\documentclass{article} 
\PassOptionsToPackage{svgnames,dvipsnames,table}{xcolor}
\usepackage{iclr2026_conference,times}

\usepackage{amsmath,amsfonts,bm}

\def\eqref#1{equation~\ref{#1}}

\def\1{\bm{1}}

\DeclareMathAlphabet{\mathsfit}{\encodingdefault}{\sfdefault}{m}{sl}
\SetMathAlphabet{\mathsfit}{bold}{\encodingdefault}{\sfdefault}{bx}{n}

\usepackage{hyperref}
\usepackage{url}
\usepackage{booktabs}
\usepackage{amsfonts}
\usepackage{nicefrac}
\usepackage{microtype}
\usepackage{xcolor}
\usepackage{colortbl}
\usepackage{amsmath}
\usepackage{amssymb}
\usepackage{graphicx}
\usepackage{wrapfig}
\usepackage{enumitem}
\usepackage{caption}
\usepackage[most]{tcolorbox}
\usepackage{tabularx}
\usepackage{amsthm}
\usepackage{bm}
\usepackage{multirow}
\tcbuselibrary{breakable}
\usepackage[capitalize,noabbrev]{cleveref}
\usepackage{titletoc}
\usepackage{tocbibind}
\usepackage{tikz}
\usepackage{marvosym}
\usepackage{subcaption}
\usepackage{threeparttable}
\usepackage[fixed]{fontawesome5}

\NewDocumentCommand{\haoran}
{ mO{} }{\textcolor{blue}{\textsuperscript{\textit{haoran}}\textsf{\textbf{\small[#1]}}}}

\NewDocumentCommand{\runzhe}
{ mO{} }{\textcolor{blue}{\textsuperscript{\textit{runzhe}}\textsf{\textbf{\small[#1]}}}}

\NewDocumentCommand{\yizhuo}
{ mO{} }{\textcolor{blue}{\textsuperscript{\textit{yizhuo}}\textsf{\textbf{\small[#1]}}}}

\usepackage[framemethod=TikZ]{mdframed}
\usepackage{xcolor}
\usetikzlibrary{shadows}
\usepackage{xpatch}
\usepackage{thmtools} 

\mdfdefinestyle{mdpurpleremark}{%
  skipabove=8pt,
  skipbelow=8pt,
  linewidth=3pt,            
  rightline=false,
  leftline=true,
  topline=false,
  bottomline=false,
  linecolor=Violet!50!black, 
  backgroundcolor=Orchid!5,  
  innertopmargin=6pt,
  innerbottommargin=6pt,
  innerleftmargin=8pt,
  innerrightmargin=8pt,
}

\declaretheoremstyle[
  headfont=\bfseries\color{Violet!50!black}, 
  bodyfont=\normalfont,
  spaceabove=0pt,
  spacebelow=0pt,
  mdframed={style=mdpurpleremark}
]{purpleremark}

\declaretheorem[style=purpleremark,name=Remark,numbered=no]{remark*}

\xpatchcmd{\endmdframed}
  {\aftergroup\endmdf@trivlist\color@endgroup}
  {\endmdf@trivlist\color@endgroup\@doendpe}
  {}{}

\mdfdefinestyle{mdpurplebox}{%
  roundcorner=10pt,
  linewidth=1pt,
  skipabove=12pt,
  skipbelow=12pt,
  innertopmargin=9pt,
  innerbottommargin=9pt,
  linecolor=black,
  nobreak=true,
  backgroundcolor=Orchid!10,
  shadow=true,
  shadowsize=6pt,
  shadowcolor=black!30,
  frametitleaboveskip=8pt,
  frametitlebelowskip=8pt,
  frametitlebackgroundcolor=Violet!50!black,
  frametitlefont=\bfseries\sffamily\color{white},
  frametitlerule=true,
}

\newcommand{\ii}{\item}

\newcommand{\NN}{\mathbb N}

\definecolor{mintblue}{RGB}{210,235,250}
\definecolor{mintframe}{RGB}{120,180,220} 
\definecolor{minttitle}{RGB}{100,150,200} 
\definecolor{minttext}{RGB}{50,80,120}    

\definecolor{runzhemilk}{RGB}{255,235,245} 
\definecolor{roseframe}{RGB}{230,120,150}  
\definecolor{runzhecotton}{RGB}{255,170,200}    

\newtcolorbox{promptbox}[1]{
  enhanced,
  breakable,
  colback= runzhemilk!30!white,   
  colframe=roseframe,                
  colbacktitle= runzhecotton!66!white, 
  coltitle=white!33,
  title=\textbf{#1},
  fonttitle=\bfseries,
  sharp corners=south, 
  borderline={0.8pt}{0pt}{roseframe},
  boxrule=0.8pt,
  arc=6pt, 
  left=6pt, right=6pt, top=6pt, bottom=6pt,
  before skip=10pt, after skip=10pt,
  drop shadow=black!12,      
}

\newtcolorbox{casebox}[1]{
enhanced,
breakable,
colback=mintblue!40!white,
colframe=mintframe,
colbacktitle=minttitle!70!white,
coltitle=white,
title=\textbf{#1},
fonttitle=\bfseries,
sharp corners=south, 
borderline={0.8pt}{0pt}{minttitle},
boxrule=0.8pt,
arc=6pt, 
left=6pt, right=6pt, top=6pt, bottom=6pt,
before skip=10pt, after skip=10pt,
drop shadow=black!15, 
}

\newtcolorbox{takeawaysbox}{
enhanced,
breakable,
colback=mintblue!40!white,
colframe=mintframe,
colbacktitle=minttitle!70!white,
coltitle=white,
title=\textbf{Key Takeaways},
fonttitle=\bfseries,
sharp corners=south, 
borderline={0.8pt}{0pt}{minttitle},
boxrule=0.8pt,
arc=6pt, 
left=6pt, right=6pt, top=6pt, bottom=6pt,
before skip=10pt, after skip=10pt,
drop shadow=black!15, 
}

\makeatletter
\newcommand{\DrawLine}{%
  \begin{tikzpicture}
  \path[use as bounding box] (0,0) -- (\linewidth,0);
  \draw[color=minttitle!70!white,dashed,dash phase=1.5pt]
        (0-\kvtcb@leftlower-\kvtcb@boxsep,0)--
        (\linewidth+\kvtcb@rightlower+\kvtcb@boxsep,0);
  \end{tikzpicture}%
  }
\makeatother

\definecolor{darkblue}{rgb}{0, 0, 0.5}
\definecolor{citeblue}{rgb}{0.35, 0.62, 0.88}
\hypersetup{colorlinks=true, citecolor=citeblue, urlcolor=citeblue, linkcolor=darkblue}

\newcommand{\goldmedalicon}{\raisebox{-0.25ex}{\includegraphics[height=1.05em]{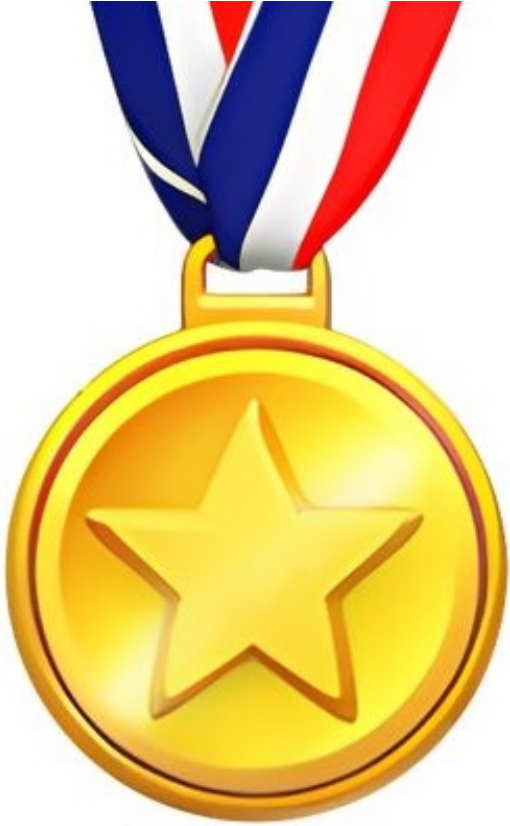}}}
\newcommand{\projectpage}{\raisebox{-1.5pt}{\faIcon{globe}}}
\newcommand{\github}{\raisebox{-1.5pt}{\includegraphics[height=1.05em]{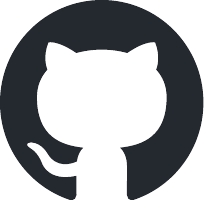}}}
\newcommand{\huggingface}{\raisebox{-1.5pt}{\includegraphics[height=1.05em]{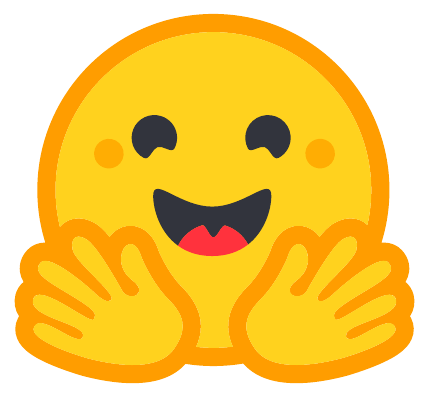}}}

\makeatletter
\def\@fnsymbol#1{%
  \ifcase#1 \or
    *%
  \or
    \textrm{\Letter}%
  \or
    \ddagger
  \or
    \S
  \or
    \P
  \or
    \|
  \or
    **%
  \or
    \dagger\dagger
  \or
    \ddagger\ddagger
  \else
    \@ctrerr
  \fi
}
\makeatother

\title{Achieving Gold-Medal-Level Olympiad Reasoning via Simple and Unified Scaling}
\author{
\textbf{Yafu Li}$^{1,2}$\thanks{~Core contributors. Yafu Li is the project lead.}\kern0.45em\thanks{~Corresponding authors. Contact: \texttt{yafuly@gmail.com} and \texttt{chengyu@cse.cuhk.edu.hk}.}, \enspace
\textbf{Runzhe Zhan}$^{1}$\footnotemark[1], \enspace
\textbf{Haoran Zhang}$^{1,4}$\footnotemark[1], \enspace
\textbf{Shunkai Zhang}$^{1,5}$\footnotemark[1], \enspace
\textbf{Yizhuo Li}$^{1}$\footnotemark[1], \enspace\\
\textbf{~Zhilin Wang}$^{1}$, \enspace
\textbf{Jiacheng Chen}$^{2}$, \enspace
\textbf{Futing Wang}$^{1}$, \enspace
\textbf{Xuyang Hu}$^{1}$, \enspace
\textbf{Yuchen Fan}$^{1}$, \enspace\\
\textbf{~Bangjie Xu}$^{3}$, \enspace
\textbf{Yucheng Su}$^{3}$, \enspace
\textbf{Xinmiao Han}$^{3}$, \enspace
\textbf{Chenxi Li}$^{1}$, \enspace
\textbf{Haodi Lei}$^{1}$, \enspace
\textbf{Yufeng Zhao}$^{1}$, \enspace\\
\textbf{~Zejin Lin}$^{3}$, \enspace
\textbf{Qianjia Cheng}$^{1}$, \enspace
\textbf{Tong Zhu}$^{1}$, \enspace
\textbf{Xiaoye Qu}$^{1}$, \enspace
\textbf{Ganqu Cui}$^{1}$, \enspace
\textbf{Peng Ye}$^{1}$\footnotemark[2], \enspace\\
\textbf{~Yun Luo}$^{1}$\footnotemark[2], \enspace
\textbf{~Zhouchen Lin}$^{5}$, \enspace
\textbf{~Yu Qiao}$^{1}$, \enspace
\textbf{Bowen Zhou}$^{1,3}$\footnotemark[2], \enspace
\textbf{Ning Ding}$^{3,1}$\footnotemark[2], \enspace
\textbf{Yu Cheng}$^{2,1}$\footnotemark[2]
\\
$^{1}$Shanghai AI Laboratory \quad
$^{2}$The Chinese University of Hong Kong\quad
$^{3}$Tsinghua University \\
$^{4}$Shanghai Jiao Tong University \quad
$^{5}$Peking University
}

\iclrfinalcopy
\begin{document}

\maketitle
\vspace{-1.45em}
\begin{abstract}
Recent progress in reasoning models has substantially advanced long-horizon
mathematical and scientific problem solving, with several systems now reaching
gold-medal-level performance on International Mathematical Olympiad (IMO) and
International Physics Olympiad (IPhO) problems.
In this
paper, we introduce a simple and unified recipe for converting a post-trained
reasoning backbone into a rigorous olympiad-level solver. The recipe first uses
a reverse-perplexity curriculum for SFT to instill rigorous
proof-search and self-checking behaviors, then scales these behaviors through a
two-stage RL pipeline that progresses from RL with
verifiable rewards to more delicate proof-level RL, and finally boosts solving
performance with test-time scaling. 
Applying this recipe, we train a \textit{30B-A3B}
backbone with SFT on around \textit{340K sub-8K-token} trajectories followed by \textit{200} RL
steps. 
The resulting model, \textbf{\textcolor{iclrdeepblue}{SU-01}}, supports
  stable reasoning on difficult problems with trajectories exceeding \textit{100K} tokens,
  while achieving gold-medal-level performance on mathematical and physical
  olympiad competitions, including \textbf{IMO 2025/USAMO 2026} and \textbf{IPhO 2024/2025}. 
  It also demonstrates strong generalization of scientific reasoning to domains beyond mathematics and physics.
\vspace{-0.5em}
\end{abstract}

\begin{center}
\vspace{-1em}
~\projectpage~\href{http://simplified-reasoning.github.io/SU-01}{{\text{Project Page}}}
\quad \quad \quad
~\github~\href{https://github.com/Simplified-Reasoning/SU-01}{{\text{Code}}}
\quad \quad \quad
~\huggingface~\href{https://huggingface.co/Simplified-Reasoning/SU-01}{{\text{Models}}}
\end{center}

\begin{center}
\vspace{-1em}
\includegraphics[width=0.8\textwidth]{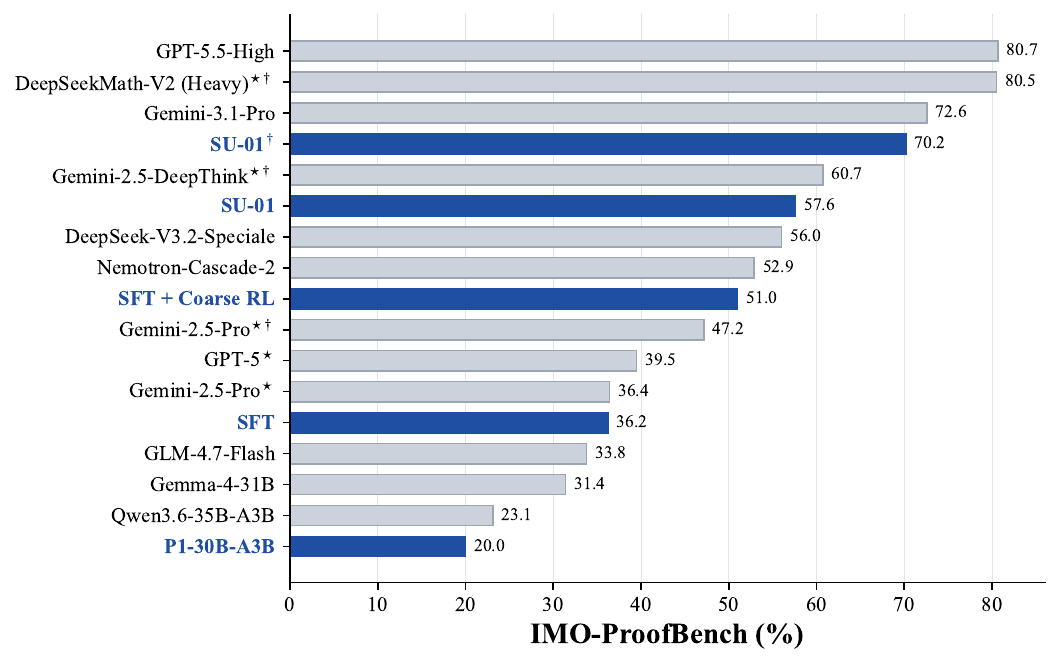}
\captionsetup{type=figure,hypcap=false,skip=2pt}
\captionof{figure}{Overall comparison on IMO-ProofBench. $\star$ denotes results reported in the original paper, and $\dagger$ denotes results with test-time scaling. Blue bars trace the evolution of our pipeline from the 30B-A3B backbone through rigorous SFT (\cref{sec:sft}), coarse RL, refined RL (\cref{sec:rl}), and test-time scaling (\cref{sec:tts}), culminating in gold-medal-level olympiad reasoning (\cref{sec:exp-results}).}
\label{fig:proofbench-overall}
\end{center}

\section{Introduction}
Olympiad competitions provide one of the clearest stress tests for long-horizon
reasoning. Unlike many standard benchmarks, these problems require a model to 
search over many possible solution paths, control assumptions precisely, verify
intermediate claims, and present a final argument that can survive strict
grading across mathematical and scientific settings. Recent systems have made
rapid progress in this direction: AlphaGeometry combined neural guidance with
symbolic search for olympiad geometry~\citep{trinh2024alphageometry}, while
AlphaProof, AlphaGeometry 2, and Gemini Deep Think reached silver- or
gold-medal standards on International Mathematical Olympiad problems with
larger search and verification budgets~\citep{deepmind2024alphaproof,deepmind2025geminiimo}.
At the same time, general reasoning models have improved through
chain-of-thought prompting, math-specialized post-training, and reinforcement
learning with verifiable rewards~\citep{wei2022chain,shao2024deepseekmath,yang2024qwen25math,r1,yan2025learning,exgrpo}, while scientific olympiad benchmarks test transfer to modeling, derivation, and competition-style justification~\citep{olympiadbench,chen2025p1,luo2026p1vl}.


A central question is therefore whether \textit{a reasoning backbone can be pushed to olympiad-level performance with a compact, domain-unified recipe} that
applies the same reasoning-centric pipeline across mathematical and scientific problems. 
Using
a 30B-A3B model, we build a modular pipeline: SFT reshapes reasoning behavior, RL scales solving capability, and TTS allocates additional inference compute to the
hardest proof-search problems. Together, these stages align behavior shaping, reward design, experience replay, and self-verification into a compact recipe for rigorous
mathematical and scientific reasoning. The desgin follows a specializable-generalist view: rather than building a narrow olympiad solver, we specialize a broadly capable post-trained model toward expert-level proof reasoning while preserving transfer across scientific domains.
  
\textbf{The first stage aims to instill a more disciplined proof-search pattern.} 
Starting from a post-trained model that is already competitive on scientific reasoning tasks,
we curate long-form solution, self-verification, and self-refinement
trajectories from mathematical, scientific, coding, and instruction-following
sources. 
After filtering, the SFT mixture contains 338K
trajectories with responses shorter than 8K tokens. SFT on this rigorous proof
data instills reasoning behaviors centered on proof search, self-checking, and
repair. We then order the examples by reverse perplexity so
that each pass starts with trajectories most mismatched to the initial policy
before consolidating on more familiar examples. This curriculum helps preserve
and recover the capability of the post-trained model with its reasoning
behavior reshaped.

\textbf{The second stage scales this behavior through two levels of RL.}
Coarse RL uses
verifiable prompts and efficient outcome checking to scale the reasoning
behaviors introduced by SFT under reliable binary rewards, following the broader RLVR paradigm for efficient reasoning improvement~\citep{r1,shao2024deepseekmath}. 
Refined RL then shifts the target from answer correctness to proof quality.  It combines a proof-level generative reward model for scoring complete proofs,
self-refinement prompts for training critique-and-repair behavior, and
experience replay for preserving rare successful trajectories on hard
problems.
Finally, we apply test-time scaling through a self-verification-and-refinement loop to elevate the trained
model to olympiad-level reasoning~\citep{huang2025winning}.

On answer-verifiable benchmarks, the resulting model,
\textbf{\textcolor{iclrdeepblue}{SU-01}}, nearly matches the strongest
similar-size baseline, Qwen3.6-35B-A3B, across AnswerBench, AMO-Bench, AIME
2025/2026, and FrontierScience-Olympiad. On proof-oriented evaluation, SU-01
reaches 57.6\% on IMO-ProofBench with direct generation and 70.2\% with TTS,
substantially outperforming similar-size models and approaching competitive
commercial systems such as Gemini 3.1 Pro Thinking. Beyond solving
competition problems, SU-01 obtains the best similar-size overall score on
FrontierScience-Research, suggesting that the recipe generalizes scientific
reasoning toward research-style problems beyond olympiad benchmarks.

On official competition problems, SU-01 shows strong end-to-end reasoning beyond benchmark-style evaluation. 
Direct SU-01 already exceeds the IPhO gold lines for
both 2024 and 2025, and clears the bronze-medal lines on IMO 2025 and USAMO 2026. 
With test-time scaling, it reaches \textbf{35 points} on both mathematical olympiads, meeting the \textbf{IMO
2025 gold} line and exceeding the \textbf{USAMO 2026 gold} line by 10 points. 
Notably, on USAMO 2026, this matches the highest reported human total among 340 competitors,
indicating that the overall recipe can elicit top-level human-like olympiad reasoning from a compact 30B-A3B model. 
The TTS traces further show how this
capability emerges at inference time: SU-01 can sustain reasoning trajectories beyond 100K tokens, condition on its own drafts and error analyses, and repeatedly
verify and repair candidate proofs. 
Overall, these results support a specializable-generalist view of compact reasoning
models: with the right training and inference recipe, a broadly capable backbone
can be driven toward expert-level proof reasoning while retaining meaningful
scientific transfer. 

\begin{figure*}[t]
\centering
\includegraphics[width=\textwidth]{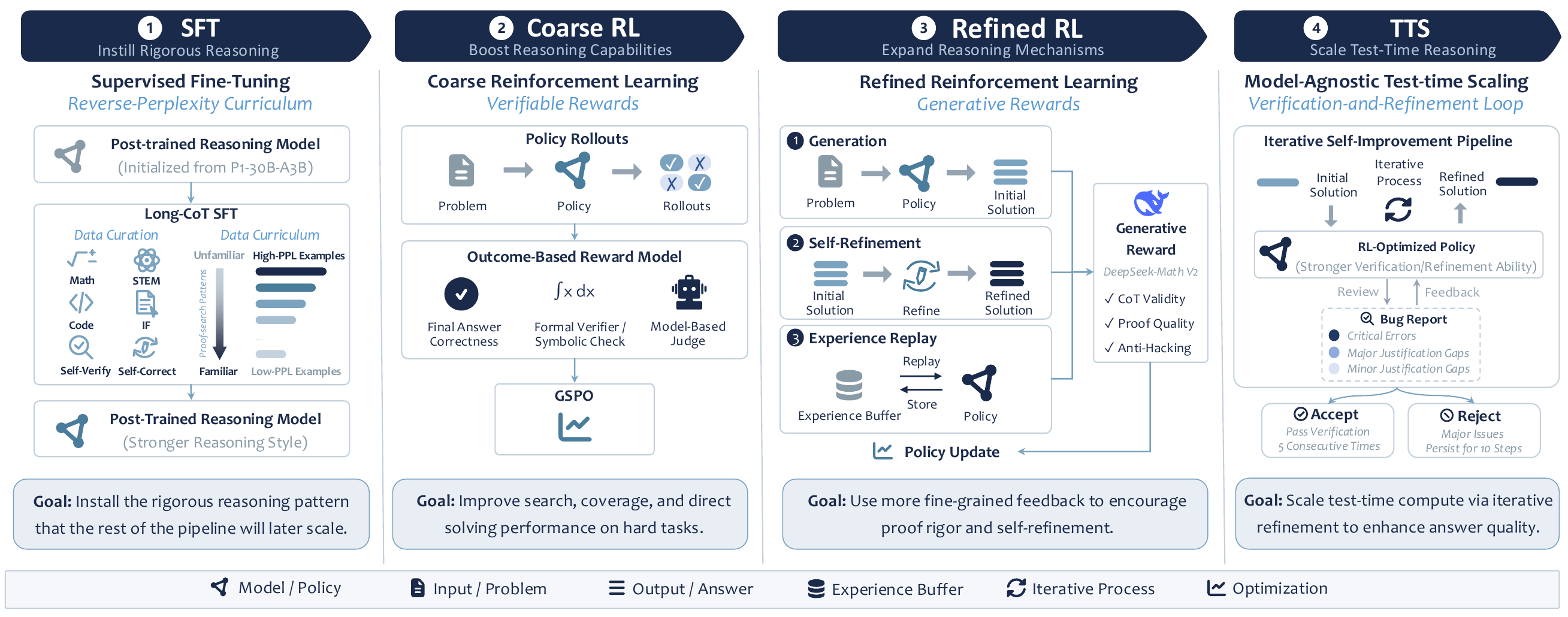}
\vspace{-1.5em}
\caption{Overview of the SU-01 training and inference pipeline. The recipe first reshapes the backbone with rigorous long-form SFT, then scales the resulting behavior through coarse and refined RL, and finally applies test-time verification and refinement for olympiad-level problem solving.}
\vspace{-0.5em}
\label{fig:simplex-pipeline}
\vspace{-4 pt}
\end{figure*}
\section{Instilling Rigorous Reasoning via SFT}
\label{sec:sft}

The first stage of the SU-01 pipeline uses supervised fine-tuning to
reshape the model's reasoning behavior. We choose P1-30B-A3B~\citep{chen2025p1} as the initial model
because it already shows competitive performance in scientific reasoning,
including both mathematics and physics. Despite its strong results on
verifiable tasks, we observe that its solutions are not always organized around
rigorous proof-search patterns. The purpose of SFT is therefore to reshape its
reasoning behavior toward more explicit, disciplined, and proof-oriented
long-form reasoning while preserving as much of its existing capability as
possible.

We empirically find that applying SFT to a post-trained backbone is more
efficient than training the same reasoning behavior from a base model. A
post-trained model already contains useful instruction-following behavior,
problem-solving ability, and broad scientific competence. Starting from
that checkpoint allows SFT to focus on changing the reasoning pattern rather
than rebuilding these capabilities from scratch. 
In this framing, SFT specializes the generalist backbone toward rigorous proof-search behavior while preserving its broad scientific competence, providing a stronger
  starting policy for subsequent RL to scale. 
The launch configuration and optimization
hyperparameters for this stage are summarized in
\cref{app:sft-training-details}. 


\subsection{SFT Data Curation}
\label{sec:sft-data-curation}

We curate SFT prompts from a broad mixture of mathematical, scientific,
instruction-following, and coding sources. The mathematical subset includes
problems from Evan Chen's olympiad materials\footnote{Evan Chen's olympiad
materials: \url{https://web.evanchen.cc/}.}, the Shuzhimi Forum\footnote{The
Shuzhimi Forum is an online Chinese mathematical problem-solving community.},
AoPS (Art of Problem Solving)\footnote{AoPS: \url{https://artofproblemsolving.com/}.},
online mathematical competition training books\footnote{The book subset is
curated from publicly available online mathematical competition training
materials.}, and DeepMath problems with difficulty at least 6
\citep{he2025deepmath}. For scientific reasoning, we include prompts from
NaturalReasoning \citep{yuan2025naturalreasoning}. To improve the
generalization of the SFT model beyond narrow olympiad-style mathematics, we
also include chat prompts from Nemotron-Instruction-Following-Chat-v1\footnote{Nemotron-Instruction-Following-Chat-v1 Hugging Face dataset card:
\href{https://huggingface.co/datasets/nvidia/Nemotron-Instruction-Following-Chat-v1}{link}.}
and coding prompts from Eurus-2-RL-Data \citep{cui2025process}
and OpenCodeReasoning-2\footnote{OpenCodeReasoning-2 Hugging Face dataset card:
\href{https://huggingface.co/datasets/nvidia/OpenCodeReasoning-2}{link}.}; the latter
extends the OpenCodeReasoning data-distillation line for competitive coding
\citep{ahmad2025opencodereasoning}.

\begin{wrapfigure}{r}{0.46\linewidth}
\vspace{-8pt}
\centering
\includegraphics[width=\linewidth]{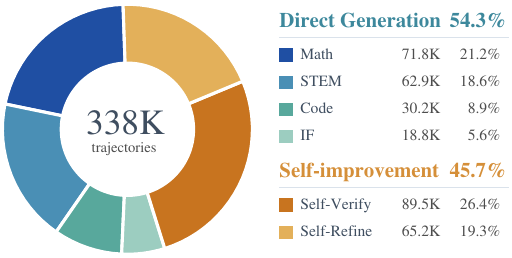}
\caption{Composition of the SFT data after filtering. Math, STEM, Code, and IF form the direct-generation group; Self-Verify and Self-Refine form the self-improvement group.}
\label{fig:sft-data-category}
\vspace{-12pt}
\end{wrapfigure}

Before generation, we first filter contaminated problems from the prompt pool.
For each remaining prompt, we use DeepSeek-V3.2-Speciale
\citep{deepseekai2025deepseekv32} to generate high-quality long-form reasoning
trajectories. We then filter noisy generations and remove trajectories longer
than 8,192 tokens. This filtering step keeps the supervised signal focused on
rigorous and usable reasoning traces, while avoiding extremely long outputs
that are more likely to introduce truncation or unstable optimization.

In addition to direct solution trajectories, we further equip the model with
self-verification and self-refinement behaviors. 
For the mathematical subset, we ask DeepSeek-V3.2-Speciale to generate verification traces for the generated solutions, followed by refinement traces that address issues identified during verification. 
These examples expose the model to the behaviors that are
especially important for olympiad-level reasoning: checking whether a proof is
actually justified and improving an argument when a flaw is found. Finally, we
obtain a filtered SFT mixture of 338K trajectories, as shown in
\cref{fig:sft-data-category}.

\subsection{Reverse-Perplexity Curriculum for SFT}
\label{sec:reverse-ppl-curriculum}

Long-CoT SFT on a post-trained reasoning model is a delicate optimization
problem. The model already contains a strong instruction-following and
reasoning policy, so SFT is not simply adding a new capability to an empty
backbone; it is modifying an existing policy while trying to preserve its
original competence. 
If the supervised signal is too narrow or the training is
stopped too early, performance can degrade substantially even when the model
starts to imitate more explicit long-form reasoning. This tension is consistent
with the long-CoT degradation phenomenon studied by \citet{luo2025through}: a
post-trained model often needs sufficient data scale and enough SFT epochs to
absorb the new reasoning style without overwriting the useful competence
installed by previous post-training stages.

In our setting, recovery depends strongly on both training duration and the
length behavior of the resulting model \citep{ren2026rethinking}. For trajectories capped at 8,192 tokens,
we empirically find that four epochs are usually sufficient to recover most of
the model capability after the initial behavioral shift, provided that the data
mixture and learning rate are well controlled. 
We also treat validation truncation rate as an operational indicator
of SFT sufficiency. A post-trained model that has
not been sufficiently adapted to rigorous long-CoT supervision often exhibits
shallow reasoning behaviors: it circles around local heuristics, repeats
intermediate claims, and continues reasoning without making decisive progress.
These repetitive and endless-reasoning patterns naturally increase truncation.
In practice, we find that a truncation rate below 5\% is a useful sign that the model has
largely adapted to the target reasoning style.

To make long-CoT SFT more stable, we use a reverse-perplexity training curriculum. Let
\(\mathcal{D}=\{(x_i,y_i)\}_{i=1}^{N}\) be the SFT set, where \(x_i\) is the
prompt and \(y_i=(y_{i,1},\ldots,y_{i,T_i})\) is the teacher trajectory. Given
the initial policy \(\pi_0\), we score each example by its length-normalized
perplexity, \(\mathrm{PPL}(x_i,y_i)=\exp\left(-\frac{1}{T_i}\sum_{t=1}^{T_i}
\log \pi_0(y_{i,t}\mid x_i,y_{i,<t})\right)\).
Instead of presenting examples in random order or in ascending perplexity, we
sort the data in descending perplexity and train from high-PPL examples to
low-PPL examples within each epoch. This order repeatedly starts each pass from
teacher trajectories that are most mismatched with the current policy, using
unfamiliar proof-search patterns for behavioral adaptation before consolidating
on more familiar examples. We discuss the empirical effect of this ordering in
\cref{sec:ppl-order-analysis}.

\section{Boosting Reasoning Capability with RL}
\label{sec:rl}

Once the model has acquired a stronger long-form reasoning pattern,
reinforcement learning provides the scalable feedback mechanism for turning this
pattern into stronger expert behavior. We split this stage into two levels.
\textbf{Coarse RL} converts the SFT reasoning pattern into stronger
answer-seeking behavior under reliable, mostly verifiable reward signals,
improving search, coverage, and direct solving performance on hard tasks.
\textbf{Refined RL} then specializes the policy toward complete, auditable proof
construction, using more fine-grained feedback to encourage proof rigor and
self-refinement.
The shared
RL launch configuration and stage-specific hyperparameters are summarized in
\cref{app:rl-training-details}.

\subsection{RL Data Curation}

RL training uses a separate prompt pool from SFT, curated to support both
answer-verifiable optimization and proof-quality refinement. The physics subset
is derived from olympiad-level physics data associated with P1
\citep{chen2025p1}. The mathematical subset follows the same source families as
our SFT data, including AoPS, online competition training books, Evan Chen's
olympiad materials, and the Shuzhimi Forum. We refer readers to
\S\ref{sec:sft-data-curation} for source attribution. We additionally include
OPC\footnote{OPC dataset card:
\href{https://huggingface.co/datasets/INSAIT-Institute/OPC}{link}.}, a
human-evaluated corpus of advanced mathematical proofs \citep{dekoninck2025opc},
to increase coverage of proof-oriented prompts.

We split the resulting RL pool into a verifiable set and a non-verifiable set.
The verifiable set contains prompts whose final answers or structured outputs
can be checked reliably, while the non-verifiable set includes proof-oriented
or open-ended reasoning prompts that require softer judgment, e.g., generative reward. 
Before training,
we first deduplicate and decontaminate the prompt pool. We then apply rejection
sampling to remove examples that are already too easy or too hard for the
current policy, and further filter noisy prompts that are poorly formatted or otherwise unreliable. The final RL pool
contains 8,967 verifiable prompts and 16,287 non-verifiable prompts.

\subsection{Coarse RL}
\label{sec:coarse-rl}

Coarse RL trains the SFT model on the 8,967 verifiable prompts described above.
We formulate this stage as reinforcement learning with verifiable rewards
(RLVR; \citealt{lambert2024tulu,r1}), using Group Sequence Policy Optimization
(GSPO; \citealt{zheng2025gspo}). GSPO is better aligned with outcome-reward
training than token-level GRPO because both reward assignment and policy
clipping operate at the complete-response level. For each prompt
$q \in \mathcal{D}_{\mathrm{ver}}$ (the verifiable prompt set), the rollout policy
$\pi_{\theta_{\mathrm{old}}}$ samples a group of $K$ candidate solutions
$\mathcal{G}_q=\{o_i\}_{i=1}^K$. The verifier converts each final answer into a
binary outcome reward $r(q,o)=1$ if the extracted final answer is verified as
correct, and $r(q,o)=0$ otherwise.
The group-relative advantage is computed from the within-prompt reward baseline.
We use the unnormalized form without group standard-deviation normalization,
$\widehat{A}_i=r(q,o_i)-\mu_{\mathcal{G}_q}$, where
$\mu_{\mathcal{G}_q}=\frac{1}{K}\sum_{j=1}^{K}r(q,o_j)$.
The key GSPO quantity is the length-normalized sequence-level importance ratio
$s_i(\theta)=\exp\{\frac{1}{|o_i|}\sum_{t=1}^{|o_i|}\log
\frac{\pi_{\theta}(o_{i,t}\mid q,o_{i,<t})}
{\pi_{\theta_{\mathrm{old}}}(o_{i,t}\mid q,o_{i,<t})}\}$. The policy is
updated with the clipped sequence-level surrogate
\begin{equation}
\mathcal{J}_{\mathrm{GSPO}}(\theta)=
\mathbb{E}_{q,\{o_i\}}
\left[
\frac{1}{K}\sum_{i=1}^{K}
\min\left(
s_i(\theta)\widehat{A}_i,
\operatorname{clip}(s_i(\theta),1-\epsilon,1+\epsilon)\widehat{A}_i
\right)
\right].
\label{eq:coarse-gspo-objective}
\end{equation}
These definitions are also the interface used by the experience replay variants in the subsequent subsection: 
replayed trajectories can reuse the same reward, advantage, and sequence-ratio
notation while changing the source policy in the denominator of $s_i(\theta)$.
Following the routing-replay motivation in GSPO \citep{zheng2025gspo}, we freeze
the MoE router during RL so replayed trajectories are evaluated under stable
expert-routing decisions, which reduces replay-induced instability.

The reward system is intentionally layered to keep high-precision automatic
checks before more expensive model-based judgments. We first extract the final
answer and apply canonicalized text matching. Unresolved cases are then checked by Math-Verify\footnote{Math-Verify
repository: \href{https://github.com/huggingface/Math-Verify}{link}.}, a
rule-based mathematical-expression evaluation pipeline for LLM outputs. Samples
that still fail these rule-based checks are sent to
gpt-oss-120b\footnote{gpt-oss-120b model card:
\href{https://huggingface.co/openai/gpt-oss-120b}{link}.} \citep{openai2025gptoss}
for generative verification. This ordering makes the reward conservative by
default, while still recovering correct solutions whose final answers are
equivalent but difficult to normalize with rule-based parsers alone.

\subsection{Refined RL}

After coarse RL has established strong search behavior, refined RL shifts the
optimization target from answer correctness to proof quality. The central issue
is that many olympiad solutions can reach a correct final answer while still
containing hidden gaps, unjustified transformations, or incomplete case
analysis. Refined RL therefore uses a stronger process-level reward and adds two
memory mechanisms: self-refinement, which turns recent failures into repair
tasks, and experience replay, which preserves rare successful proofs long
enough for the policy to learn from them.

\paragraph{Generative proof reward.}
We use DeepSeekMath-V2 as a generative reward model for refined RL
\citep{deepseekmathv2}, except for physics prompts. 
For every rollout from both the verifiable and
non-verifiable subsets, the reward model reads the problem and the complete
solution or proof, then outputs a binary score $r_{\mathrm{proof}}(q,o)\in
\{0,1\}$. 
Unlike the coarse verifier in \S\ref{sec:coarse-rl}, this score is not restricted to
checking whether the final answer matches a reference answer. 
It evaluates
whether the full reasoning path is mathematically valid, sufficiently rigorous,
and complete. This makes the reward more aligned with the final goal of
olympiad reasoning, but also more expensive and more vulnerable to judge
artifacts. We therefore apply anti-hack preprocessing before sending a response
to the reward model: malformed generations with leaked chat-template tokens,
unbalanced thinking delimiters, or severe repetition are replaced by a safe
fallback answer. This prevents the policy from receiving reward by exploiting
formatting or verifier-input pathologies rather than by improving the proof. The
reward-model serving configuration is summarized in
\cref{app:inference-serving-details}.

\paragraph{Self-refinement.}
Self-refinement exposes the policy to the same repair pattern that we use at
test time: propose a solution, inspect it, locate gaps, and produce a corrected
proof. After each rollout, responses are grouped by query. If a query group has
average proof reward below a threshold $\tau_{\mathrm{ref}}=0.5$, failed
responses from that group are converted into refinement prompts. Each prompt
contains the original problem, the previous incorrect solution, and an
instruction to critique the argument, fix proof 
errors, fill
missing justifications, and output a complete final solution. These prompts are
stored in a self-refinement buffer and mixed into subsequent batches with target
ratio $\eta_{\mathrm{ref}}=0.2$. Normal samples displaced by refinement queries
are returned to a buffer, so refinement does not silently discard fresh
training data. We also do not recursively enqueue failed refinement attempts,
which avoids spending repeated updates on examples that remain outside the
current policy's learnable region.

\paragraph{Experience replay.}
On difficult proof problems,
the policy may occasionally discover a valid solution trajectory even though it
usually fails on the same query. Immediately discarding such a trajectory wastes
a high-value training signal. Following ExGRPO
\citep{exgrpo}, we keep a replay buffer $\mathcal{E}$ keyed by query, but our
implementation is simpler: it uses the same GSPO-style update and does not apply
the policy-shaping transform introduced in ExGRPO. After each rollout, a query
is admitted to the replay buffer only when it is hard but solvable, operationalized as
$0<n_+(q)<2$, where $n_+(q)$ is the number of successful trajectories in the
current group. In answer-only RLVR, such a unique success can be a lucky hit: a
trajectory may end with the correct final answer while still containing brittle
or invalid reasoning\citep{exgrpo}. 
In our refined RL setting, however, success is assigned by the
DeepSeekMath-V2 proof reward, which inspects the full solution rather than only
the final answer. This does not eliminate reward-model noise, but it makes a
rare successful rollout substantially more likely to encode a reusable proof
pattern and therefore a safer replay target. Stored trajectories are
deduplicated, and a query is retired once fresh on-policy rollouts solve it
often enough, with threshold $n_+(q)\ge 4$.

Replay is mixed with fresh proof-reward training rather than run as a separate
mode. A replay rollout injects one stored successful trajectory for the selected
query, and the replay ratio is controlled by $\rho=0.25$ over the non-refinement
portion of the batch. When multiple successful trajectories are stored for a
query, we select the lowest-entropy one, $o^*=\arg\min_{o\in\mathcal{E}(q)}
H(o;\pi_\theta)$, using rollout-side top-$k$ log probabilities as an efficient
entropy estimate, following the trajectory-selection principle in ExGRPO
\citep{exgrpo}. The resulting refined objective is
\begin{equation}
\mathcal{J}_{\mathrm{refined}}(\theta)=
(1-\rho)\,\mathbb{E}_{\mathcal{B}_{\mathrm{fresh}}}
\left[\mathcal{J}_{\mathrm{GSPO}}(q,\mathcal{G}_q;\theta,\pi_{\theta_{\mathrm{old}}})\right]
+\rho\,\mathbb{E}_{\mathcal{B}_{\mathrm{exp}}}
\left[\mathcal{J}_{\mathrm{GSPO}}(q^*,\{o^*\}\cup\mathcal{G}_{q^*};\theta,\pi_{\theta_{\mathrm{src}}})\right],
\label{eq:refined-replay-objective}
\end{equation}
where $\pi_{\theta_{\mathrm{src}}}=\pi_{\theta_{\mathrm{past}}}$ for the replayed
trajectory and $\pi_{\theta_{\mathrm{src}}}=\pi_{\theta_{\mathrm{old}}}$ for fresh
rollouts. This replay design is targeted rather than exhaustive: it stores rare
valid proofs, prefers the most stable stored trajectory, replays it at a
controlled ratio, and removes it once the current policy can reliably reproduce
the behavior on-policy.

\section{Achieving Gold-Medal-Level Reasoning via Test-time Scaling}
\label{sec:tts}

Even with a
strong reasoning policy, the hardest problems often require substantial search
and revision at inference time. This is not merely a matter of sampling more
answers. IMO-style tasks demand complete and rigorous proofs, and a solution
with the right final conclusion can still fail if it contains a hidden gap or a
logical fallacy. Recent work on IMO 2025 makes this point explicit: strong
frontier models already contain significant mathematical capability, but their
single-pass outputs and even best-of-many selection can remain far below the
level obtained by a structured verification-and-refinement pipeline
\citep{huang2025winning}. 

The need for test-time scaling is also tied to reasoning budget. A single generation has a
finite context and thinking budget, while an olympiad proof may require several
rounds of exploration, lemma checking, counterexample search, and exposition
repair. A model can spend most of its budget discovering a promising approach
and still fail to fully close the proof. Breaking inference into repeated
solve--verify--refine stages effectively allocates additional computation to the
same problem while keeping each step focused and auditable. This extra budget is
useful only when the model can remain coherent across repeated drafts,
critiques, and repairs. 
After the full training pipeline, SU-01 is able to use
this budget productively on difficult problems, sustaining coherent reasoning
trajectories longer than 100K tokens during inference.

Our TTS procedure follows the verification-and-refinement paradigm of
\citet{huang2025winning} as a self-verification and refinement loop. The model
first produces an initial solution under a solver prompt that prioritizes proof
rigor rather than merely reaching a final answer. It then enters refinement,
where it revisits the draft, repairs weak steps, and tries to turn a promising
argument into a complete proof. The refined candidate is next checked through a
verification prompt: the model inspects the full solution and writes a structured
bug report, identifying issues such as critical errors, unjustified claims, or
missing cases. A verdict step interprets this report and decides whether the
candidate should be accepted, rejected, or sent back for another refinement
round. This loop is repeated until the solution consistently passes
self-verification or the refinement budget is exhausted. Multiple independent
runs can be executed in parallel or serial, and accepted candidates are selected
only after the proof is stable under repeated verification. The corresponding
inference setting is summarized in \cref{app:inference-serving-details}.


\section{Experimental Results}
\label{sec:exp-results}
The experimental section is organized around three complementary evaluation
views: answer-verifiable reasoning tasks, non-verifiable or proof-oriented
tasks, and official olympiad competition problems. 

\subsection{Benchmarks}

We organize evaluation into three benchmark families. The first family contains
answer-verifiable reasoning tasks, where correctness can be checked by a final
answer or a high-confidence automatic verifier. It includes AMO-Bench~\citep{an2025amobench},
AIME 2025 and AIME 2026\footnote{AIME official competition page:
\url{https://maa.org/math-competitions/aime}.}, AnswerBench from the
IMO-Bench evaluation suite~\citep{luong2025imobench}, and FrontierScience-Olympiad,
the Olympiad subset of FrontierScience~\citep{wang2026frontierscience}. These
benchmarks mainly test whether the model can produce correct final answers
under single-pass or fixed-budget inference.

The second family contains non-verifiable or proof-oriented tasks. We include
ProofBench from IMO-Bench~\citep{luong2025imobench}, which emphasizes proof
quality rather than only final-answer matching, and FrontierScience-Research, the
research subset of FrontierScience~\citep{wang2026frontierscience}. These tasks
are used to probe whether training improves rigorous reasoning and scientific
problem solving beyond answer-checkable settings.

The third family contains official olympiad competition problems, including
IMO 2025\footnote{International Mathematical Olympiad official archive:
\url{https://www.imo-official.org/}.}, USAMO 2026\footnote{USA
Mathematical Olympiad:
\url{https://maa.org/math-competitions/usamo}.} and IPhO (2024, 2025)\footnote{International Physics
Olympiad official site: \url{https://www.ipho-new.org/}.}. 
Detailed grading and verifier settings are summarized
in \cref{app:evaluation-details}.

\definecolor{MathReasoningBg}{RGB}{255,244,204}
\definecolor{ScienceReasoningBg}{RGB}{234,242,255}
\begin{table*}[t]
\centering
\caption{Performance on answer-verifiable reasoning tasks. Results for AnswerBench, AMO-Bench, AIME 25/26, and FrontierScience-Olympiad are averaged over 4, 8, 8, and 4 runs, respectively. FrontierScience-Olympiad abbreviates the Olympiad subset of FrontierScience. Avg. is the mean of AnswerBench, AMO-Bench, AIME 2025, AIME 2026, and FrontierScience-Olympiad. Within each comparison block, \textbf{bold} marks the best score and underline marks the second best.}
\label{tab:verifiable-single-pass}
\vspace{-5 pt}
\setlength{\tabcolsep}{2.2pt}
\renewcommand{\arraystretch}{1.14}
\small
\resizebox{\textwidth}{!}{%
\begin{tabular}{lcccccccc}
\toprule
\multirow{2}{*}{\textbf{Model}} & \multirow{2}{*}{\textbf{AnswerBench}} & \multirow{2}{*}{\textbf{AMO-Bench}} & \multirow{2}{*}{\textbf{AIME 25/26}} & \multicolumn{4}{c}{\textbf{FrontierScience-Olympiad}} & \multirow{2}{*}{\textbf{Avg.}} \\
\cmidrule(lr){5-8}
 & & & & \textbf{Physics} & \textbf{Chemistry} & \textbf{Biology} & \textbf{Overall} & \\
\midrule
P1-30B-A3B & 69.3\% & 41.3\% & 90.4\% / 89.6\% & 57.5\% & 57.5\% & \underline{27.5\%} & 54.5\% & 69.0\% \\
GLM-4.7-Flash & 73.8\% & 53.8\% & 91.3\% / 88.3\% & 54.5\% & 60.0\% & 17.5\% & 53.0\% & 72.0\% \\
Nemotron-Cascade-2 & \textbf{80.5\%} & 40.8\% & \underline{94.2\%} / 90.0\% & 56.0\% & 56.3\% & \textbf{30.0\%} & 53.5\% & 71.8\% \\
Qwen3.6-35B-A3B & \underline{78.0\%} & \underline{58.8\%} & 92.5\% / \underline{92.9\%} & \underline{65.5\%} & \textbf{74.4\%} & 25.0\% & \textbf{65.0\%} & \textbf{77.4\%} \\
Gemma-4-31B & 74.0\% & 39.3\% & 88.8\% / 91.3\% & \textbf{69.0\%} & 61.9\% & 17.5\% & 61.0\% & 70.9\% \\
\rowcolor{ScienceReasoningBg!35}
\textbf{\textcolor{iclrdeepblue}{SU-01}} & 77.5\% & \textbf{59.8\%} & \textbf{94.6\%} / \textbf{93.3\%} & 62.5\% & \underline{69.4\%} & 25.0\% & \underline{61.5\%} & \underline{77.3\%} \\
\bottomrule
\end{tabular}%
}
\end{table*}

\subsection{Verifiable Problems}

  As shown in
  \cref{tab:verifiable-single-pass}, SU-01 reaches a 77.3\% average score across
  AnswerBench, AMO-Bench, AIME 2025, AIME 2026, and FrontierScience-Olympiad,
  nearly matching the strongest similar-size baseline, Qwen3.6-35B-A3B (77.4\%).
  Importantly, SU-01 achieves this level of performance with a simpler unified
  post-training recipe and substantially lower training cost, highlighting the
  efficiency of our approach.
The mathematical benchmarks show where this improvement is most pronounced.
SU-01 achieves the best similar-size results on AMO-Bench (59.8\%) and AIME
2025/2026 (94.6\%/93.3\%), which are closer to competition-style problem solving
than routine answer extraction. On AnswerBench, SU-01 remains competitive at
77.5\%, close to Qwen3.6-35B-A3B (78.0\%) and behind only Nemotron-Cascade-2
(80.5\%). 
  FrontierScience-Olympiad tests whether this behavior transfers beyond pure
  mathematics. Although the RL stages use only math and physics signals, SU-01
  reaches 61.5\% overall and shows strong transfer to untrained STEM domains,
  including 69.4\% on Chemistry and 25.0\% on Biology. This cross-domain transfer
  supports the specializable-generalist framing: the model is specialized through
  math and physics reasoning signals, yet the resulting capability does not
  collapse into a narrow contest solver.

\subsection{Non-verifiable Problems}

Non-verifiable benchmarks test whether the training recipe improves the quality
of full reasoning traces, rather than only optimizing final-answer rewards. On
IMO-ProofBench, \cref{tab:nonverifiable-benchmarks} shows that SU-01 reaches
57.6\% overall in direct generation, already the strongest result among
similar-size models. Test-time scaling further raises the score to 70.2\%,
including 91.0\% on the basic split and 49.5\% on the advanced split, bringing a
30B-A3B model close to much larger frontier systems such as Gemini 3.1 Pro. This improvement indicates
that self-verification and refinement are especially useful when correctness
depends on the complete proof, not merely on producing the right final answer.

\begin{wraptable}{r}{0.54\textwidth}
  \centering
  \captionsetup{type=table,hypcap=false}
  \caption{Performance on physics olympiad problems, averaged over 8 runs. Gold lines for IPhO 2024/2025 are
  20.8/19.7 points. For SU-01, $x/y$ reports scores without and with test-time scaling. 
  }
  \vspace{-0.75em}
  \label{tab:ipho-problems}
  \setlength{\tabcolsep}{3.0pt}
  \renewcommand{\arraystretch}{1.}
  \small
  \resizebox{0.85\linewidth}{!}{%
  \begin{tabular}{@{}lc@{\hspace{1.0em}}c@{}}
  \toprule
  \textbf{Model} & \textbf{IPhO 2024} & \textbf{IPhO 2025} \\
  \midrule
  \multicolumn{3}{l}{\textit{Larger models}} \\
  \midrule
  Gemini 3.1 Pro Thinking & \textbf{25.9} & \textbf{25.1} \\
  GPT-5.5-High & \underline{25.8} & \underline{23.2} \\
  DeepSeek-V3.2-Speciale & 25.1 & 21.9 \\
  \midrule
  \multicolumn{3}{l}{\textit{Similar-size models}} \\
  \midrule
  P1-30B-A3B & 23.1 & 17.7 \\
  GLM-4.7-Flash & 22.2 & 19.5 \\
  Nemotron-Cascade-2 & 21.2 & 16.7 \\
  Qwen3.6-35B-A3B & 24.3 & 19.9 \\
  Gemma-4-31B & \underline{24.4} & \underline{20.3} \\
  \rowcolor{ScienceReasoningBg!35}
  \textbf{\textcolor{iclrdeepblue}{SU-01}} & 23.5\,/\,\textbf{25.3} & \underline{20.3}\,/\,\textbf{21.7} \\
  \bottomrule
  \end{tabular}
  }
  \vspace{-1.0em}
  \end{wraptable}
  
FrontierScience-Research is a substantially harder research-oriented subset of
FrontierScience, covering physics, chemistry, and biology problems that require
scientific modeling and multi-step reasoning beyond standard contest formats.
Absolute scores remain low even for frontier systems, but SU-01 obtains the
best similar-size overall score at 11.7\%. It also leads the
similar-size block on Physics, ties for the best Chemistry score, and ranks
second on Biology, despite our RL stages using only mathematics and physics
signals. This cross-domain pattern suggests that the recipe learns a more
general scientific reasoning behavior rather than only specializing to the
training domains, providing early evidence of transferable research-level
reasoning in a compact model.

\begin{table*}[t]
\centering
\caption{Performance on non-verifiable benchmarks. FrontierScience-Research refers to the research subset of FrontierScience. For SU-01, $x/y$ reports scores without and with TTS on ProofBench.}
\label{tab:nonverifiable-benchmarks}
\vspace{-5 pt}
\setlength{\tabcolsep}{2.4pt}
\renewcommand{\arraystretch}{1.12}
\small
\resizebox{0.97\textwidth}{!}{%
\begin{tabular}{lccccccc}
\toprule
\multirow{2}{*}{\textbf{Model}} & \multicolumn{3}{c}{\textbf{IMO-ProofBench}} & \multicolumn{4}{c}{\textbf{FrontierScience-Research}} \\
\cmidrule(lr){2-4}\cmidrule(lr){5-8}
 & \textbf{Basic} & \textbf{Advanced} & \textbf{Overall} & \textbf{Physics} & \textbf{Chemistry} & \textbf{Biology} & \textbf{Overall} \\
\midrule
\multicolumn{8}{l}{\textit{Larger models}} \\
\midrule
Gemini 3.1 Pro Thinking & \underline{95.2\%} & \underline{50.0\%} & \underline{72.6\%} & 0.0\% & \underline{30.0\%} & 10.0\% & 13.3\% \\
GPT-5.5-High & \textbf{96.7\%} & \textbf{64.8\%} & \textbf{80.7\%} & \textbf{25.0\%} & \textbf{40.0\%} & \textbf{45.0\%} & \textbf{36.7\%} \\
DeepSeek-V3.2-Speciale & 77.6\% & 34.3\% & 56.0\% & \underline{10.0\%} & 20.0\% & \underline{15.0\%} & \underline{15.0\%} \\
\midrule
\multicolumn{8}{l}{\textit{Similar-size models}} \\
\midrule
P1-30B-A3B & 33.8\% & 6.2\% & 20.0\% & 0.0\% & \textbf{10.0\%} & 0.0\% & 3.3\% \\
GLM-4.7-Flash & 51.0\% & 16.7\% & 33.8\% & 0.0\% & 0.0\% & 0.0\% & 0.0\% \\
Nemotron-Cascade-2 & \underline{77.1\%} & 28.6\% & 52.9\% & \underline{5.0\%} & 5.0\% & \textbf{20.0\%} & \underline{10.0\%} \\
Qwen3.6-35B-A3B & 39.1\% & 7.1\% & 23.1\% & 0.0\% & 5.0\% & 10.0\% & 5.0\% \\
Gemma-4-31B & 46.7\% & 16.2\% & 31.4\% & 0.0\% & \textbf{10.0\%} & 5.0\% & 5.0\% \\
\rowcolor{ScienceReasoningBg!35}
\textbf{\textcolor{iclrdeepblue}{SU-01}} & \underline{77.1\%}\,/\,\textbf{91.0\%} & \underline{38.1\%}\,/\,\textbf{49.5\%} & \underline{57.6\%}\,/\,\textbf{70.2\%} & \textbf{10.0\%} & \textbf{10.0\%} & \underline{15.0\%} & \textbf{11.7\%} \\
\bottomrule
\end{tabular}%
}
\end{table*}

\subsection{Olympiad Competition Problems}


\begin{wraptable}{r}{0.58\textwidth}
  \vspace{-1.3em}
  \centering
  \captionsetup{type=table,hypcap=false}
  \caption{Performance on mathematical olympiad competition problems. Medal lines for
  IMO 2025 are 35/28/19 points, and medal lines for USAMO 2026 are 25/18/11 points.
  $^{\star}$ indicates that TTS results are evaluated by human experts, while direct generation results are evaluated automatically (\cref{app:evaluation-details}).
  }
  \vspace{-0.75em}
  \label{tab:olympiad-competition-problems}
  \setlength{\tabcolsep}{5.8pt}
  \renewcommand{\arraystretch}{1.0}
  \small
  \begin{tabular}{@{}lccccccc@{}}
  \toprule
  \multicolumn{8}{c}{\textbf{IMO 2025}} \\
  \midrule
  \textbf{Model} & \textbf{P1} & \textbf{P2} & \textbf{P3} & \textbf{P4} & \textbf{P5} & \textbf{P6} & \textbf{Total} \\
  \midrule
  \textbf{\textcolor{iclrdeepblue}{SU-01}} & 1 & 7 & 1 & 6 & 6 & 0 & 21 \\
  \textbf{\textcolor{iclrdeepblue}{SU-01} w/ TTS} & $7^{\star}$ & $7^{\star}$ & $7^{\star}$ & $7^{\star}$ & $7^{\star}$ & $0^{\star}$ & $35^{\star}$\,\goldmedalicon \\
  \midrule
  \multicolumn{8}{c}{\textbf{USAMO 2026}} \\
  \midrule
  \textbf{Model} & \textbf{P1} & \textbf{P2} & \textbf{P3} & \textbf{P4} & \textbf{P5} & \textbf{P6} & \textbf{Total} \\
  \midrule
  \textbf{\textcolor{iclrdeepblue}{SU-01}} & 7 & 0 & 0 & 7 & 0 & 1 & 15 \\
  \textbf{\textcolor{iclrdeepblue}{SU-01} w/ TTS} & $7^{\star}$ & $0^{\star}$ & $7^{\star}$ & $7^{\star}$ & $7^{\star}$ & $7^{\star}$ & $35^{\star}$\,\goldmedalicon \\
  \bottomrule
  \end{tabular}
  \end{wraptable}

Even without TTS, SU-01 averages 23.5 points on IPhO 2024 and 20.3 points on
IPhO 2025, exceeding the corresponding gold lines of 20.8 and 19.7 points, as shown in \cref{tab:ipho-problems}. TTS
further raises the scores to 25.3 and 21.7 points, making SU-01 the strongest
similar-size model in both years among models with available scores. 

\cref{tab:olympiad-competition-problems} reports the final competition-style
mathematics results. In direct generation, SU-01 reaches 21 points on IMO 2025
and 15 points on USAMO 2026, already clearing the bronze-medal lines for both
competitions. 
The direct
model obtains full credit on IMO 2025 P2 and USAMO 2026 P1/P4, near-complete
solutions on IMO 2025 P4/P5, and still fails several harder problems in a
single pass. 
This indicates that the base model has acquired substantial
olympiad reasoning ability, but still benefits from additional search and
self-correction on the most brittle proof attempts.

With test-time scaling, SU-01 reaches \textbf{35 points} on both IMO 2025 and USAMO 2026, meeting the \textbf{IMO gold} line exactly and exceeding the \textbf{USAMO gold} line by 10
points. TTS upgrades five IMO 2025 problems to full credit while P6 remains unsolved, and recovers full-credit solutions on five of six USAMO 2026 problems while
P2 remains unresolved. The USAMO 2026 score summary reports 340 competitors, a median score of 6, a top-12 cutoff of 26, and a maximum score of 35\footnote{https://web.evanchen.cc/exams/posted-usamo-statistics.pdf}. 
SU-01
therefore matches the \textbf{highest} reported human total on this contest, while still exposing a concrete failure mode on P2. 
This result suggests that our overall
recipe can elicit top-level human-like olympiad reasoning from a compact 30B-A3B model.

\paragraph{Case study.} 
We include the corresponding
model-generated solutions and expert verdicts in
\cref{app:olympiad-solutions}. 
Across the twelve IMO 2025 and USAMO 2026 problems, the model gives full-credit solutions to ten problems, with failures on IMO 2025 P6 and USAMO 2026 P2. Its main strength is translating olympiad problems into formal frameworks: coordinates or complex numbers for geometry, modular classifications for number theory, recurrences for functional equations, and automata-based dynamic programming for digit problems. A particularly striking example is USAMO 2026 P3: rather than following the standard synthetic geometry route, the model elegantly uses complex numbers to unify the unit circle, equilateral-triangle rotations, chord relations, and tangent conditions within a single algebraic framework. This yields an ingenious analytic reformulation of a configuration that olympiad solvers would typically approach through angle chasing and carefully chosen auxiliary constructions. IMO 2025 P2 shows a complementary strength: the model reduces a configuration involving two intersecting circles, an orthocenter, and a tangency claim to coordinate and distance computations. Other strong examples include the carry-state dynamic programming approach for USAMO P4 and the number-theoretic proof using totients, congruences, Vieta jumping, and Fibonacci structure in USAMO P6. However, the failures show a clear limitation: the model can miss subtle structural constraints, as in the invalid column-permutation reduction in IMO P6, or leave gaps in delicate global strategy arguments, as in USAMO P2. Overall, the model performs well when a problem admits a rigid formal representation, but is less reliable when the core challenge is preserving combinatorial structure or proving a finely tuned process invariant.
\section{Analysis and Discussion}








\subsection{Progressive Rigorous Reasoning}


\begin{wrapfigure}{r}{0.48\textwidth}
    \vspace{-70 pt}
    \centering
    \includegraphics[width=\linewidth]{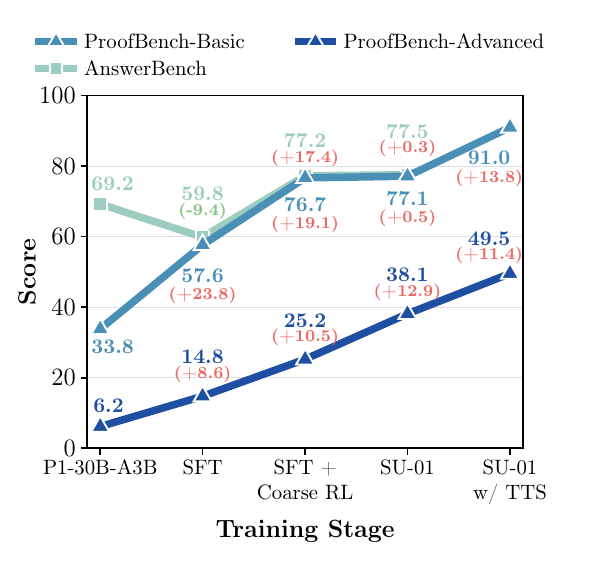}
    \vspace{-18 pt}
    \caption{{Progressive reasoning performance across training stages.} 
    }
    \label{fig:progressive-rigorous-reasoning}
    \vspace{-1.em}
\end{wrapfigure}

\cref{fig:progressive-rigorous-reasoning} separates two kinds of reasoning
progress. AnswerBench measures whether the model can recover a correct final
answer under verifiable evaluation, whereas ProofBench grades the total solution
and therefore exposes gaps in rigor, justification, and proof completion. The
starting P1-30B-A3B model is already strong on AnswerBench, but its
ProofBench scores are much lower, especially on the Advanced split, indicating that answer-seeking ability alone does not imply
olympiad-style proof reliability.

The staged trend matches the intended role of each method component. \textbf{SFT} lowers
AnswerBench from 69.2 to 59.8, but raises ProofBench-Basic from 33.8 to 57.6
and ProofBench-Advanced from 6.2 to 14.8. This is consistent with behavior
shaping: the model is moved away from short answer recovery and toward longer
proof-search, self-checking, and refinement patterns. \textbf{Coarse RL} then uses
verifiable rewards to recover and improve direct solving ability, lifting
AnswerBench to 77.2 while also improving ProofBench-Basic to 76.7 and
ProofBench-Advanced to 25.2. This suggests that RLVR scales the rigorous
reasoning behavior introduced by SFT into stronger problem-solving capability.

The final SU-01 model keeps AnswerBench essentially saturated at 77.5 and ProofBench-Basic nearly unchanged at 77.1, but improves ProofBench-Advanced from 25.2 to 38.1. The gain is therefore concentrated on harder non-verifiable proof problems, matching the role of \textbf{Refined RL}: proof-level generative rewards provide
supervision beyond final-answer correctness, self-refinement prompts train the model to critique and repair its own solutions, and experience replay keeps rare
successful hard-problem trajectories available long enough for the policy to learn more robust proof construction. \textbf{Test-time scaling} then further lifts the
proof-oriented evaluation, reaching 91.0 on ProofBench-Basic and 49.5 on ProofBench-Advanced, showing that the trained self-verification and refinement behavior
remains useful when additional inference compute is spent on checking and repairing candidate proofs.


\subsection{Characterizing Inference Scaling}

We further inspect the TTS traces from USAMO 2026 to understand where inference
compute is spent during difficult proof search. The key distinction is not only
between short and long responses, but between qualitatively different reasoning
contexts: initial generation starts from the problem statement, whereas
refinement must condition on an existing solution together with verifier
feedback or a bug report and then produce a revised proof.

\begin{wrapfigure}{r}{0.64\linewidth}
\vspace{-8pt}
\centering
\includegraphics[width=\linewidth]{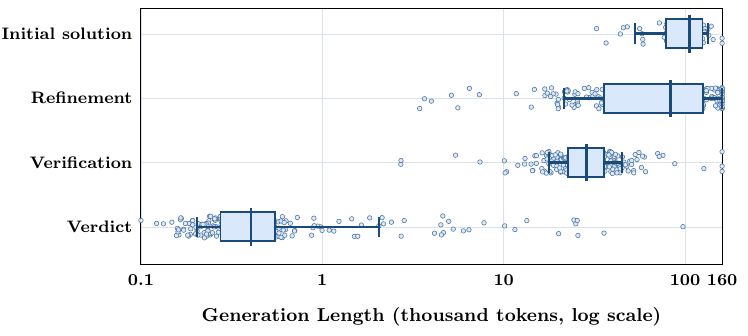}
\vspace{-15 pt}
\caption{{Generation-length distribution of actions in the TTS pipeline on USAMO 2026. }
}
\label{fig:tts-action-length}
\vspace{-6 pt}
\end{wrapfigure}

\cref{fig:tts-action-length} shows a clear allocation of computation across TTS actions. Initial solution generation is the longest stage, with a median length
of 106K tokens, reflecting broad proof search and candidate construction. Refinement remains length-intensive, with a median of 83K tokens and a heavier upper
tail, consistent with substantial proof repair. Verification is shorter but still substantive, with a median of 28.7K tokens, reflecting its role in auditing
complete arguments for hidden gaps. Verdict parsing is lightweight, with a median of only 404 tokens.

The long refinement traces indicate that the model can reason over complicated
conditioning contexts rather than merely produce long first-pass solutions:
given a candidate proof and a structured critique, it often sustains another
long reasoning trajectory to localize the flaw, preserve useful parts of the
argument, and synthesize a corrected proof. The pattern therefore suggests that
the training recipe enables the 30B model to generalize beyond direct solution
generation, sustaining complex reasoning beyond 100K tokens and repeatedly
verifying and refining its own solutions toward stronger candidates.

\subsection{Reverse-Perplexity Ordering}
\label{sec:ppl-order-analysis}

\begin{wrapfigure}{r}{0.37\textwidth}
\vspace{-55pt}
\centering
\includegraphics[width=\linewidth]{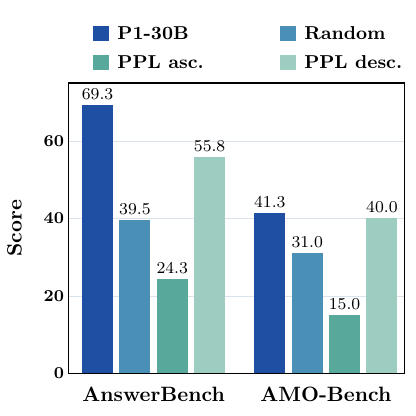}
\vspace{-16pt}
\caption{{Validation results for SFT data ordering.}}
\label{fig:sft-ppl-curriculum}
\vspace{-15pt}
\end{wrapfigure}

 We compare the effect of different SFT data orderings on validation performance, as shown in \cref{fig:sft-ppl-curriculum}. The results come from validation
  experiments whose SFT training data differ from the final training mixture used for SU-01. Data ordering has a large effect on both score recovery and generation
  stability. Random ordering substantially under-recovers the P1-30B baseline, reaching 39.5 on AnswerBench and 31.0 on AMO-Bench, with truncation rates of 7.3\%
  and 8.0\%, respectively. Descending-PPL ordering recovers much more of the original capability, reaching 55.8 on AnswerBench and 40.0 on AMO-Bench, while reducing
  truncation to 0.3\% and 0.0\%. The low-PPL-first setting is the weakest curriculum, degrading to 24.3 on AnswerBench and 15.0 on AMO-Bench. These results suggest
  that descending-PPL ordering helps preserve the capability of the post-trained model while reshaping its reasoning behavior, and also prevents training from
  falling into a superficial long-generation regime with high truncation rates.

\subsection{Cost Analysis}
\label{sec:cost_analysis}



SU-01 uses a compact and transparent post-training setup. Starting from a
30B-A3B backbone, we train on 338K SFT trajectories shorter than 8K tokens for
four epochs with batch size 128. The RL stage uses 25K prompts for 200 steps,
with batch size 128, 8 rollouts per prompt, and a 160K-token maximum response
length. 
For reference, DeepSeek-V3.2~\citep{deepseekai2025deepseekv32} reports
continued pre-training with 1,000 indexer warm-up steps over 2.1B tokens and
15,000 sparse-training steps over 943.7B tokens, followed by post-training with
specialist distillation and mixed RL over thousands of continued-RL steps; its
high-compute Speciale variant further trains on reasoning data with DeepSeekMath-V2-style proof rewards. 
Nemotron-Cascade 2~\citep{yang2026nemotroncascade2}, a same-size 30B-A3B
reference model, reports a substantially broader SFT mixture. Summing the
disclosed category counts gives roughly 26.6M SFT samples across math, proof,
code, science, long-context, chat, instruction-following, tool-use, and
software-engineering data, trained with packed 256K-token sequences for 33K
steps. Its post-training then continues with a multi-stage Cascade RL pipeline
covering IF-RL, multi-domain RL, on-policy distillation, RLHF, long-context RL,
CodeRL, and SWE RL. 
These comparisons highlight the main design point of SU-01:
a simple and unified recipe can elicit strong olympiad-level reasoning from a
compact 30B-A3B model while preserving scientific transfer.
\section{Related Work}

\paragraph{Post-training for Large Reasoning Models.}
Post-training has become the main mechanism for turning strong pretrained
language models into reliable reasoning systems. Early self-improvement work
showed that models can bootstrap rationales from their own successful attempts
\citep{zelikman2022star}, while open post-training recipes combine instruction
tuning, preference optimization, and reinforcement learning to improve general
assistant behavior \citep{lambert2024tulu}. For mathematical and long-CoT
reasoning, DeepSeekMath introduced large-scale mathematical pretraining together
with GRPO \citep{shao2024deepseekmath}, Qwen2.5-Math emphasized math-specific
self-improvement and tool-augmented data construction
\citep{yang2024qwen25math}, DeepSeek-R1 demonstrated that large-scale RL can
induce long reasoning traces and self-correction behaviors \citep{r1}, and Kimi
k1.5 highlighted long-context RL, curriculum design, and sampling-based
inference \citep{kimi2025k15}. Recent work further studies how to stabilize and
reuse learning signal: off-policy guidance and experience replay improve sample
reuse for reasoning policies \citep{yan2025learning,exgrpo}, entropy analyses
explain exploration, entropy collapse, and selective entropy regularization in
RLVR \citep{cui2025entropy,jiang2025rethinkingentropy}, and GSPO optimizes at
the sequence level for MoE reasoning models \citep{zheng2025gspo}. Contemporary
technical reports such as MiniMax-M2.5, Kimi-K2.5, and GLM-5 also show the
growing importance of agentic post-training, efficient reasoning, long-context
execution, and tool-oriented RL \citep{minimax2026m25,moonshot2026kimi25,zai2026glm5}.

\paragraph{Toward Olympiad-Level Reasoning.}
Olympiad reasoning pushes beyond benchmark math accuracy because solutions must
be complete, rigorous, and robust to hidden gaps. One line of work addresses this
challenge with specialized symbolic or neuro-symbolic systems: AlphaGeometry
solves geometry problems by combining a neural language model with symbolic
deduction \citep{trinh2024alphageometry}, and AlphaProof/AlphaGeometry 2 reached
silver-medal-level IMO performance through formal reasoning and search
\citep{deepmind2024alphaproof}. More recent frontier systems have moved toward
broader natural-language reasoning and test-time search, including Gemini Deep
Think's gold-medal-level IMO result \citep{deepmind2025geminiimo}. In parallel,
model-agnostic verification-and-refinement pipelines show that repeated
generation, critique, repair, and acceptance decisions can substantially improve
proof quality without relying on a single-pass answer \citep{huang2025winning},
and DeepSeekMath-V2 studies self-verifiable mathematical reasoning as a training
and inference target \citep{deepseekmathv2}. Nemotron-Cascade 2 provides another
recent example of a compact MoE reasoning model approaching frontier mathematical
and olympiad performance through cascade RL and multi-domain on-policy
distillation \citep{yang2026nemotroncascade2}. Our contribution is a simple and
unified recipe that enables a 30B-A3B model to develop rigorous proof behavior
through post-training and reach olympiad-level performance with
self-verification, refinement, and test-time scaling.

\section{Conclusion}

This report presents a simple and unified recipe for turning a compact
post-trained reasoning model into a stronger mathematical and scientific
reasoner. Starting from a broadly capable 30B-A3B backbone, SU-01 combines
reverse-perplexity curriculum SFT, efficient coarse RL with outcome verification,
refined RL with proof-level rewards, self-refinement and experience replay, and
test-time scaling through self-verification and refinement. Together, these
stages decompose rigorous reasoning improvement into behavior shaping, scalable
reward feedback, proof-level specialization, and inference-time repair. 
The resulting model reaches gold-medal-level performance on mathematical and
physical olympiad competitions, sustains reasoning trajectories beyond 100K
tokens during inference, and shows transfer to scientific domains beyond the
main math and physics training signals. 
In summary, SU-01 supports a specializable-generalist view of compact reasoning models: with the right training and inference recipe, a broadly capable backbone can be driven toward expert-level proof reasoning while retaining meaningful scientific transfer.

\section*{Acknowledgments}
This work was supported by the Shanghai Artificial Intelligence Laboratory. 
We thank the authors and maintainers of prior open research and infrastructure
that made this work possible. 
In particular, we are
grateful to DeepSeek for open-sourcing strong reasoning policies and generative reward models, which provided an important
reference point for our work~\citep{deepseekai2025deepseekv32,deepseekmathv2}.
IMO-Bench, AMO-Bench, and FrontierScience
helped guide the overall system optimization by offering challenging
mathematical and scientific reasoning benchmarks and evaluation
protocols~\citep{luong2025imobench,an2025amobench,wang2026frontierscience}. We
also thank prior data efforts that supported our SFT and RL data curation,
including DeepMath, NaturalReasoning, Eurus, OpenCodeReasoning, P1, and
OPC~\citep{he2025deepmath,yuan2025naturalreasoning,cui2025process,ahmad2025opencodereasoning,chen2025p1,dekoninck2025opc},
as well as the many public problem sources and communities that cannot all be
listed here. We further acknowledge the broader open-source infrastructure
ecosystem, including slime for training and SGLang for efficient inference and
serving~\citep{thudm2026slime,sglang2026repo}.

\bibliography{iclr2026_conference}
\bibliographystyle{iclr2026_conference}

\appendix
\clearpage

\section*{Appendix}
\addcontentsline{toc}{part}{Appendix}  

\startcontents[appendix]
\printcontents[appendix]{}{1}{\subsection*{Contents}}

\section{Implementation and Evaluation Details}
\label{app:training-details}

This appendix collects the implementation details needed to interpret and
reproduce the reported training, inference, reward-model serving, and evaluation
procedures. We summarize the effective modeling, optimization, decoding,
serving, and grading settings extracted from the corresponding launch scripts
and evaluation protocols, while omitting infrastructure-only commands that do
not affect the training objective, data semantics, inference behavior, reward
definition, or evaluation criteria.

\section{Problem-Solving Prompt}
\label{app:problem-solving-prompt}

Unless otherwise specified, all RL training and inference stages use the following fixed problem-solving prompt for SU-01:

\begin{mdframed}[style=mdpurplebox,frametitle={Problem-Solving Prompt}]
\small
\ttfamily
Please solve the following olympiad problem. Show your complete reasoning and proof.\\[2pt]
1. Please use LaTeX format to represent the variables and formulas used in the solution process and results.\\
2. If the problem asks you to find specific values, please put the final answer(s) in \textbackslash boxed\{\}.\\
3. If the problem requires a proof, present a clear and rigorous argument.
\end{mdframed}

\section{SFT Training Details}
\label{app:sft-training-details}

The SFT stage is implemented with slime~\citep{thudm2026slime}, trained on 8 GPUs, initializes from P1-30B, and optimizes the reverse-perplexity-ordered mixture described in \cref{sec:sft-data-curation}; rollout shuffling is disabled so that the curriculum order is preserved throughout training. We train for four epochs with batch size 128, using Adam with learning rate $1\times 10^{-5}$, cosine decay to a minimum learning rate of $1\times 10^{-6}$, warmup fraction 0.1, weight decay 0.1, and momentum parameters $\beta_1=0.9$ and $\beta_2=0.95$. This configuration treats SFT as a controlled behavioral adaptation step: the objective is not to maximize benchmark performance directly, but to expose the post-trained backbone to a stable sequence of rigorous long-form trajectories while preserving as much of its existing mathematical and scientific competence as possible.

\section{RL Training Details}
\label{app:rl-training-details}

The RL stage is implemented with slime~\citep{thudm2026slime}, trained on 64 GPUs, continues from the SFT checkpoint, and trains with a GSPO objective over a balanced mixture of verifiable prompts, proof-reward prompts, self-refinement prompts, and replayed experience. In total, the model is trained for 200 RL steps, consisting of 96 coarse-RL steps followed by 104 refined-RL steps. Each rollout uses prompt batch size 128, 8 samples per prompt, maximum response length 160k tokens, temperature 1.0, and 4 policy-update steps per rollout. The data pipeline applies dynamic sampling with a non-zero standard-deviation reward requirement, partial rollout, and oversampling batch size 160; the oversampling and partial-rollout mechanism increases the candidate prompt pool under filtering while recycling completed partial generations, and replay filtering prevents repeatedly training on queries that have become too easy for the current policy. The policy objective uses the GSPO advantage estimator with KL coefficient 0, entropy coefficient 0, symmetric clip range $10^{-3}$, and trajectory importance sampling enabled. 

The self-refinement mechanism uses a refinement prompt template that includes the original problem and the previous failed answer. The self-refinement ratio is set to $20\%$, which denotes the fraction of training queries sampled from refinement data, and the group reward threshold selects failed or partially failed rollout groups with an average reward below $0.5$ for refinement. Experience replay largely follows the ExGRPO setting without reward shaping and is re-implemented in the \texttt{slime} training framework. We use an experience admission threshold of online rollout correctness below $25\%$, one replayed trajectory per replay query, and entropy-based selection estimated from SGLang top-$16$ log probabilities. A query is admitted to the experience pool if only one sampled rollout succeeds, making the successful trajectory a hard positive example. Experiences are retired once online rollout correctness reaches $50\%$, indicating improved model performance on the corresponding query. Optimization is performed using Adam with a constant learning rate of $1\times 10^{-6}$, weight decay of 0.1, and momentum parameters $\beta_1=0.9$ and $\beta_2=0.98$.

\section{Inference and Reward-Model Serving Details}
\label{app:inference-serving-details}


For test-time scaling, inference is served with SGLang~\citep{sglang2026repo} by default; the Nemotron-Cascade-2 and DeepSeek-V3.2-Speciale models are served with vLLM~\citep{vllm2026repo} \footnote{Nemotron-Cascade-2:vLLM-0.17.2rc1.dev148+g47b7af0d8.cu128 ; DeepSeek-V3.2-Speciale: vLLM-v0.20-CUDA12.9.}. We follow the same model-agnostic verification-and-refinement setting as \citet{huang2025winning}. 

Concretely, inference is organized as an iterative solve--verify--refine procedure: the solver first produces a candidate proof, the candidate is inspected by a verifier that returns a structured critique or bug report, and the solver then revises the proof conditioned on this feedback until the candidate is accepted or the refinement budget is exhausted. 
Unless otherwise specified, each generation uses a maximum length of 160,000 tokens, temperature 1.0, and top-p 0.95. We use the following exceptions for model- or benchmark-specific constraints. For Nemotron model, we set top-p to 1.0. The maximum generation length is set to 131K tokens for AIME25/26, AMOBench, IPhO, and FrontierScience-Olympic, and to 256K tokens for IMO-AnswerBench, IMO-ProofBench, and FrontierScience-Research subset. 
For API-limited models, we set the maximum decoding completion length to the largest value allowed by the API request constraints: 128,000 tokens for GPT-5.5 API and 65,535 tokens for Gemini 3.1-Pro.

The TTS loop uses explicit stopping rules following the verification-and-refinement protocol. Within a single run, \texttt{MAX\_VERIFICATION\_TRUE\_ROUNDS=5} means that a candidate is accepted once it passes verification for five consecutive rounds, while \texttt{MAX\_VERIFICATION\_FALSE\_ROUNDS=10} terminates the run early after ten consecutive failed verification rounds. Each run allows at most \texttt{MAX\_EXPLORATION\_ROUNDS=30} solve--verify--refine cycles, corresponding to the repeated exploration and correction loop in the TTS pipeline. For each problem, we set \texttt{MAX\_RUNS=10}, so at most ten independent runs are launched. This setting is intentionally aligned with the self-refinement behavior used during refined RL, so the model is evaluated under the same type of proof-repair workflow that it sees during training.

For proof-reward evaluation during refined RL, the DeepSeekMath-V2 reward server is deployed on 32 GPUs with maximum context length 32k and data-parallel degree 64. 
Speculative decoding is enabled during server-side inference using an MTP-based configuration. We adopt a $3$-step speculative decoding scheme with a \textit{single} top-ranked EAGLE draft candidate and $4$ draft tokens per step. This setting improves verifier efficiency while maintaining the evaluation correctness required by the refined-RL training.

\section{Compared Models}
\label{app:compared-models}

The comparison tables use a mixture of public technical reports, official model
cards, and official benchmark reports. For the answer-verifiable comparisons in
\cref{tab:verifiable-single-pass}, the larger-model group contains
DeepSeek-V3.2~\citep{deepseekai2025deepseekv32}, GPT-5.5~\citep{openai2026gpt55},
and Gemini 3.1 Pro Thinking~\citep{googledeepmind2026gemini31pro}. The
similar-size group contains GLM-4.7-Flash~\citep{zai2026glm47flash},
Nemotron-Cascade-2-30B-A3B~\citep{yang2026nemotroncascade2},
Qwen3.6-35B-A3B~\citep{qwen2026qwen36_35b}, and
Gemma-4-31B~\citep{google2026gemma4_31b}, together with our
\textbf{SU-01}. These same active comparison families are reused in the
non-verifiable benchmark table in \cref{tab:nonverifiable-benchmarks}, with
missing benchmark entries marked by --. GPT-5.5-High is treated as the
high-reasoning variant of GPT-5.5, Gemini 3.1 Pro Thinking is drawn from the
Gemini 3.1 Pro family, and DeepSeek V3.2 uses the same DeepSeek-V3.2 source as
above. When no paper-style technical report is available, we cite the
corresponding official model card, repository, or product page.

\section{Evaluation Details}
\label{app:evaluation-details}

For answer-verifiable tasks, we use a layered automatic grading pipeline. We first apply rule-based verification, including canonicalized answer matching and symbolic or expression-level checks with Math-Verify\footnote{Math-Verify repository: \url{https://github.com/huggingface/Math-Verify}.}. If a response is not resolved by these checks, we send it to gpt-oss-120b for generative verification~\citep{openai2025gptoss}. For FrontierScience-O, we use the same verifiable grading pipeline, but adopt the dedicated prompt described in the original FrontierScience paper~\citep{wang2026frontierscience}.

For non-verifiable and proof-oriented tasks, we follow each benchmark's official evaluation protocol. ProofBench is evaluated following the IMO-Bench framework~\citep{luong2025imobench}: solutions are graded with the four-level score set \(\{0,1,6,7\}\), and we use Gemini-2.5-Pro as the grading model to align with the official reported setting. FrontierScience-R is evaluated with the official FrontierScience procedure using GPT-5-high as the judge~\citep{wang2026frontierscience}.

For IPhO, we follow the evaluation paradigms used in P1 and P1-VL~\citep{chen2025p1,luo2026p1vl}, which combine physics-specific grading prompts with competition-style scoring criteria. For IMO and USAMO, we construct a ProofBench-style automatic grading framework from annotated reference solutions and grading guidelines prepared by expert annotators who have won gold medals in mathematical olympiads, such as IMO or CMO. For the final verdict, three gold-medal experts independently assign scores from \(\{0,1,6,7\}\), and we report the worst of the three scores as a conservative final score.

\clearpage
\section{Model Solutions for IMO 2025 and USAMO 2026}
\label{app:olympiad-solutions}
This section lists the model-generated solutions used for the human-expert scoring reported in \cref{tab:olympiad-competition-problems}. Each solution is followed by its final expert verdict. For USAMO 2026 Problem 2, we show only an excerpt because the original response is substantially longer than the other solutions.

\begingroup
\footnotesize
\sloppy
\setlength{\parindent}{0pt}
\setlength{\parskip}{0.08em}
\setlength{\abovedisplayskip}{3pt}
\setlength{\belowdisplayskip}{3pt}
\setlength{\abovedisplayshortskip}{2pt}
\setlength{\belowdisplayshortskip}{2pt}
\newcommand{\solutionheading}[1]{\par\smallskip\noindent\textbf{#1}\par}
\newcommand{\compactbullet}[1]{\par\noindent$\triangleright$~#1\par}
\newcommand{\compactnumber}[2]{\par\noindent\textbf{#1.}~#2\par}

\subsection{IMO 2025}
\vspace{1em}
\paragraph{IMO 2025 Problem 1.}
\begin{mdframed}[style=mdpurplebox,frametitle={IMO 2025 Problem 1}]
A line in the plane is called \emph{sunny}
if it is not parallel to any of the $x$-axis, the $y$-axis, or the line $x+y=0$.

Let $n \ge 3$ be a given integer.
Determine all nonnegative integers $k$ such that there exist $n$ distinct lines
in the plane satisfying both of the following:
\begin{itemize}
\ii for all positive integers $a$ and $b$ with $a+b\le n+1$,
  the point $(a,b)$ lies on at least one of the lines; and
\ii exactly $k$ of the $n$ lines are sunny.
\end{itemize}
\end{mdframed}
\solutionheading{Solution}
\begin{remark*}
  7/7.
\end{remark*}
\textbf{Definition.} For an integer \(n\ge 3\) let\par
\[
P_n = \{(a,b)\in\mathbb{N}^2 \mid a\ge 1,\; b\ge 1,\; a+b\le n+1\}.
\]
We are asked for which nonnegative integers \(k\) there exist \(n\) distinct lines in the plane such that\par
\compactbullet{every point of \(P_n\) lies on at least one of the lines, and}
\compactbullet{exactly \(k\) of the lines are \emph{sunny} (i.e. not parallel to the \(x\)-axis, the \(y\)-axis, or the line \(x+y=0\)).}
We will prove that for every \(n\ge 3\) the only possible values of \(k\) are \(0,1,3\).\par
\smallskip
\solutionheading{1. Constructions for \(k = 0,1,3\)}
We give explicit families of \(n\) distinct lines that satisfy the two conditions.\par
\solutionheading{\(k = 0\)}
Take the \(n\) horizontal lines\par
\[
y = 1,\; y = 2,\; \dots,\; y = n.
\]
Each point \((a,b)\in P_n\) lies on the line \(y = b\). No line is sunny (all are parallel to the \(x\)-axis). Hence \(k = 0\) is attainable.\par
\solutionheading{\(k = 1\)}
Take the \(n-1\) horizontal lines\par
\[
y = 1,\; y = 2,\; \dots,\; y = n-1.
\]
These cover all points with \(b \le n-1\). The only point not covered is \((1,n)\) (since \(a+n\le n+1\) forces \(a = 1\)). Add one extra line, for instance\par
\[
L:\; y = x + (n-1).
\]
\(L\) has slope \(1\), so it is sunny. It passes through \((1,n)\). The total number of lines is \((n-1)+1 = n\), and exactly one line is sunny. Thus \(k = 1\) is attainable.\par
\solutionheading{\(k = 3\)}
\emph{If \(n = 3\):}Use the three sunny lines\par
\[
\begin{aligned}
L_1 &: y = x,\\
L_2 &: y = -2x + 5,\\
L_3 &: y = -\tfrac12 x + \tfrac52.
\end{aligned}
\]
Check: \(L_1\) contains \((1,1)\) and \((2,2)\); \(L_2\) contains \((1,3)\) and \((2,1)\); \(L_3\) contains \((1,2)\) and \((3,1)\). The six points of \(P_3\) are all covered. All three lines are sunny, so \(k = 3\).\par
\emph{If \(n > 3\):}First take the \(n-3\) horizontal lines\par
\[
y = 1,\; y = 2,\; \dots,\; y = n-3.
\]
These cover all points with \(b \le n-3\). The remaining points are those with \(b = n-2,\, n-1,\, n\); there are exactly \(3+2+1 = 6\) such points. Translate the situation by \((0,\,-(n-3))\): the map\par
\[
(x,\,y) \;\longmapsto\; (x,\,y-(n-3))
\]
sends these six points bijectively onto the set \(P_3\) (because \(a+b \le n+1\) becomes \(a + (b'+(n-3)) \le n+1 \iff a+b' \le 4\)). Therefore, if we take the three sunny lines that cover \(P_3\) in the translated coordinates and then translate them back, we obtain three sunny lines in the original plane:\par
\[
\begin{aligned}
S_1 &: y = x + (n-3),\\
S_2 &: y = -2x + (n+2),\\
S_3 &: y = -\tfrac12 x + \tfrac{2n-1}{2}.
\end{aligned}
\]
Together with the \(n-3\) horizontal lines we have \(n\) distinct lines, exactly three of which are sunny. Hence \(k = 3\) is attainable for every \(n\ge 3\).\par
\smallskip
\solutionheading{2. Upper bound on points per sunny line}
\textbf{Lemma 1.} For any sunny line \(L\), we have \(|L \cap P_n| \le \left\lfloor\dfrac{n+1}{2}\right\rfloor\).\par
\emph{Proof.} If \(L\) contains no integer point, then obviously \(|L\cap P_n|\le 1\le\lfloor(n+1)/2\rfloor\) (since \(n\ge 3\)). So assume \(L\) contains at least one integer point. Since \(L\) is sunny, it is not vertical, so we can write its equation as \(y = mx + d\).\par
\compactbullet{If \(m\) is irrational, then \(L\) can contain at most one integer point (two integer points would give a rational slope). Hence the bound holds. Thus we may assume \(m\) is rational.}
Write \(m = p/q\) in lowest terms, \(q>0\), \(\gcd(p,q)=1\). Choose an integer point \((x_0,y_0)\) on \(L\). Then all integer points on \(L\) are\par
\[
(x,y) = (x_0 + q t,\; y_0 + p t),\qquad t\in\mathbb Z.
\]
Let the set of \(t\) for which the point lies in \(P_n\) be a consecutive integer interval \(t_{\min},\dots,t_{\max}\); let \(N\) be the number of such \(t\). Then \(t_{\max}-t_{\min} = N-1\).\par
For any such point we have the constraints\par
\[
1 \le x \le n,\qquad 1 \le y \le n,\qquad 2 \le x+y \le n+1.
\]
In particular,\par
\begin{align*}
q\,(N-1) &= |x_{\max}-x_{\min}| \le n-1, \quad\text{(A)}\\
|p|\,(N-1) &= |y_{\max}-y_{\min}| \le n-1, \quad\text{(B)}\\
|p+q|\,(N-1) &= |(x+y)_{\max}-(x+y)_{\min}| \le n-1. \quad\text{(C)}
\end{align*}
Set \(M = \max\{q,\ |p|,\ |p+q|\}\). Then (A)-(C) give\par
\[
N-1 \le \frac{n-1}{M}.
\]
Now, because \(L\) is sunny, we have \(p \neq 0\) (otherwise horizontal) and \(p+q \neq 0\) (otherwise slope \(-1\)). We claim that \(M \ge 2\). Suppose \(M = 1\). Then \(q = 1\) (since \(q>0\)), \(|p| \le 1\), \(|p+q| \le 1\). As \(p \neq 0\), either \(p = 1\) or \(p = -1\).\par
\compactbullet{If \(p = 1\), then \(|p+q| = |1+1| = 2 > 1\), contradiction.}
\compactbullet{If \(p = -1\), then \(p+q = 0\), which is forbidden.}
Hence \(M \ge 2\). Consequently,\par
\[
N-1 \le \frac{n-1}{2}\quad\Longrightarrow\quad N \le \frac{n+1}{2}.
\]
Since \(N\) is an integer, \(N \le \left\lfloor\frac{n+1}{2}\right\rfloor\). \(\square\)\par
\smallskip
\solutionheading{3. No covering by all sunny lines for \(n \ge 4\)}
\textbf{Lemma 2.} For \(n \ge 4\) there is no set of \(n\) distinct sunny lines that covers \(P_n\).\par
\emph{Proof.}\par
\emph{Even \(n\):} Write \(n = 2m\). By Lemma 1 each sunny line contains at most \(m\) points. Therefore the total number of incidences (counting each point as many times as lines through it) is at most \(n \cdot m = 2m^2\). But\par
\[
|P_n| = \frac{n(n+1)}{2} = m(2m+1) = 2m^2 + m > 2m^2 \qquad (m\ge1),
\]
so it is impossible to cover all points.\par
\emph{Odd \(n\):} Write \(n = 2m+1\) with \(m \ge 2\) (i.e. \(n \ge 5\)). Lemma 1 gives each sunny line at most \(m+1\) points. Hence\par
\[
\sum_{i=1}^{n} |L_i \cap P_n| \le n (m+1) = (2m+1)(m+1).
\]
But\par
\[
|P_n| = \frac{(2m+1)(2m+2)}{2} = (2m+1)(m+1).
\]
Thus the inequality must be an equality. In particular, each line contains exactly \(m+1\) points, and the sum of the sizes equals \(|P_n|\). Since every point is covered at least once, equality forces each point to be covered exactly once, and therefore the \(n\) lines are pairwise disjoint.\par
We now classify all sunny lines that can contain exactly \(m+1\) points in \(P_n\).\par
\textbf{Lemma 3.} Let \(n = 2m+1 \ge 3\). If a sunny line \(L\) satisfies \(|L \cap P_n| = m+1\), then \(L\) is one of the three lines\par
\[
L_A: y = x,\qquad L_B: y = -2x + (n+2),\qquad L_C: y = -\tfrac12 x + \tfrac{n+2}{2}.
\]
\emph{Proof.} Write \(L\) in reduced form \(y = \frac{p}{q}x + d\) with \(q>0\), \(\gcd(p,q)=1\), \(p \neq 0\), \(p+q \neq 0\). Let \(N = m+1\).\par
From (A)-(C) we have (with \(N-1 = m\) and \(n-1 = 2m\))\par
\[
q\,m \le 2m,\quad |p|\,m \le 2m,\quad |p+q|\,m \le 2m,
\]
hence\par
\[
q \le 2,\quad |p| \le 2,\quad |p+q| \le 2.
\]
We examine the possibilities.\par
\textbf{Case }\(q = 1\).\textbf{ Then }\(|p| \le 2\) and \(|p+1| \le 2\). Since \(p \neq 0\) and \(p+1 \neq 0\) (i.e. \(p \neq -1\)), the admissible integers are \(p = 1\) and \(p = -2\).\par
\textbf{Case }\(q = 2\).\textbf{ Then }\(\gcd(p,2)=1\), so \(p\) is odd. \(|p| \le 2\) forces \(p = \pm 1\). \(|p+2| \le 2\) eliminates \(p = 1\) because \(|3| > 2\); thus only \(p = -1\) remains.\par
Now we determine the intercept \(d\) so that exactly \(m+1\) integer points of \(P_n\) lie on the line.\par
\smallskip
\textbf{Subcase (1, 1):} \(L: y = x + d\).\par
Let the integer points be \((a+i,\ a+i+d)\) for \(i = 0,1,\dots,m\) (since \(x\) increases by \(q=1\)). The \(x\)-coordinates range from \(a\) to \(a+m\) and must satisfy \(1 \le a+i \le 2m+1\). Hence \(1 \le a \le m+1\). The constraints from the bounds on \(y\) and on \(x+y\) yield:\par
\[
\begin{cases}
a+d \ge 1,\\
a+m+d \le 2m+1,\\
2a+2m+d \le 2m+2.
\end{cases}
\]
The third inequality gives \(d \le 2-2a\); the second gives \(d \le m+1 - a\); the first gives \(d \ge 1-a\).\par
For consistency we need \(1-a \le 2-2a\), i.e. \(a \le 1\). Thus \(a = 1\). Then the bounds become \(0 \le d \le \min\{m,\,0\} = 0\), so \(d = 0\). Hence \(L: y = x\).\par
\smallskip
\textbf{Subcase (1, -2):} \(L: y = -2x + d\).\par
Again the points are \((a+i,\ -2(a+i)+d)\) with \(i=0,\dots,m\) and \(1 \le a \le m+1\). The tightest constraints come from \(y \ge 1\) (at \(i=m\)), \(y \le 2m+1\) (at \(i=0\)), and \(x+y \le 2m+2\) (at \(i=0\)):\par
\[
\begin{aligned}
-2(a+m)+d \ge 1 \;&\Longrightarrow\; d \ge 1 + 2a + 2m,\\
-2a + d \le 2m+1 \;&\Longrightarrow\; d \le 2m+1 + 2a,\\
-a + d \le 2m+2 \;&\Longrightarrow\; d \le 2m+2 + a.
\end{aligned}
\]
The first two give\par
\[
1+2a+2m \le d \le 2m+1+2a.
\]
Hence \(d\) must equal \(2m+1+2a\) (the only integer satisfying both). Substituting into the third inequality yields\par
\[
2m+1+2a \le 2m+2 + a \;\Longrightarrow\; a \le 1.
\]
Since \(a \ge 1\), we have \(a = 1\). Then \(d = 2m+1+2 = 2m+3 = n+2\). So \(L: y = -2x + (n+2)\).\par
\smallskip
\textbf{Subcase (2, -1):} \(L: y = -\tfrac12 x + d\).\par
The \(x\)-coordinates differ by \(q=2\). Let the smallest be \(a\). Then the points are \((a+2i,\ -\tfrac12(a+2i)+d)\) for \(i=0,\dots,m\). To stay within the bounds \(1 \le x \le 2m+1\) we need\par
\[
a \ge 1 \quad\text{and}\quad a+2m \le 2m+1 \;\Longrightarrow\; a \le 1,
\]
so \(a = 1\). Thus the \(x\)-coordinates are \(1,3,5,\dots,2m+1\).\par
For the \(y\)-coordinates to be integers, \(d\) must have the form \(d = d' + \tfrac12\) with \(d' \in \mathbb Z\). Then\par
\[
y_i = -\tfrac12(2i+1) + d = -i - \tfrac12 + d' + \tfrac12 = d' - i.
\]
Now apply the constraints:\par
\compactbullet{\(y_i \ge 1\) for all \(i\): the smallest \(y\) occurs at \(i=m\) (since slope negative), so \(d' - m \ge 1 \;\Rightarrow\; d' \ge m+1\).}
\compactbullet{\(y_i \le 2m+1\) for all \(i\): the largest \(y\) occurs at \(i=0\), giving \(d' \le 2m+1\).}
\compactbullet{\(x_i + y_i = (2i+1) + (d' - i) = d' + i + 1 \le 2m+2\): the most restrictive is at \(i=m\), giving \(d' + m + 1 \le 2m+2 \;\Rightarrow\; d' \le m+1\).}
Thus \(d' \ge m+1\) and \(d' \le m+1\), so \(d' = m+1\). Then \(d = m+1 + \tfrac12 = \tfrac{2m+3}{2} = \tfrac{n+2}{2}\). Hence \(L: y = -\tfrac12 x + \tfrac{n+2}{2}\).\par
\smallskip
We have exhausted all possibilities; therefore any sunny line with exactly \(m+1\) points must be one of the three listed. \(\square\)\par
Returning to the proof of Lemma 2 for odd \(n\ge 5\): we have \(n\) distinct sunny lines, each must contain exactly \(m+1\) points, hence each must be one of \(L_A, L_B, L_C\). But there are only three distinct such lines, while \(n \ge 5\) gives a contradiction. Hence no covering exists for odd \(n\ge 5\) either. (For \(n=3\) the three lines are exactly these three, and that case is covered by the construction.) This completes the proof of Lemma 2. \(\square\)\par
\smallskip
\solutionheading{4. Induction on \(n\)}
We now prove by strong induction that for every \(n\ge 3\) the number \(k\) of sunny lines in any covering of \(P_n\) by \(n\) distinct lines can only be \(0,1,3\).\par
\textbf{Base case }\(n = 3\).\textbf{ Constructions in 1 show that }\(k = 0,1,3\) are attainable. We must show that \(k = 2\) is impossible.\par
Assume three lines cover \(P_3\) with exactly two sunny. Let the sunny lines be \(L_1, L_2\) and the third line \(L_3\) be non-sunny.\par
Lemma 1 implies that each sunny line contains at most two points of \(P_3\). The only pairs of points in \(P_3\) that are collinear with a sunny line (i.e. with slope different from \(0\), undefined, or \(-1\)) are exactly the three pairs:\par
\[
\{(1,1),(2,2)\}\ (\text{slope }1),\quad \{(1,3),(2,1)\}\ (\text{slope }-2),\quad \{(1,2),(3,1)\}\ (\text{slope }-1/2).
\]
Hence any sunny line that contains two points must be one of the three lines\par
\[
\ell_1: y=x,\quad \ell_2: y=-2x+5,\quad \ell_3: y=-\tfrac12 x+\tfrac52.
\]
Now consider the possibilities.\par
\emph{If both \(L_1\) and \(L_2\) contain two points and are disjoint.} Then they must be two of the three lines \(\ell_1,\ell_2,\ell_3\). Those three lines are pairwise disjoint, so they cover four points. The two remaining points lie on the third \(\ell\), which is sunny. The third line \(L_3\) is non-sunny, so it cannot cover both of those points (the unique line through them is sunny). Contradiction.\par
\emph{If at least one of \(L_1, L_2\) contains at most one point.} Then the two sunny lines together cover at most three points. Consequently, the non-sunny line \(L_3\) must cover at least three points. The only lines that contain three points of \(P_3\) are the three "boundary" lines:\par
\[
y=1,\quad x=1,\quad x+y=4.
\]
(One verifies that any other line contains at most two points.) So \(L_3\) is one of these three.\par
\compactbullet{\textbf{}\(L_3 = y=1\)\textbf{: covers }\((1,1),(2,1),(3,1)\). The uncovered points are \((1,2),(1,3),(2,2)\). Each of these three points lies on a different one of \(\ell_1,\ell_2,\ell_3\) (namely \((2,2)\in\ell_1\), \((1,3)\in\ell_2\), \((1,2)\in\ell_3\)). Two sunny lines can cover at most two of them (each \(\ell_i\) contains exactly one). Hence impossible.}
\compactbullet{\textbf{}\(L_3 = x=1\)\textbf{: covers }\((1,1),(1,2),(1,3)\). Uncovered: \((2,1),(2,2),(3,1)\). Again these are one per \(\ell_i\); two sunny lines cannot cover all three. Contradiction.}
\compactbullet{\textbf{}\(L_3 = x+y=4\)\textbf{: covers }\((1,3),(2,2),(3,1)\). Uncovered: \((1,1),(1,2),(2,1)\), again one per \(\ell_i\). Contradiction.}
Thus \(k = 2\) is impossible, so the only possible values for \(n=3\) are \(0,1,3\).\par
\textbf{Inductive step.} Assume the statement holds for all \(m\) with \(3 \le m < n\), where \(n \ge 4\). Consider any covering of \(P_n\) by \(n\) distinct lines.\par
Define three special lines:\par
\[
R:\ y=1,\qquad C:\ x=1,\qquad D:\ x+y = n+1.
\]
Note that \(R, C, D\) are not sunny (horizontal, vertical, slope \(-1\) respectively).\par
\textbf{Claim.} At least one of \(R, C, D\) belongs to the covering.\par
\emph{Proof of claim.} Suppose none of them is present. Then:\par
\compactbullet{The line \(y=1\) contains \(n\) distinct points \((a,1)\). To cover a point \((a,1)\), the line must intersect \(y=1\) at that point. A horizontal line different from \(y=1\) does not intersect \(y=1\). Hence every line that covers a point on \(y=1\) must be non-horizontal. Moreover, a non-horizontal line meets \(y=1\) in at most one point. Therefore we need at least \(n\) non-horizontal lines; since there are exactly \(n\) lines, \emph{all} lines are non-horizontal, and each covers exactly one point on \(y=1\).}
\compactbullet{The line \(x=1\) contains \(n\) points \((1,b)\). Since \(C\) is absent, a vertical line different from \(x=1\) does not contain any point with \(x=1\). Hence every line covering a point on \(x=1\) must be non-vertical. Since all lines are already non-horizontal, they must all be non-vertical to cover those points (a vertical line would miss \(x=1\) unless it is \(x=1\) itself). Thus all lines are non-vertical.}
\compactbullet{The line \(x+y=n+1\) contains \(n\) points. Since \(D\) is absent, a line with slope \(-1\) other than \(D\) is parallel to it and does not intersect it. Therefore every line covering a point on \(D\) must have slope different from \(-1\). Hence all lines have slope \(\neq -1\).}
Therefore every line is non-horizontal, non-vertical, and slope \(\neq -1\); i.e., every line is sunny. But Lemma 2 states that for \(n\ge 4\) there is no covering of \(P_n\) by \(n\) sunny lines. Contradiction. Hence the claim holds. \(\square\)\par
Now we examine which of \(R, C, D\) is present. (If more than one is present, we may choose any one; the argument works in each case.)\par
\emph{If \(R\) is present:} Remove \(R\) from the set. The remaining \(n-1\) lines still cover all points of \(P_n\) with \(y\ge 2\) (because \(R\) only covered points with \(y=1\)). Apply the translation \(\phi_R: (x,y) \mapsto (x, y-1)\). This map sends the set \(\{(x,y)\in P_n \mid y\ge 2\}\) bijectively onto \(P_{n-1}\). Moreover, \(\phi_R\) sends each line to a line (translations preserve lines). The resulting family consists of \(n-1\) distinct lines (translation preserves distinctness) and covers \(P_{n-1}\) (each point of \(P_{n-1}\) corresponds to a point with \(y\ge 2\) that was covered). Since \(R\) is not sunny, the number of sunny lines among the \(n-1\) lines remains \(k\).\par
Thus we have a covering of \(P_{n-1}\) by \(n-1\) distinct lines with exactly \(k\) sunny lines. By the induction hypothesis (applied to \(m=n-1\)), we must have \(k \in \{0,1,3\}\).\par
\emph{If \(C\) is present:} Remove \(C\) and use \(\phi_C: (x,y) \mapsto (x-1, y)\). It maps \(\{(x,y)\in P_n \mid x\ge 2\}\) onto \(P_{n-1}\). \(C\) is non-sunny, so \(k\) unchanged. Induction gives \(k \in \{0,1,3\}\).\par
\emph{If \(D\) is present:} Remove \(D\). The remaining \(n-1\) lines cover all points of \(P_n\) that are not on \(D\). Observe that \(P_{n-1} \subseteq P_n\) and \(P_{n-1} \cap D = \emptyset\) (since points in \(P_{n-1}\) satisfy \(a+b\le n\), while \(D\) requires \(a+b = n+1\). Hence they cover \(P_{n-1}\). \(D\) is not sunny, so \(k\) stays the same. Induction yields \(k \in \{0,1,3\}\).\par
In every possible case we conclude \(k \in \{0,1,3\}\), which completes the inductive step.\par
By strong induction, the statement holds for all integers \(n \ge 3\).\par
\smallskip
\solutionheading{5. Conclusion}
We have shown that for every \(n \ge 3\) the only possible numbers of sunny lines in a covering of \(P_n\) by \(n\) distinct lines are \(0\), \(1\), and \(3\). Explicit constructions demonstrate that each of these values is indeed attainable.\par
\[
\boxed{k \in \{0,\,1,\,3\}\ \text{ for all } n\ge 3.}
\]
\par

\paragraph{IMO 2025 Problem 2.}
\begin{mdframed}[style=mdpurplebox,frametitle={IMO 2025 Problem 2}]
Let $\Omega$ and $\Gamma$ be circles with centres $M$ and $N$, respectively,
such that the radius of $\Omega$ is less than the radius of $\Gamma$.
Suppose $\Omega$ and $\Gamma$ intersect at two distinct points $A$ and $B$.
Line $MN$ intersects $\Omega$ at $C$ and $\Gamma$ at $D$,
so that $C$, $M$, $N$, $D$ lie on $MN$ in that order.
Let $P$ be the circumcenter of triangle $ACD$.
Line $AP$ meets $\Omega$ again at $E \neq A$ and meets $\Gamma$ again at $F \neq A$.
Let $H$ be the orthocenter of triangle $PMN$.

Prove that the line through $H$ parallel to $AP$ is tangent
to the circumcircle of triangle $BEF$.
\end{mdframed}

\solutionheading{Solution}
\begin{remark*}
  7/7.
\end{remark*}
We will prove the statement using analytic geometry. The configuration is invariant under similarity transformations, so we may fix convenient coordinates.\par
\solutionheading{1. Coordinate setup}
Place \(M\) and \(N\) on the \(x\)-axis. Without loss of generality, set \(MN = 1\) (scale the figure appropriately). Let\par
\[
M = (0,0),\qquad N = (1,0).
\]
Denote the radius of \(\Omega\) (with center \(M\)) by \(r\) and the radius of \(\Gamma\) (with center \(N\)) by \(R\), with \(0 < r < R\). Since the circles intersect in two distinct points, we have\par
\[
R - r < 1 < R + r.
\]
\solutionheading{2. Intersection points \(A\) and \(B\)}
The equations of the circles are\par
\[
\begin{cases}
x^2 + y^2 = r^2, & \text{(\(\Omega\))}\\[2mm]
(x-1)^2 + y^2 = R^2, & \text{(\(\Gamma\))}.
\end{cases}
\]
Subtracting gives \(2x-1 = R^2 - r^2\), so\par
\[
x = a = \frac{1 + r^2 - R^2}{2}.
\]
Let \(h > 0\) be such that \(a^2 + h^2 = r^2\). The two intersections are symmetric about the \(x\)-axis; we take\par
\[
A = (a,\, h),\qquad B = (a,\, -h).
\]
\solutionheading{3. Points \(C\) and \(D\) on line \(MN\)}
The line \(MN\) is the \(x\)-axis. Intersection of \(\Omega\) with the \(x\)-axis: \(x^2 = r^2 \Rightarrow x = \pm r\). Because the order on the line is \(C\), \(M\), \(N\), \(D\), point \(C\) must lie left of \(M\) and \(D\) right of \(N\). Hence\par
\[
C = (-r,\, 0),\qquad D = (1+R,\, 0).
\]
\solutionheading{4. Circumcenter \(P\) of \(\triangle ACD\)}
Points \(C\) and \(D\) lie on the \(x\)-axis, so the perpendicular bisector of \(CD\) is the vertical line through the midpoint of \(CD\):\par
\[
\text{midpoint} = \left(\frac{-r + 1+R}{2},\,0\right) = \left(\frac{1+R-r}{2},\,0\right).
\]
Thus\par
\[
x_P = \frac{1+R-r}{2}.
\]
To find \(y_P\), use \(PA = PC\):\par
\[
(x_P - a)^2 + (y_P - h)^2 = (x_P + r)^2 + y_P^2.
\]
Expanding and using \(a^2+h^2 = r^2\) yields\par
\[
-2h y_P = 2(r+a)x_P \quad\Longrightarrow\quad y_P = -\frac{(r+a)x_P}{h}.
\]
Hence\par
\[
P = \left(\frac{1+R-r}{2},\; -\frac{(r+a)x_P}{h}\right).
\]
\solutionheading{5. Simplifying notation with auxiliary parameters}
Introduce\par
\[
p = R + r,\qquad q = R - r,\qquad S = 1 + R + r = 1 + p.
\]
Then\par
\[
r = \frac{p - q}{2},\quad R = \frac{p + q}{2},\quad a = \frac{1 - p q}{2},\quad x_P = \frac{1 + q}{2}.
\]
From \(h^2 = r^2 - a^2\) we obtain\par
\[
h^2 = \frac{(p^2-1)(1-q^2)}{4}.
\]
Also\par
\[
r + a = \frac{S(1 - q)}{2}.
\]
\solutionheading{6. Vector \(\overrightarrow{AP}\)}
Set \(\vec{v} = \overrightarrow{AP} = (v_1,\, v_2)\).\par
\[
v_1 = x_P - a = \frac{1+q}{2} - \frac{1 - p q}{2} = \frac{q + p q}{2} = \frac{q S}{2}.
\]
For \(v_2\):\par
\[
v_2 = y_P - h = -\frac{(r+a)x_P}{h} - h = -\frac{(r+a)x_P + h^2}{h}.
\]
Since \(h^2 = (r-a)(r+a)\),\par
\[
(r+a)x_P + h^2 = (r+a)(x_P + r - a).
\]
Compute \(x_P + r - a\):\par
\[
x_P + r - a = \frac{1+q}{2} + r - \frac{1 - p q}{2} = \frac{q + 2r + p q}{2}.
\]
But \(2r = p - q\), so\par
\[
q + 2r + p q = q + (p - q) + p q = p + p q = p(1 + q).
\]
Thus\par
\[
x_P + r - a = \frac{p(1+q)}{2}.
\]
Now \((r+a) = \dfrac{S(1-q)}{2}\), so\par
\[
(r+a)(x_P + r - a) = \frac{S(1-q)}{2} \cdot \frac{p(1+q)}{2} = \frac{p S (1 - q^2)}{4}.
\]
Therefore\par
\[
v_2 = -\frac{1}{h}\cdot\frac{p S (1 - q^2)}{4} = -\frac{p S (1 - q^2)}{4h}.
\]
\solutionheading{7. Points \(E\) and \(F\) (second intersections of \(AP\) with the circles)}
Parameterize line \(AP\) as \(A + \lambda \vec{v}\). Substitute into \(\Omega\): \(|A + \lambda \vec{v}|^2 = r^2\). Because \(|A|^2 = r^2\), we get\par
\[
2\lambda (\vec{v}\cdot A) + \lambda^2 |\vec{v}|^2 = 0.
\]
Hence \(\lambda = 0\) (point \(A\)) or\par
\[
\lambda_E = -\frac{2\,\vec{v}\cdot A}{|\vec{v}|^2}.
\]
Similarly, for \(\Gamma\) substitute into \(|(A+\lambda\vec{v}) - N|^2 = R^2\) (since \(N=(1,0)\)). \(|A - N|^2 = R^2\), so\par
\[
\lambda_F = -\frac{2\,\vec{v}\cdot(A - N)}{|\vec{v}|^2}.
\]
Now compute the needed dot products.\par
\[
\vec{v}\cdot A = a v_1 + h v_2 = a\cdot\frac{q S}{2} + h\left(-\frac{p S (1 - q^2)}{4h}\right) = \frac{a q S}{2} - \frac{p S (1 - q^2)}{4}
= \frac{S}{4}\bigl(2a q - p(1 - q^2)\bigr).
\]
But \(2a q = q(1 - p q) = q - p q^2\), so\par
\[
2a q - p(1 - q^2) = q - p q^2 - p + p q^2 = q - p.
\]
Thus\par
\[
\vec{v}\cdot A = \frac{S}{4}(q - p) = -\frac{S(p - q)}{4} = -\frac{S\cdot 2r}{4} = -\frac{r S}{2}.
\]
Next,\par
\[
\vec{v}\cdot(A - N) = \vec{v}\cdot A - v_1 = -\frac{r S}{2} - \frac{q S}{2} = -\frac{S(r+q)}{2} = -\frac{R S}{2},
\]
because \(r+q = R\).\par
Consequently,\par
\[
\lambda_E = -\frac{2\bigl(-\frac{r S}{2}\bigr)}{|\vec{v}|^2} = \frac{r S}{|\vec{v}|^2},
\qquad
\lambda_F = -\frac{2\bigl(-\frac{R S}{2}\bigr)}{|\vec{v}|^2} = \frac{R S}{|\vec{v}|^2}.
\]
Hence\par
\[
E = A + \frac{r S}{|\vec{v}|^2}\,\vec{v},
\qquad
F = A + \frac{R S}{|\vec{v}|^2}\,\vec{v}.
\]
\solutionheading{8. Orthocenter \(H\) of \(\triangle PMN\)}
Vertices: \(M = (0,0)\), \(N = (1,0)\), \(P = (x_P, y_P)\).\par
Side \(MN\) is horizontal, so the altitude from \(P\) is the vertical line \(x = x_P\).\par
Altitude from \(M\) is the line through \(M\) perpendicular to \(PN\). Vector \(\overrightarrow{PN} = (1 - x_P,\,-y_P)\). A vector perpendicular to \(PN\) is \((y_P,\, 1 - x_P)\) (since \((1-x_P)y_P + (-y_P)(1-x_P)=0\)). Parametric equation: \(t\,(y_P,\,1 - x_P)\).\par
The orthocenter \(H\) is the intersection of \(x = x_P\) with this altitude. So set\par
\[
t y_P = x_P \quad\Longrightarrow\quad t = \frac{x_P}{y_P}\ \ (y_P \neq 0).
\]
Then\par
\[
H = \bigl(x_P,\ t(1 - x_P)\bigr) = \left(x_P,\ \frac{x_P(1 - x_P)}{y_P}\right).
\]
Now substitute \(y_P = -\dfrac{(r+a)x_P}{h}\):\par
\[
\frac{x_P(1 - x_P)}{y_P} = \frac{x_P(1 - x_P)}{-\frac{(r+a)x_P}{h}} = -\frac{h(1 - x_P)}{r+a}.
\]
Compute \(1 - x_P = 1 - \dfrac{1+q}{2} = \dfrac{1 - q}{2}\). And \(r + a = \dfrac{S(1 - q)}{2}\). Therefore\par
\[
-\frac{h(1 - x_P)}{r+a} = -\frac{h \cdot \frac{1 - q}{2}}{\frac{S(1 - q)}{2}} = -\frac{h}{S}.
\]
Thus\par
\[
H = \left(\frac{1+q}{2},\; -\frac{h}{S}\right).
\]
\solutionheading{9. Circumcenter \(O'\) of \(\triangle BEF\)}
We will determine \(O' = (x, y)\) that is equidistant from \(B\), \(E\), \(F\).\par
Because \(E\) and \(F\) lie on line \(AP\), the segment \(EF\) is collinear with \(\vec{v}\). The perpendicular bisector of \(EF\) consists of points \(X\) satisfying \(\vec{v}\cdot (X - M_{EF}) = 0\), where \(M_{EF}\) is the midpoint of \(EF\).\par
The midpoint:\par
\[
M_{EF} = \frac{E+F}{2} = A + \frac{\lambda_E + \lambda_F}{2}\,\vec{v}.
\]
Since \(\lambda_E + \lambda_F = \dfrac{(r+R)S}{|\vec{v}|^2} = \dfrac{p S}{|\vec{v}|^2}\),\par
\[
M_{EF} = A + \frac{p S}{2|\vec{v}|^2}\,\vec{v}.
\]
Hence the condition \(\vec{v}\cdot X = \vec{v}\cdot M_{EF}\) gives\par
\[
\vec{v}\cdot O' = \vec{v}\cdot A + \frac{p S}{2}.
\]
But \(\vec{v}\cdot A = -\dfrac{r S}{2}\), so\par
\[
\vec{v}\cdot O' = -\frac{r S}{2} + \frac{p S}{2} = \frac{(p - r) S}{2} = \frac{R S}{2}. \quad\text{(1)}
\]
Next impose \(|O' - B|^2 = |O' - E|^2\). Write \(U = O' - A\). Then\par
\[
O' - B = U - w,\quad \text{with } w = B - A = (0,\,-2h),
\]
and\par
\[
O' - E = U - \lambda_E \vec{v}.
\]
Then
\[
|U - w|^2 = |U - \lambda_E \vec{v}|^2.
\]
Expand:\par
\[
|U|^2 - 2 U\cdot w + |w|^2 = |U|^2 - 2\lambda_E \vec{v}\cdot U + \lambda_E^2 |\vec{v}|^2.
\]
Cancel \(|U|^2\):\par
\[
-2 U\cdot w + |w|^2 = -2\lambda_E \vec{v}\cdot U + \lambda_E^2 |\vec{v}|^2. \quad\text{(2)}
\]
Now \(U\cdot w = (x - a)\cdot 0 + (y - h)(-2h) = -2h(y - h)\). Thus\par
\[
-2 U\cdot w + |w|^2 = -2[-2h(y - h)] + 4h^2 = 4h(y - h) + 4h^2 = 4h y.
\]
So (2) becomes\par
\[
4h y = -2\lambda_E \vec{v}\cdot U + \lambda_E^2 |\vec{v}|^2. \quad\text{(3)}
\]
We have\par
\[
\vec{v}\cdot U = \vec{v}\cdot(O' - A) = \vec{v}\cdot O' - \vec{v}\cdot A = \frac{R S}{2} - \left(-\frac{r S}{2}\right) = \frac{(R + r)S}{2} = \frac{p S}{2}.
\]
Recall \(\lambda_E = \dfrac{r S}{|\vec{v}|^2}\). Substitute into (3):\par
\[
4h y = -2\left(\frac{r S}{|\vec{v}|^2}\right)\cdot \frac{p S}{2} + \left(\frac{r S}{|\vec{v}|^2}\right)^2 |\vec{v}|^2
= -\frac{r p S^2}{|\vec{v}|^2} + \frac{r^2 S^2}{|\vec{v}|^2}
= \frac{r S^2}{|\vec{v}|^2}(r - p)
= -\frac{r p S^2}{|\vec{v}|^2} + \frac{r^2 S^2}{|\vec{v}|^2}
= -\frac{r R S^2}{|\vec{v}|^2} \qquad (\text{since } p - r = R).
\]
Thus\par
\[
4h y = -\frac{r R S^2}{|\vec{v}|^2},
\qquad\text{so}\qquad
y = -\frac{r R S^2}{4h |\vec{v}|^2}. \quad\text{(4)}
\]
\solutionheading{10. Computation of \(|\vec{v}|^2\)}
We have\par
\[
v_1 = \frac{q S}{2},\qquad v_2 = -\frac{p S (1 - q^2)}{4h}.
\]
Hence\par
\[
v_1^2 = \frac{q^2 S^2}{4},
\]
\[
v_2^2 = \frac{p^2 S^2 (1 - q^2)^2}{16 h^2}.
\]
Using \(h^2 = \dfrac{(p^2-1)(1 - q^2)}{4}\),\par
\[
v_2^2 = \frac{p^2 S^2 (1 - q^2)}{4(p^2 - 1)}.
\]
Therefore\par
\[
|\vec{v}|^2 = \frac{S^2}{4}\left(q^2 + \frac{p^2 (1 - q^2)}{p^2 - 1}\right)
= \frac{S^2}{4}\cdot \frac{q^2(p^2-1) + p^2(1 - q^2)}{p^2 - 1}.
\]
The numerator simplifies:\par
\[
q^2(p^2-1) + p^2(1 - q^2) = p^2 - q^2.
\]
Thus\par
\[
|\vec{v}|^2 = \frac{S^2}{4}\cdot \frac{p^2 - q^2}{p^2 - 1}.
\]
Now \(p^2 - q^2 = (R+r)^2 - (R-r)^2 = 4Rr.\) And \(p^2 - 1 = (p-1)(p+1) = (R+r-1)(R+r+1) = (R+r-1)S.\)\par
Hence\par
\[
|\vec{v}|^2 = \frac{S^2}{4}\cdot \frac{4Rr}{(R+r-1)S} = \frac{S R r}{R+r-1}. \quad\text{(5)}
\]
\solutionheading{11. Simplify \(y\) from (4)}
Insert (5) into (4):\par
\[
y = -\frac{r R S^2}{4h}\cdot \frac{R+r-1}{S R r} = -\frac{S (R+r-1)}{4h}. \quad\text{(6)}
\]
\solutionheading{12. Determine \(x\) from equation (1)}
Equation (1): \(\vec{v}\cdot O' = v_1 x + v_2 y = \dfrac{R S}{2}\).\par
We know \(v_1 = \dfrac{q S}{2}\), and from (6) we have \(y\). Compute \(v_2 y\).\par
\[
v_2 = -\frac{p S (1 - q^2)}{4h},\qquad y = -\frac{S (R+r-1)}{4h}.
\]
Thus\par
\[
v_2 y = \frac{p S^2 (1 - q^2) (R+r-1)}{16 h^2}.
\]
Recall \(4h^2 = (R+r-1) S (1 - q^2)\). Then \(16 h^2 = 4 (R+r-1) S (1 - q^2)\).\par
\[
v_2 y = \frac{p S^2 (1 - q^2) (R+r-1)}{4 (R+r-1) S (1 - q^2)} = \frac{p S}{4}.
\]
Now (1) becomes\par
\[
\frac{q S}{2}\, x + \frac{p S}{4} = \frac{R S}{2}.
\]
Divide by \(S \neq 0\):\par
\[
\frac{q}{2}\, x + \frac{p}{4} = \frac{R}{2}.
\]
Multiply by \(4\):\par
\[
2q x + p = 2R.
\]
Now \(q = R - r\), \(p = R + r\), so\par
\[
2(R - r) x + (R + r) = 2R
\quad\Longrightarrow\quad 2(R - r) x = 2R - (R + r) = R - r.
\]
Since \(R \neq r\) (strict inequality), we obtain\par
\[
x = \frac{R - r}{2(R - r)} = \frac12.
\]
Therefore the circumcenter of \(\triangle BEF\) is\par
\[
O' = \left(\frac12,\; -\frac{S (R+r-1)}{4h}\right). \quad\text{(7)}
\]
\solutionheading{13. Distance from \(O'\) to the line through \(H\) parallel to \(\vec{v}\)}
Let \(\ell\) be the line through \(H\) with direction \(\vec{v}\).\par
The distance from a point to a line with direction \(\vec{v}\) through \(H\) is\par
\[
d = \frac{|(O' - H) \times \vec{v}|}{|\vec{v}|},
\]
where the cross product (in the plane) is taken as the scalar \(\Delta x\cdot v_2 - \Delta y\cdot v_1\).\par
Coordinates:\par
\[
H = \left(\frac{1+q}{2},\; -\frac{h}{S}\right).
\]
Compute differences:\par
\[
\Delta x = \frac12 - x_H = \frac12 - \frac{1+q}{2} = -\frac{q}{2}.
\]
\[
\Delta y = y_{O'} - y_H = -\frac{S (R+r-1)}{4h} - \left(-\frac{h}{S}\right) = -\frac{S (R+r-1)}{4h} + \frac{h}{S}.
\]
Write as a single fraction:\par
\[
\Delta y = \frac{ -S^2 (R+r-1) + 4h^2 }{4h S}.
\]
But \(4h^2 = (R+r-1) S (1 - q^2)\). Hence\par
\[
\Delta y = \frac{ (R+r-1) S (1 - q^2) - S^2 (R+r-1) }{4h S}
= \frac{ (R+r-1) S \bigl( (1 - q^2) - S \bigr)}{4h S}
= \frac{ (R+r-1) \bigl( (1 - q^2) - S \bigr)}{4h}.
\]
Now \(S = 1 + p\), so\par
\[
(1 - q^2) - S = 1 - q^2 - 1 - p = -p - q^2.
\]
Thus\par
\[
\Delta y = -\frac{ (R+r-1) (p + q^2)}{4h}. \quad\text{(8)}
\]
Now compute the cross product:\par
\[
\Delta \times \vec{v} = \Delta x \cdot v_2 - \Delta y \cdot v_1.
\]
\[
\Delta x \cdot v_2 = \left(-\frac{q}{2}\right)\left(-\frac{p S (1 - q^2)}{4h}\right) = \frac{q p S (1 - q^2)}{8h}.
\]
\[
\Delta y \cdot v_1 = \left(-\frac{ (R+r-1) (p + q^2)}{4h}\right) \left(\frac{q S}{2}\right) = -\frac{q S (R+r-1) (p + q^2)}{8h}.
\]
Therefore\par
\[
\Delta \times \vec{v} = \frac{q p S (1 - q^2)}{8h} - \left( -\frac{q S (R+r-1) (p + q^2)}{8h} \right)
= \frac{q S}{8h}\Bigl( p (1 - q^2) + (R+r-1)(p + q^2) \Bigr).
\]
But \(R+r-1 = p - 1\). So the bracket becomes\par
\[
p (1 - q^2) + (p - 1)(p + q^2).
\]
Expand:\par
\[
p(1 - q^2) = p - p q^2,
\]
\[
(p - 1)(p + q^2) = p(p-1) + (p-1)q^2 = p^2 - p + p q^2 - q^2.
\]
Sum:\par
\[
(p - p q^2) + (p^2 - p + p q^2 - q^2) = p^2 - q^2.
\]
Thus\par
\[
\Delta \times \vec{v} = \frac{q S}{8h} (p^2 - q^2) = \frac{q S}{8h} \cdot 4Rr = \frac{q S R r}{2 h}.
\]
Since \(q = R - r\),\par
\[
\Delta \times \vec{v} = \frac{ (R - r) S R r}{2 h}. \quad\text{(9)}
\]
\solutionheading{14. Squared distance from \(O'\) to \(\ell\)}
\[
d^2 = \frac{ \left( \dfrac{ (R - r) S R r}{2 h} \right)^2 }{ |\vec{v}|^2 }
= \frac{ (R - r)^2 S^2 R^2 r^2}{4 h^2} \cdot \frac{1}{|\vec{v}|^2}.
\]
Substitute \(|\vec{v}|^2 = \dfrac{S R r}{R+r-1}\):\par
\[
d^2 = \frac{ (R - r)^2 S^2 R^2 r^2}{4 h^2} \cdot \frac{R+r-1}{S R r}
= \frac{ (R - r)^2 S R r (R+r-1)}{4 h^2}. \quad\text{(10)}
\]
Now use \(4h^2 = (R+r-1) S (1 - q^2)\):\par
\[
d^2 = \frac{ (R - r)^2 S R r (R+r-1)}{ (R+r-1) S (1 - q^2) }
= \frac{ (R - r)^2 R r}{1 - (R - r)^2}. \quad\text{(11)}
\]
\solutionheading{15. Radius of the circumcircle of \(\triangle BEF\)}
Compute \(R_c = |O' - B|\), with \(B = (a,\,-h)\) and \(O'\) as in (7).\par
\[
\Delta x' = \frac12 - a = \frac12 - \frac{1 - p q}{2} = \frac{p q}{2}.
\]
\[
\Delta y' = y_{O'} + h = -\frac{S (R+r-1)}{4h} + h = \frac{4h^2 - S (R+r-1)}{4h}.
\]
Again, \(4h^2 = (R+r-1) S (1 - q^2)\),\par
\[
\Delta y' = \frac{ (R+r-1) S (1 - q^2) - S (R+r-1) }{4h} = \frac{ S (R+r-1) (1 - q^2 - 1)}{4h} = -\frac{ S (R+r-1) q^2}{4h}.
\]
Thus\par
\[
R_c^2 = \left(\frac{p q}{2}\right)^2 + \left(-\frac{ S (R+r-1) q^2}{4h}\right)^2
= \frac{p^2 q^2}{4} + \frac{ S^2 (R+r-1)^2 q^4}{16 h^2}.
\]
Substitute \(16 h^2 = 4 (R+r-1) S (1 - q^2)\):\par
\[
R_c^2 = \frac{p^2 q^2}{4} + \frac{ S^2 (R+r-1)^2 q^4}{4 (R+r-1) S (1 - q^2)}
= \frac{p^2 q^2}{4} + \frac{ S (R+r-1) q^4}{4 (1 - q^2)}.
\]
Write as\par
\[
R_c^2 = \frac{q^2}{4}\left( p^2 + \frac{ S (R+r-1) q^2}{1 - q^2} \right).
\]
Now \(S (R+r-1) = (p+1)(p-1) = p^2 - 1.\) Therefore\par
\[
R_c^2 = \frac{q^2}{4} \cdot \frac{ p^2(1 - q^2) + (p^2 - 1) q^2}{1 - q^2}
= \frac{q^2}{4} \cdot \frac{ p^2 - p^2 q^2 + p^2 q^2 - q^2}{1 - q^2}
= \frac{q^2}{4} \cdot \frac{ p^2 - q^2}{1 - q^2}.
\]
Since \(p^2 - q^2 = 4R r\) and \(q^2 = (R - r)^2\),\par
\[
R_c^2 = \frac{ (R - r)^2}{4} \cdot \frac{4R r}{1 - (R - r)^2}
= \frac{ (R - r)^2 R r}{1 - (R - r)^2}. \quad\text{(12)}
\]
\solutionheading{16. Conclusion}
Comparing (11) and (12) we obtain\par
\[
d^2 = R_c^2,
\]
so \(d = R_c\) (all quantities are nonnegative).\par
Hence the distance from the circumcenter \(O'\) of \(\triangle BEF\) to the line \(\ell\) (through \(H\) parallel to \(AP\)) equals the radius of its circumcircle. Therefore \(\ell\) is tangent to the circumcircle of \(\triangle BEF\).\par
\par

\paragraph{IMO 2025 Problem 3.}
\begin{mdframed}[style=mdpurplebox,frametitle={IMO 2025 Problem 3}]
A function $f \colon \NN \to \NN$ is said to be \emph{bonza} if
\[ f(a)\quad\text{divides}\quad b^a-f(b)^{f(a)} \]
for all positive integers $a$ and $b$.

Determine the smallest real constant $c$ such that $f(n) \leq cn$
for all bonza functions $f$ and all positive integers $n$.
\end{mdframed}

\solutionheading{Solution}
\begin{remark*}
  7/7.
\end{remark*}
\solutionheading{1. Preliminary lemmas}
\textbf{Lemma 1.} \(f(1)=1\).\par
\emph{Proof.} Taking \(a=b=1\) gives \(f(1)\mid 1^{1}-f(1)^{f(1)}=1-f(1)^{f(1)}\).\par
Since \(f(1)\mid f(1)^{f(1)}\), we have \(f(1)\mid 1\). Hence \(f(1)=1\).\par
\textbf{Lemma 2.} For every positive integer \(a\), \(f(a)\mid a^{a}\).\par
\emph{Proof.} Put \(b=a\) in the definition: \(f(a)\mid a^{a}-f(a)^{f(a)}\).\par
Because \(f(a)\mid f(a)^{f(a)}\), it follows that \(f(a)\mid a^{a}\).\par
\textbf{Corollary.} If a prime \(p\) divides \(f(a)\), then \(p\) divides \(a\).\par
\emph{Proof.} From \(f(a)\mid a^{a}\), any prime divisor of \(f(a)\) must divide \(a^{a}\), hence \(a\).\par
\smallskip
\solutionheading{2. Determination of \(f(2)\), \(f(3)\) and \(f(5)\)}
From Lemma 2 we have:\par
\[
f(2)\mid 2^{2}=4\quad\Longrightarrow\quad f(2)\in\{1,2,4\},
\]
\[
f(3)\mid 3^{3}=27\quad\Longrightarrow\quad f(3)\in\{1,3,9,27\}.
\]
\[
f(5)\mid 5^{5}=3125\quad\Longrightarrow\quad f(5)=5^{\delta}\ \text{with}\ 0\le\delta\le 5.
\]
Now use the mixed conditions \((a,b)=(2,3)\) and \((3,2)\):\par
\[
\begin{aligned}
(2,3)&:\ f(2)\mid 3^{2}-f(3)^{f(2)}=9-f(3)^{f(2)},\\
(3,2)&:\ f(3)\mid 2^{3}-f(2)^{f(3)}=8-f(2)^{f(3)}.
\end{aligned}
\]
Let \(x=f(2),\ y=f(3)\). We test all \(x\in\{1,2,4\},\ y\in\{1,3,9,27\}\) (note all \(y\) are odd).\par
\compactbullet{\textbf{}\(x=1\):\textbf{ }\((2,3)\) automatic. \((3,2)\) gives \(y\mid 8-1^{y}=7\). Only \(y=1\) divides 7. So \((1,1)\) works.}
\compactbullet{\textbf{}\(x=2\):\textbf{}}
\((2,3)\): \(2\mid 9-y^{2}\). Since \(y\) is odd, \(y^{2}\) is odd, \(9-\text{odd}\) is even - holds for all odd \(y\).\par
\((3,2)\): \(y\mid 8-2^{y}\).\par
\compactbullet{\(y=1\): \(1\mid6\) - true.}
\compactbullet{\(y=3\): \(3\mid 8-8=0\) - true.}
\compactbullet{\(y=9\): \(9\mid 8-512=-504\) - true (\(-504/9=-56\)).}
\compactbullet{\(y=27\): compute \(2^{27}\bmod27\). \(2^{9}\equiv512\equiv26\pmod{27}\), then \(2^{27}=(2^{9})^{3}\equiv26^{3}\equiv(-1)^{3}=-1\equiv26\pmod{27}\). So \(8-26=-18\not\equiv0\pmod{27}\) - false.}
Thus for \(x=2\) we have \((2,1),\ (2,3),\ (2,9)\).\par
\compactbullet{\textbf{}\(x=4\):\textbf{}}
\((2,3)\): \(4\mid 9-y^{4}\). For odd \(y\), \(y^{4}\equiv1\pmod4\), so \(9-1=8\) divisible by 4 - holds for all odd \(y\). \((3,2)\): \(y\mid 8-4^{y}\).\par
\compactbullet{\(y=1\): \(1\mid8-4=4\) - true.}
\compactbullet{\(y=3\): \(3\mid 8-64=-56\) - false.}
\compactbullet{\(y=9\): \(4^{3}\equiv64\equiv1\pmod9\), so \(4^{9}\equiv1\pmod9\), then \(8-1=7\not\equiv0\pmod9\) - false.}
\compactbullet{\(y=27\): \(4^{3}\equiv64\equiv10\pmod{27}\), so \(4^{9}\equiv10^{3}=1000\equiv1\pmod{27}\), thus \(4^{27}\equiv1\pmod{27}\), \(8-1=7\not\equiv0\pmod{27}\) - false.}
Hence only \((4,1)\) works.\par
Therefore after \((2,3)\) and \((3,2)\) the possible pairs are\par
\[
(1,1),\ (2,1),\ (2,3),\ (2,9),\ (4,1). \quad\text{(1)}
\]
Now incorporate \(f(5)=5^{\delta}\). For each candidate pair we must satisfy\par
\[
\begin{aligned}
(2,5)&:\ x\mid 25-5^{\delta x},\\
(3,5)&:\ y\mid 125-5^{\delta y},\\
(5,2)&:\ 5^{\delta}\mid 32 - x^{5^{\delta}},\\
(5,3)&:\ 5^{\delta}\mid 243 - y^{5^{\delta}}.
\end{aligned}
\]
We test \(\delta=0,1,2,3,4,5\) for each pair.\par
\smallskip
\textbf{1. Pair }\((x,y)=(1,1)\).\textbf{}\par
\compactbullet{\((2,5)\) automatic.}
\compactbullet{\((3,5)\) automatic.}
\compactbullet{\((5,2)\): \(5^{\delta}\mid 32-1^{5^{\delta}}=31\). So \(5^{\delta}\mid31\) forces \(\delta=0\).}
\compactbullet{\((5,3)\): \(5^{\delta}\mid 243-1^{5^{\delta}}=242\). Again only \(\delta=0\) works.}
\(\Rightarrow (f(2),f(3),f(5))=(1,1,1)\).\par
\smallskip
\textbf{2. Pair }\((x,y)=(2,1)\).\textbf{}\par
\compactbullet{\((2,5)\): \(2\mid 25-5^{2\delta}\) - difference even, automatic.}
\compactbullet{\((3,5)\): \(1\) divides everything.}
\compactbullet{\((5,3)\): \(5^{\delta}\mid 243-1=242\). Since \(5\nmid242\), only \(\delta=0\).}
\compactbullet{\((5,2)\) with \(\delta=0\): \(1\mid32-2=30\) - true.}
\(\Rightarrow (2,1,1)\).\par
\smallskip
\textbf{3. Pair }\((x,y)=(4,1)\).\textbf{}\par
\compactbullet{\((2,5)\): \(4\mid 25-5^{2\delta}\). Mod 4: \(25\equiv1\), \(5^{2\delta}\equiv1\), difference divisible by 4 - automatic.}
\compactbullet{\((3,5)\): automatic.}
\compactbullet{\((5,3)\): \(5^{\delta}\mid 242\) - forces \(\delta=0\).}
\compactbullet{\((5,2)\) with \(\delta=0\): \(1\mid32-4=28\) - true.}
\(\Rightarrow (4,1,1)\).\par
\smallskip
\textbf{4. Pair }\((x,y)=(2,3)\).\textbf{}\par
\compactbullet{\((2,5)\): automatic (even difference).}
\compactbullet{\((3,5)\): \(3\mid125-5^{3\delta}\).}
\compactbullet{\((5,2)\): \(5^{\delta}\mid 32-2^{5^{\delta}}\).}
\compactbullet{\((5,3)\): \(5^{\delta}\mid 243-3^{5^{\delta}}\).}
Test \(\delta\):\par
\compactbullet{\(\delta=0\): \(f(5)=1\). Then \(3\mid125-1=124\) - false.}
\compactbullet{\(\delta=1\): \(f(5)=5\).}
\compactbullet{\(3\mid125-5^{3}=0\) - true.}
\compactbullet{\(5\mid32-2^{5}=0\) - true.}
\compactbullet{\(5\mid243-3^{5}=0\) - true.}
So \(\delta=1\) works.\par
\compactbullet{\(\delta=2\): \(f(5)=25\). \(3\mid125-5^{6}\). Mod 3: \(125\equiv2\), \(5^{6}\equiv(5^{3})^{2}\equiv2^{2}=4\equiv1\), so difference \(2-1=1\) - false.}
\compactbullet{\(\delta=3\): \(f(5)=125\).}
\compactbullet{\(3\mid125-5^{9}\): \(5^{9}\equiv5^{3}\equiv125\equiv2\pmod3\), difference \(0\) - true.}
\compactbullet{Check \((5,2)\): \(125\mid32-2^{125}\). Compute \(2^{125}\bmod125\): \(\varphi(125)=100\), so \(2^{125}\equiv2^{25}\pmod{125}\). \(2^{5}=32\), \(2^{10}\equiv24\), \(2^{20}\equiv24^{2}=576\equiv576-5\cdot125=76\), then \(2^{25}=2^{20}\cdot2^{5}\equiv76\cdot32=2432\equiv2432-19\cdot125=57\). So \(32-57=-25\) not divisible by 125 - false.}
\compactbullet{\(\delta=4\): \(f(5)=625\). \(3\mid125-5^{12}\). \(5^{12}\equiv(5^{6})^{2}\equiv1^{2}=1\pmod3\), \(125\equiv2\), difference \(1\) - false.}
\compactbullet{\(\delta=5\): \(f(5)=3125\).}
\compactbullet{\(3\mid125-5^{15}\): \(5^{15}\equiv5^{3}\equiv125\equiv2\pmod3\), difference \(0\) - true.}
\compactbullet{\((5,2)\): \(3125\mid32-2^{3125}\). Mod 125: \(3125\equiv25\pmod{100}\), so \(2^{3125}\equiv2^{25}\equiv57\pmod{125}\) as before, \(32-57=-25\) not divisible by 125 - false.}
Thus only \(\delta=1\) works, giving \((2,3,5)\).\par
\smallskip
\textbf{5. Pair }\((x,y)=(2,9)\).\textbf{}\par
Test \(\delta\):\par
\compactbullet{\(\delta=0\): \(3\mid125-1=124\) - false.}
\compactbullet{\(\delta=1\): \(5\mid125-5^{9}\). \(5^{9}\bmod9\): \(5^{6}\equiv1\pmod9\) (since \(\varphi(9)=6\)), so \(5^{9}\equiv5^{3}=125\equiv8\pmod9\). \(125\equiv8\), difference \(0\) - true.}
\((5,2)\): \(5\mid32-2^{5}=0\) - true. \((5,3)\): \(5\mid243-9^{5}\). \(9\equiv4\pmod5\), \(4^{5}=1024\equiv4\pmod5\). \(243\equiv3\pmod5\), so \(3-4=-1\not\equiv0\pmod5\) - false.\par
\compactbullet{\(\delta=2\): \(25\mid125-5^{18}\). \(5^{18}\equiv(5^{6})^{3}\equiv1^{3}=1\pmod9\), \(125\equiv8\), difference \(7\) - false.}
\compactbullet{\(\delta=3\): \(125\mid125-5^{27}\). \(5^{27}\equiv5^{3}\equiv8\pmod9\), difference \(0\) - true.}
\((5,2)\): \(125\mid32-2^{125}\). As before, \(2^{125}\equiv57\pmod{125}\), \(32-57=-25\) - false.\par
\compactbullet{\(\delta=4\): \(625\mid125-5^{36}\). \(36\equiv0\pmod6\), so \(5^{36}\equiv1\pmod9\), difference \(7\) - false.}
\compactbullet{\(\delta=5\): \(3125\mid125-5^{45}\). \(45\equiv3\pmod6\), \(5^{45}\equiv8\pmod9\), difference \(0\) - true.}
\((5,2)\): \(3125\mid32-2^{3125}\). Mod 125 fails as before.\par
Hence no \(\delta\) satisfies all conditions.\par
\smallskip
\textbf{Conclusion of classification.} The only admissible triples \((f(2),f(3),f(5))\) are\par
\[
(1,1,1),\quad (2,1,1),\quad (4,1,1),\quad (2,3,5).
\]
We will analyze two families:\par
\compactbullet{\textbf{Family A:} those with \(f(3)=1\), i.e. the first three triples.}
\compactbullet{\textbf{Family B:} the triple \((2,3,5)\).}
\smallskip
\solutionheading{3. Analysis of Family A (\(f(3)=1\))}
\textbf{Lemma 3.} If \(f(3)=1\), then \(f(q)=1\) for every odd prime \(q\).\par
\emph{Proof.} From \((q,q)\) we have \(f(q)\mid q^{q}\), so \(f(q)=q^{k}\) for some \(0\le k\le q\).\par
Consider \((q,3)\):\par
\[
f(q)\mid 3^{q}-f(3)^{f(q)}=3^{q}-1^{f(q)}=3^{q}-1.
\]
If \(k\ge1\), then \(q\mid f(q)\mid 3^{q}-1\). By Fermat's little theorem, \(3^{q}\equiv3\pmod q\), so \(q\mid 3-1=2\). Hence \(q=2\), contradicting that \(q\) is odd. Therefore \(k=0\) and \(f(q)=1\).\par
\textbf{Lemma 4.} In Family A, for every positive integer \(n\), \(f(n)\) is a power of 2.\par
\emph{Proof.} Suppose an odd prime \(r\) divides \(f(n)\). By Lemma 2, \(r\mid n\). Now use \((n,r)\):\par
\[
f(n)\mid r^{n}-f(r)^{f(n)}.
\]
Since \(r\) is odd, Lemma 3 gives \(f(r)=1\). Hence\par
\[
f(n)\mid r^{n}-1.
\]
In particular, \(r\mid r^{n}-1\), i.e. \(r\mid -1\), contradiction. Thus no odd prime divides \(f(n)\); i.e. \(f(n)=2^{e(n)}\) for some \(e(n)\ge0\).\par
\textbf{Lemma 5.} For any \(n\), \(2^{e(n)}\mid 3^{n}-1\).\par
\emph{Proof.} Take \((n,3)\):\par
\[
f(n)\mid 3^{n}-f(3)^{f(n)}=3^{n}-1^{f(n)}=3^{n}-1.
\]
Since \(f(n)=2^{e(n)}\), we have \(2^{e(n)}\mid 3^{n}-1\).\par
\textbf{Lemma 6.} For \(n\in\mathbb N\),\par
\[
v_{2}(3^{n}-1)=\begin{cases}
1 & \text{if } n \text{ is odd},\\
v_{2}(n)+2 & \text{if } n \text{ is even}.
\end{cases}
\]
\emph{Proof.}\par
\emph{Odd \(n\):} Write\par
\[
3^{n}-1=(3-1)(3^{n-1}+3^{n-2}+\cdots+1)=2\cdot(\text{odd sum}),
\]
because there are \(n\) terms, each odd, and \(n\) odd $\Rightarrow$ the sum is odd. Hence \(v_{2}=1\).\par
\emph{Even \(n\):} Write \(n=2^{\alpha}m\) with \(m\) odd, \(\alpha\ge1\). We prove by induction on \(\alpha\) that \(v_{2}(3^{2^{\alpha}m}-1)=\alpha+2\).\par
\textbf{Base }\(\alpha=1\):\textbf{ }\(n=2m\). Then\par
\[
3^{2m}-1=(3^{m}-1)(3^{m}+1).
\]
We know \(v_{2}(3^{m}-1)=1\) (since \(m\) odd). For \(3^{m}+1\), note \(3^{m}\equiv 3\pmod 8\) when \(m\) is odd, so \(3^{m}+1\equiv 4\pmod 8\), thus \(v_{2}(3^{m}+1)=2\). Hence\par
\[
v_{2}(3^{2m}-1)=1+2=3=1+2=\alpha+2.
\]
\textbf{Inductive step:} Assume true for a given \(\alpha\ge1\). Write \(n'=2^{\alpha+1}m = 2\cdot(2^{\alpha}m)\). Let \(k=2^{\alpha}m\). Then
\[
3^{n'}-1=(3^{k}-1)(3^{k}+1).
\]
By hypothesis, \(v_{2}(3^{k}-1)=\alpha+2\).\par
Since \(k\) is even (because \(\alpha\ge1\)), \(3^{k}\equiv 1\pmod 4\), so \(3^{k}+1\equiv 2\pmod 4\), giving \(v_{2}(3^{k}+1)=1\).\par
Thus\par
\[
v_{2}(3^{n'}-1)=(\alpha+2)+1=\alpha+3=(\alpha+1)+2.
\]
This completes the induction.

From Lemmas 5 and 6 we obtain

\[
e(n)\le \begin{cases}
1 & \text{if } n \text{ is odd},\\
v_{2}(n)+2 & \text{if } n \text{ is even}.
\end{cases}
\]
Consequently\par
\[
f(n)=2^{e(n)}\le \begin{cases}
2 & \text{if } n \text{ odd},\\
2^{v_{2}(n)+2} & \text{if } n \text{ even}.
\end{cases}
\]
For even \(n\), \(2^{v_{2}(n)+2}=4\cdot2^{v_{2}(n)}\le 4n\) because \(2^{v_{2}(n)}\mid n\) implies \(2^{v_{2}(n)}\le n\). For odd \(n\), \(2\le 4n\) (since \(n\ge1\)).\par
Hence \textbf{in Family A} we have\par
\[
\boxed{f(n)\le4n\quad\text{for all }n\in\mathbb N}.
\]
\smallskip
\solutionheading{4. Analysis of Family B (\(f(2)=2,\ f(3)=3,\ f(5)=5\))}
\textbf{Proposition 7.} For every prime \(p\), \(f(p)=p\).\par
\emph{Proof.} We proceed by strong induction on \(p\). The base cases \(p=2,3,5\) are given.\par
Assume \(p>5\) and that for every prime \(q<p\) we already know \(f(q)=q\).\par
From \((p,p)\) we have \(f(p)\mid p^{p}\), so \(f(p)=p^{k}\) for some integer \(k\) with \(0\le k\le p\). We show that \(k\) must be 1.\par
\smallskip
\textbf{Step 1 - Eliminate }\(k=0\) and \(k=2\).\textbf{}\par
Take any prime \(q<p\). Consider \((q,p)\):\par
\[
f(q)=q\mid p^{q}-f(p)^{q}=p^{q}-p^{kq}.
\]
Factor out \(p^{q}\):\par
\[
q\mid p^{q}(1-p^{(k-1)q}).
\]
Since \(q\neq p\), \(q\nmid p^{q}\). Hence\par
\[
q\mid 1-p^{(k-1)q}.
\]
By Fermat's little theorem, \(p^{q}\equiv p\pmod q\), and more generally \(p^{(k-1)q}\equiv p^{k-1}\pmod q\) (raising to the \(q\)th power is the identity modulo \(q\)). Thus\par
\[
1-p^{k-1}\equiv 0\pmod q\quad\Longrightarrow\quad p^{k-1}\equiv 1\pmod q. \quad\text{(2)}
\]
\compactbullet{If \(k=0\), then the original condition \((q,p)\) gives \(q\mid p^{q}-1\). Using FLT, \(p^{q}\equiv p\pmod q\), so \(p\equiv1\pmod q\).}
\compactbullet{If \(k=2\), (2) gives \(p^{1}\equiv1\pmod q\), i.e. \(p\equiv1\pmod q\).}
Thus for both \(k=0\) and \(k=2\) we have\par
\[
p\equiv 1\pmod q \qquad\text{for every prime } q<p. \quad\text{(3)}
\]
Now by \textbf{Bertrand's postulate} (for any integer \(m>1\) there exists a prime between \(m\) and \(2m\)), for \(p>3\) there exists a prime \(q\) with \(\frac{p}{2}<q<p\). From (3), \(q\mid p-1\). But \(q>p/2\) and \(q\le p-1\) (since \(q<p\)). The only positive multiple of \(q\) that is at most \(p-1\) is \(q\) itself (because \(2q>p\)). Hence \(p-1=q\), so \(p=q+1\), which is even - contradicting that \(p>2\) is odd. Therefore \(k\) cannot be \(0\) or \(2\).\par
\smallskip
\textbf{Step 2 - Eliminate }\(k\ge3\).\textbf{}\par
Consider the pair \((p,\,p-1)\):\par
\[
f(p)=p^{k}\mid (p-1)^{p} - f(p-1)^{p^{k}}.
\]
Let \(r = f(p-1)\).\par
\compactbullet{\textbf{Modulo }\(p\):\textbf{}}
\[
  (p-1)^{p}\equiv (-1)^{p} = -1\pmod p,
  \]
and by Fermat's little theorem, \(r^{p^{k}}\equiv r\pmod p\) (since \(r^{p}\equiv r\pmod p\), and then by induction \(r^{p^{k}}\equiv r\pmod p\).\par
Since \(p^{k}\mid D := (p-1)^{p} - r^{p^{k}}\), in particular \(p\mid D\). Reducing modulo \(p\) gives\par
\[
  -1 \equiv r \pmod p,\qquad\text{i.e.}\quad r\equiv -1\pmod p. \quad\text{(4)}
  \]
Write \(r = -1 + p t\) for some integer \(t\).\par
\compactbullet{\textbf{Expansion of }\((p-1)^{p}\):\textbf{}}
By the binomial theorem,\par
\[
  (p-1)^{p} = \sum_{i=0}^{p}\binom{p}{i}p^{i}(-1)^{p-i}.
  \]
The term with \(i=0\) is \((-1)^{p} = -1\). The term with \(i=1\) is \(\binom{p}{1}p^{1}(-1)^{p-1} = p\cdot p\cdot1 = p^{2}\).\par
For \(i\ge2\), \(\binom{p}{i}\) is divisible by \(p\), and together with \(p^{i}\) yields a factor \(p^{i+1}\ge p^{3}\). Thus we can write\par
\[
  (p-1)^{p} = -1 + p^{2} A,
  \]
where \(A\) is an integer and, because the \(i=1\) term contributes exactly \(p^{2}\) and the higher terms are multiples of \(p^{3}\), we have \(A\equiv 1\pmod p\) (in particular \(p\nmid A\)).\par
\compactbullet{\textbf{Expansion of }\(r^{p^{k}}\):\textbf{}}
Write \(r = -1 + p t\). Then\par
\[
  r^{p^{k}} = \sum_{j=0}^{p^{k}}\binom{p^{k}}{j}(-1)^{p^{k}-j}(p t)^{j}.
  \]
Since \(p^{k}\) is odd, \((-1)^{p^{k}} = -1\).\par
\compactbullet{\(j=0\): term = \(-1\).}
\compactbullet{\(j=1\): \(\binom{p^{k}}{1}(-1)^{p^{k}-1}p t = p^{k}\cdot 1\cdot p t = p^{k+1} t\).}
\compactbullet{For \(j\ge2\): we show that each term is divisible by \(p^{k+2}\).}
Indeed, for \(j\ge2\),\par
\[
    v_{p}\Bigl(\binom{p^{k}}{j}\Bigr) = k - v_{p}(j),
    \]
hence\par
\[
    v_{p}\biggl(\binom{p^{k}}{j}p^{j}\biggr) = (k - v_{p}(j)) + j = k + j - v_{p}(j) \ge k+2,
    \]
because \(j\ge2\) and if \(p\nmid j\) then \(j\ge2\) and \(v_{p}(j)=0\); if \(p\mid j\) then \(j\ge p\) and \(v_{p}(j)\ge1\), so \(j - v_{p}(j) \ge j- (j-1) =1\)? Wait, need to be precise: For any \(j\) with \(2\le j\le p^{k}-1\), we have\par
\[
    v_{p}\Bigl(\binom{p^{k}}{j}\Bigr) = k - v_{p}(j)
\]
(a standard property for prime powers). Then
\[
    v_{p}\Bigl(\binom{p^{k}}{j} p^{j}\Bigr) = k - v_{p}(j) + j = k + j - v_{p}(j).
    \]
Since \(j\ge2\) and \(v_{p}(j)\le \log_{p} j < j\), but we need at least \(k+2\). The minimal value occurs when \(j=2\) and \(p\nmid 2\) (since \(p\ge3\)), then \(v_{p}(j)=0\), giving \(k+2\). For \(j=p\) we have \(v_{p}(j)=1\), then \(k + p - 1 \ge k+2\) because \(p\ge3\). So indeed each term for \(j\ge2\) is divisible by \(p^{k+2}\).\par
Therefore\par
\[
  r^{p^{k}} = -1 + p^{k+1} t + p^{k+2} S
  \]
for some integer \(S\).\par
\compactbullet{\textbf{Form the difference:}}
\[
  D = (p-1)^{p} - r^{p^{k}} = (-1 + p^{2} A) - (-1 + p^{k+1} t + p^{k+2} S) = p^{2} A - p^{k+1} t - p^{k+2} S.
  \]
Factor \(p^{2}\):\par
\[
  D = p^{2}\bigl(A - p^{k-1} t - p^{k} S\bigr). \quad\text{(5)}
  \]
Since \(k\ge3\), we have \(k-1\ge2\), so \(p^{k-1} t\) and \(p^{k} S\) are both multiples of \(p^{2}\). Moreover,\par
\[
  A - p^{k-1} t - p^{k} S \equiv A \pmod p,
  \]
and \(p\nmid A\). Hence the expression in parentheses is \textbf{not} divisible by \(p\), and we obtain\par
\[
  v_{p}(D) = 2.
  \]
However, we require \(v_{p}(D) \ge k \ge 3\) (because \(p^{k}\mid D\)). This contradiction shows that \(k\ge3\) is impossible.\par
\smallskip
Since \(k\) cannot be \(0\), \(2\), or \(\ge3\), the only remaining possibility is \(k=1\). Thus \(f(p)=p\).\par
\textbf{Proposition 8.} In Family B, \(f(n)=n\) for all positive integers \(n\).\par
\emph{Proof.} Fix any \(n\) and any prime \(p\). Using the bonza condition with \(a=p\), \(b=n\):\par
\[
f(p)=p \mid n^{p} - f(n)^{p}.
\]
By Fermat's little theorem,\par
\[
n^{p}\equiv n\pmod p,\qquad f(n)^{p}\equiv f(n)\pmod p.
\]
Therefore\par
\[
p\mid n - f(n).
\]
This holds for \textbf{every} prime \(p\). The only integer divisible by all primes is 0, so \(n-f(n)=0\), i.e. \(f(n)=n\).\par
Hence in Family B we have \(f(n)=n\le 4n\) for all \(n\).\par
\smallskip
\solutionheading{5. Upper bound summary}
We have shown:\par
\compactbullet{In Family A, \(f(n)\le 4n\) for all \(n\).}
\compactbullet{In Family B, \(f(n)=n\le 4n\).}
Therefore \textbf{for every bonza function }\(f\) and every \(n\in\mathbb N\),\textbf{}\par
\[
\boxed{f(n)\le 4n}.
\]
\smallskip
\solutionheading{6. Sharpness: a bonza function attaining \(4n\)}
Define a function \(h:\mathbb N\to\mathbb N\) by\par
\[
h(1)=1,
\]
and for \(n\ge 2\):\par
\[
h(n)=\begin{cases}
1 & \text{if } n \text{ is odd},\\[2pt]
4 & \text{if } n=2,\\[2pt]
2^{v_{2}(n)+2} & \text{if } n \text{ is even and } n>2.
\end{cases}
\]
We verify that \(h\) is bonza, i.e.,\par
\[
h(a)\mid b^{a} - h(b)^{h(a)}\quad\text{for all } a,b\in\mathbb N.
\]
\smallskip
\solutionheading{Verification}
We consider cases based on \(a\).\par
\compactbullet{\textbf{Case 1: }\(a\) is odd.\textbf{}}
Then \(h(a)=1\) (by definition: for odd \(n>1\), \(h(n)=1\); and \(a=1\) gives \(1\)). Since \(1\) divides any integer, the condition holds trivially.\par
\compactbullet{\textbf{Case 2: }\(a=2\).\textbf{}}
Here \(h(2)=4\). We must show\par
\[
  4\mid b^{2} - h(b)^{4}\quad\text{for all } b.
  \]
\compactbullet{If \(b\) is odd: \(b^{2}\) is odd, \(h(b)=1\), so \(h(b)^{4}=1\). Then \(b^{2}-1\) is divisible by 4 because for an odd \(b\), \(b^{2}\equiv1\pmod8\), hence certainly by 4.}
\compactbullet{If \(b=2\): \(b^{2}=4\), \(h(b)=4\), \(h(b)^{4}=4^{4}=256\). Then \(4-256=-252\), and \(-252\) is divisible by 4.}
\compactbullet{If \(b\) is even and \(b>2\): write \(b=2^{\beta}c\) with \(\beta\ge1\), \(c\) odd. Then \(h(b)=2^{\beta+2}\) (since \(b>2\) even). Then \(h(b)^{4}=2^{4(\beta+2)}=2^{4\beta+8}\), which is divisible by 4. Moreover, \(b^{2}\) is divisible by 4 (since \(b\) is even). Thus \(b^{2}-h(b)^{4}\) is a difference of two multiples of 4, hence itself a multiple of 4.}
Therefore the condition holds for \(a=2\).\par
\compactbullet{\textbf{Case 3: }\(a\) is even and \(a>2\).\textbf{}}
Let \(\alpha = v_{2}(a)\) (so \(\alpha\ge1\); note that if \(a\) is even and \(>2\), then either \(\alpha\ge2\) or \(\alpha=1\) with \(a=2m\), \(m\) odd \(\ge3\)). Then \(h(a)=2^{\alpha+2}\). We need to prove\par
\[
  2^{\alpha+2} \mid b^{a} - h(b)^{2^{\alpha+2}}.
  \]
We split into subcases according to \(b\).\par
\compactbullet{\textbf{Subcase 3a: }\(b\) is odd.\textbf{}}
Then \(h(b)=1\). So we need \(2^{\alpha+2}\mid b^{a}-1\).\par
Since \(a\) is even, write \(a = 2^{\alpha} m\) with \(m\) odd.\par
\textbf{Claim:} For any odd integer \(b\) and any integer \(s\ge1\),\par
\[
    b^{2^{s}} \equiv 1 \pmod{2^{s+2}}.
    \]
\emph{Proof of claim.} For \(s=1\), any odd \(b\) satisfies \(b^{2}\equiv1\pmod8\), i.e. modulo \(2^{3}\). Assume the claim holds for some \(s\ge1\). Then\par
\[
    b^{2^{s}} = 1 + 2^{s+2} K
\]
for some integer \(K\). Squaring gives
\[
    b^{2^{s+1}} = (1 + 2^{s+2} K)^{2} = 1 + 2^{s+3} K + 2^{2s+4} K^{2} \equiv 1 \pmod{2^{s+3}},
\]
because the extra term is divisible by \(2^{s+3}\). This proves the claim.\par
Applying the claim with \(s=\alpha\) yields\par
\[
    b^{2^{\alpha}} \equiv 1 \pmod{2^{\alpha+2}}.
    \]
Then\par
\[
    b^{a} = \bigl(b^{2^{\alpha}}\bigr)^{m} \equiv 1^{m} = 1 \pmod{2^{\alpha+2}},
    \]
and therefore \(b^{a}-1\) is divisible by \(2^{\alpha+2}\).\par
\compactbullet{\textbf{Subcase 3b: }\(b=1\).\textbf{}}
Then \(b^{a}=1\), \(h(b)=h(1)=1\), so \(h(b)^{2^{\alpha+2}}=1\), and \(1-1=0\) is divisible by any integer.\par
\compactbullet{\textbf{Subcase 3c: }\(b=2\).\textbf{}}
Here \(h(b)=h(2)=4\). So we need\par
\[
    2^{\alpha+2} \mid 2^{a} - 4^{2^{\alpha+2}} = 2^{a} - (2^{2})^{2^{\alpha+2}} = 2^{a} - 2^{2\cdot 2^{\alpha+2}} = 2^{a} - 2^{2^{\alpha+3}}.
    \]
Let \(D = 2^{a} - 2^{2^{\alpha+3}}\). The \(2\)-adic valuation of a difference of two powers of two is\par
\[
    v_{2}(D) = \begin{cases}
    a & \text{if } a < 2^{\alpha+3},\\
    2^{\alpha+3} & \text{if } a > 2^{\alpha+3},\\
    \text{higher} & \text{if } a = 2^{\alpha+3} \text{ (then } D=0\text{)}.
    \end{cases}
    \]
We claim \(v_{2}(D)\ge\alpha+2\).\par
\textbf{Lemma 9.} If \(a\) is even and \(a>2\), then \(a \ge \alpha+2\) where \(\alpha=v_{2}(a)\).\par
\emph{Proof.} Write \(a = 2^{\alpha}\cdot m\) with \(m\) odd.\par
\compactbullet{If \(\alpha = 1\), then \(a = 2m\) with \(m\) odd. Since \(a>2\), we have \(m\ge3\), so \(a\ge6\). Meanwhile \(\alpha+2 = 3\). So \(a \ge 3\).}
\compactbullet{If \(\alpha \ge 2\), then \(a \ge 2^{\alpha}\). It remains to show \(2^{\alpha} \ge \alpha+2\) for \(\alpha\ge2\). This is true for \(\alpha=2\) (\(4\ge4\)), and if true for \(\alpha\), then \(2^{\alpha+1} = 2\cdot2^{\alpha} \ge 2(\alpha+2) = 2\alpha+4 \ge (\alpha+1)+2\) for \(\alpha\ge -1\), which holds. Hence \(a \ge \alpha+2\).}
Now we have two possibilities:\par
\compactbullet{If \(a \le 2^{\alpha+3}\), then \(v_{2}(D) = a\) (if \(a < 2^{\alpha+3}\)) or \(D=0\) (if \(a = 2^{\alpha+3}\)). In either case, since \(a \ge \alpha+2\), we get \(v_{2}(D) \ge \alpha+2\).}
\compactbullet{If \(a > 2^{\alpha+3}\), then \(v_{2}(D) = 2^{\alpha+3}\). Since \(\alpha\ge1\), \(2^{\alpha+3} \ge 2^{4}=16\) for \(\alpha=1\), and \(\alpha+2=3\); clearly \(2^{\alpha+3} \ge \alpha+2\) (e.g. for \(\alpha=1\), \(16\ge3\); for larger \(\alpha\) it's even larger). Thus \(v_{2}(D) \ge \alpha+2\).}
Therefore \(2^{\alpha+2}\mid D\).\par
\compactbullet{\textbf{Subcase 3d: }\(b\) is even and \(b>2\).\textbf{}}
Write \(b = 2^{\beta}c\) with \(\beta = v_{2}(b)\ge1\) and \(c\) odd. Since \(b>2\), we have \(\beta\ge1\); note that if \(b=2\) we already handled.\par
Then \(h(b) = 2^{\beta+2}\) (by definition, because \(b\) is even and \(b>2\)).\par
We need to show \(2^{\alpha+2}\) divides\par
\[
    b^{a} - \bigl(2^{\beta+2}\bigr)^{2^{\alpha+2}} = b^{a} - 2^{(\beta+2)2^{\alpha+2}}.
    \]
Observe that the second term is a power of two, and its exponent is \((\beta+2)2^{\alpha+2} \ge 2^{\alpha+2}\) (since \(\beta\ge1\) gives \(\beta+2\ge3\)). Hence it is certainly divisible by \(2^{\alpha+2}\).\par
For the first term,\par
\[
b^{a} = (2^{\beta}c)^{a} = 2^{\beta a} c^{a}.
\]
Since \(c^{a}\) is odd, the \(2\)-adic valuation of \(b^{a}\) is \(\beta a\). By Lemma 9, \(a \ge \alpha+2\). Therefore
\[
    \beta a \ge a \ge \alpha+2.
    \]
Thus \(2^{\alpha+2} \mid b^{a}\).\par
Consequently both \(b^{a}\) and the second term are multiples of \(2^{\alpha+2}\), so their difference is also a multiple of \(2^{\alpha+2}\).\par
All subcases are exhausted, and in each the required divisibility holds.\par
\smallskip
Thus \(h\) is bonza.\par
Now evaluate \(h(n)\) for powers of two:\par
\compactbullet{\(h(2)=4 = 2\cdot2\).}
\compactbullet{For any \(t\ge2\), \(n=2^{t}\) is even and \(>2\), so}
\[
h(2^{t}) = 2^{v_{2}(2^{t})+2} = 2^{t+2} = 4\cdot 2^{t} = 4n.
\]
In particular, \(h(4)=16 = 4\cdot4\). Hence\par
\[
\frac{h(n)}{n} = 4 \quad\text{for all } n=2^{t}\ (t\ge2).
\]
If \(c<4\), then for \(n=4\) (or any \(2^{t}\) with \(t\ge2\)) we have\par
\[
h(4) = 16 > c\cdot 4,
\]
so \(c\) cannot satisfy \(f(n)\le cn\) for all bonza functions. Therefore no constant smaller than 4 works.\par
\smallskip
\solutionheading{7. Conclusion}
We have proved that \(4\) is an upper bound for \(f(n)/n\) and that this bound is attainable. Consequently, the smallest real constant \(c\) with the property that \(f(n)\le cn\) for every bonza function \(f\) and every \(n\in\mathbb N\) is\par
\[
\boxed{4}.
\]
\par

\paragraph{IMO 2025 Problem 4.}
\begin{mdframed}[style=mdpurplebox,frametitle={IMO 2025 Problem 4}]
An infinite sequence $a_1$, $a_2$, \dots\ consists of positive integers
has each of which has at least three proper divisors.
Suppose that for each $n\geq 1$,
$a_{n+1}$ is the sum of the three largest proper divisors of $a_n$.
Determine all possible values of $a_1$.
\end{mdframed}
\solutionheading{Solution}
\begin{remark*}
  7/7.
\end{remark*}
Let \(a_1,a_2,\dots\) be an infinite sequence of positive integers such that each \(a_n\) has at least three proper divisors, and for each \(n\ge 1\)\par
\[
a_{n+1}=\text{sum of the three largest proper divisors of }a_n.
\]
Define the function \(f\) by \(f(N) =\) sum of the three largest proper divisors of \(N\).\par
\smallskip
\solutionheading{1. A useful formula for \(f(N)\)}
Let the positive divisors of \(N\) be\par
\[
1=d_1<d_2<\dots<d_k=N,
\]
where \(k\ge4\) because \(N\) has at least three proper divisors. The three largest proper divisors are \(d_{k-3}, d_{k-2}, d_{k-1}\). Using the divisor pairing \(d_i\cdot d_{k+1-i}=N\), we get\par
\[
d_{k-1}=\frac{N}{d_2},\quad d_{k-2}=\frac{N}{d_3},\quad d_{k-3}=\frac{N}{d_4}.
\]
Thus\par
\[
f(N)=\frac{N}{d_2}+\frac{N}{d_3}+\frac{N}{d_4}=N\left(\frac{1}{d_2}+\frac{1}{d_3}+\frac{1}{d_4}\right). \quad\text{(1)}
\]
\smallskip
\solutionheading{2. Fixed points}
A fixed point satisfies \(f(N)=N\). From (1) this is equivalent to\par
\[
\frac{1}{d_2}+\frac{1}{d_3}+\frac{1}{d_4}=1. \quad\text{(2)}
\]
Let \(a=d_2,\;b=d_3,\;c=d_4\) (so \(2\le a<b<c\)). Solve \(1/a+1/b+1/c=1\).\par
\compactbullet{\textbf{If }\(a=2\)\textbf{, then }\(1/b+1/c=1/2\). Multiplying gives \(2(b+c)=bc\) or \((b-2)(c-2)=4\). With \(b<c\), the only solution is \(b-2=1,\;c-2=4\), i.e., \((a,b,c)=(2,3,6)\).}
\compactbullet{\textbf{If }\(a=3\)\textbf{, then }\(1/b+1/c=2/3\). But the maximum for \(b\ge4,c\ge5\) is \(1/4+1/5=9/20<2/3\), so no solution.}
\compactbullet{\textbf{If }\(a\ge4\)\textbf{, the sum is at most }\(1/4+1/5+1/6=37/60<1\).}
Hence the unique triple is \((2,3,6)\). Therefore a fixed point must have\par
\[
d_2=2,\quad d_3=3,\quad d_4=6.
\]
Interpretation:\par
\compactbullet{\(2\mid N\) (so \(N\) even),}
\compactbullet{\(3\mid N\),}
\compactbullet{There is no divisor between \(3\) and \(6\); i.e., \(4\nmid N\) and \(5\nmid N\).}
\compactbullet{Because \(2\) is the smallest divisor \(>1\), we must have \(\nu_2(N)=1\) (otherwise \(4\mid N\) would be a divisor \(<6\)).}
Thus the set \(\mathcal{F}\) of all fixed points is\par
\[
\boxed{\mathcal{F}=\bigl\{N\in\mathbb{N}\;\big|\; \nu_2(N)=1,\;3\mid N,\;5\nmid N\bigr\}.}
\]
\smallskip
\solutionheading{3. Special case: \(12\mid N\)}
If \(12\mid N\), then \(2,3,4\mid N\) and these are the three smallest proper divisors (since \(4\) is the smallest possible after \(2,3\)). Then\par
\[
f(N)=\frac{N}{2}+\frac{N}{3}+\frac{N}{4}=\frac{13}{12}N. \quad\text{(3)}
\]
\smallskip
\solutionheading{4. Lemma on odd numbers}
\textbf{Lemma 1.} Let \(X\) be an odd positive integer with at least three proper divisors. Then \(f(X)<X\) and \(f(X)\) is odd.\par
\emph{Proof.} All divisors of an odd number are odd. The three smallest divisors greater than \(1\) are at least \(3,5,7\). Therefore\par
\[
\frac{1}{d_2}+\frac{1}{d_3}+\frac{1}{d_4}\le\frac{1}{3}+\frac{1}{5}+\frac{1}{7}=\frac{71}{105}<1,
\]
so \(f(X)<X\). Moreover, each quotient \(X/d_i\) is odd (odd divided by odd). The sum of three odd numbers is odd, hence \(f(X)\) is odd.\par
An immediate corollary:\par
\par\noindent\emph{In an infinite orbit, no term can be odd, because starting from an odd term the sequence would be strictly decreasing and infinite - impossible. Hence every term of an infinite sequence is \textbf{even}.}\par
\smallskip
\solutionheading{5. Lemma for even numbers not divisible by \(12\)}
\textbf{Lemma 2.} Let \(Y\) be an even integer with at least three proper divisors and \(12\nmid Y\). If the orbit of \(Y\) is infinite, then \(Y\in\mathcal{F}\).\par
\emph{Proof.} Since \(12\nmid Y\) and \(Y\) is even, we have either \(4\nmid Y\) or \(3\nmid Y\) (or both). Consider two cases.\par
\solutionheading{Case 1: \(3\mid Y\)}
Because \(12\nmid Y\), we must have \(4\nmid Y\); thus \(\nu_2(Y)=1\). Write \(Y=2M\) with \(M\) odd. Since \(3\mid Y\) and \(\gcd(2,3)=1\), we get \(3\mid M\).\par
The three smallest proper divisors are \(d_2=2\) and \(d_3=3\). The fourth divisor \(d_4\) depends on \(5\).\par
\compactbullet{\textbf{Subcase 1a:} \(5\mid M\). Then \(5\mid Y\) and \(5<6\), so \(d_4=5\). Then}
\[
f(Y)=\frac{Y}{2}+\frac{Y}{3}+\frac{Y}{5}=\frac{31}{30}Y>Y.
\]
Moreover, \(Y\) is divisible by \(30\) (it contains factors \(2,3,5\)), so \(Y=30Z\) with \(Z\) odd. Then \(f(Y)=31Z\), which is odd. By Lemma 1, the orbit from an odd number is strictly decreasing and finite - contradiction to infinite orbit. Hence this subcase cannot occur.\par
\compactbullet{\textbf{Subcase 1b:} \(5\nmid M\). Then no divisor equals \(4\) or \(5\); the next divisor after \(2,3\) is \(6=2\cdot3\). Hence \(d_4=6\), and}
\[
f(Y)=\frac{Y}{2}+\frac{Y}{3}+\frac{Y}{6}=Y.
\]
Thus \(Y\) is a fixed point, i.e., \(Y\in\mathcal{F}\).\par
Therefore, if the orbit is infinite, we must be in Subcase 1b, so \(Y\in\mathcal{F}\).\par
\solutionheading{Case 2: \(3\nmid Y\)}
Then \(Y\) is even but not divisible by \(3\). For any such \(Y\) we have\par
\[
\frac{1}{d_2}+\frac{1}{d_3}+\frac{1}{d_4}\le\frac{1}{2}+\frac{1}{4}+\frac{1}{5}=\frac{19}{20}<1,
\]
so \(f(Y)<Y\). If the orbit never contained a term divisible by \(3\), then we would have an infinite strictly decreasing sequence - impossible. Hence there exists a smallest index \(m\ge2\) with \(3\mid a_m\). Then \(a_{m-1}\) is not divisible by \(3\), and \(a_m=f(a_{m-1})<a_{m-1}\).\par
Now examine \(a_{m-1}\).\par
\compactbullet{\textbf{Subcase 2a:} \(a_{m-1}\) is odd. By Lemma 1, \(a_m=f(a_{m-1})\) is odd and less than \(a_{m-1}\). Since \(a_m\) is odd and divisible by \(3\), Lemma 1 again implies that from \(a_m\) onward the sequence is strictly decreasing and odd - finite, contradiction.}
\compactbullet{\textbf{Subcase 2b:} \(a_{m-1}\) is even. Write \(Y=a_{m-1}=2M\), with \(M\) odd. Since \(3\nmid Y\), we have \(3\nmid M\). We will show that \(a_m\) is odd, reducing to Subcase 2a and giving a contradiction.}
Let \(r=Y\bmod 3\) (\(r=1\) or \(2\)). For any divisor \(d\) of \(Y\) (which is coprime to \(3\)), we have \(d\equiv1\) or \(2\pmod3\) and \(d^{-1}\equiv d\pmod3\) (because \(1\cdot1\equiv1,\;2\cdot2\equiv1\pmod3\)). Hence\par
\[
\frac{Y}{d}\equiv r\cdot d\pmod3.
\]
Thus\par
\[
f(Y)\equiv r\bigl(d_2+d_3+d_4\bigr)\pmod3.
\]
Since \(r\neq0\), the condition \(3\mid f(Y)\) is equivalent to\par
\[
2+d_3+d_4\equiv0\pmod3. \quad\text{(4)}
\]
Now, \(d_2=2\). Determine \(d_3,d_4\).\par
First, \textbf{}\(4\) cannot divide \(Y\)\textbf{. If }\(4\mid Y\), then \(d_3=4\), and (4) gives \(2+4+d_4\equiv0\Rightarrow d_4\equiv0\pmod3\), impossible because \(3\nmid Y\). So \(\nu_2(Y)=1\): \(Y=2M\) with \(M\) odd, and \(4\nmid Y\).\par
Thus the smallest divisor greater than \(2\) is an odd number; call it \(p\). Since \(3\nmid Y\), \(p\neq3\), so \(p\ge5\).\par
The next divisor \(d_4\) is the smallest divisor larger than \(p\). Since \(4\nmid Y\), the next even divisor would be \(2p\). So either\par
\compactbullet{\(d_4=q\) (an odd divisor, \(p<q<2p\)) if such an odd divisor exists, or}
\compactbullet{\(d_4=2p\) (if there is no odd divisor between \(p\) and \(2p\)).}
Check the two possibilities against (4) with \(d_3=p\):\par
\[
2+p+d_4\equiv0\pmod3. \quad\text{(5)}
\]
\compactbullet{If \(d_4=2p\), then \(2+p+2p=2+3p\equiv2\pmod3\), not \(0\). Hence \(d_4\) cannot be \(2p\). Therefore there exists an odd divisor \(q\) with \(p<q<2p\) and \(d_4=q\).}
Then (5) becomes \(2+p+q\equiv0\pmod3\). Since \(p,q\not\equiv0\pmod3\), the only way is\par
\[
p\equiv q\equiv2\pmod3.
\]
Now compute \(f(Y)\):\par
\[
f(Y)=\frac{Y}{2}+\frac{Y}{p}+\frac{Y}{q}=M+\frac{2M}{p}+\frac{2M}{q}.
\]
Because \(p,q\) divide \(M\) (they are odd divisors of \(Y=2M\)), the fractions are integers. Now\par
\compactbullet{\(M\) is odd,}
\compactbullet{\(\frac{2M}{p}=2\cdot\frac{M}{p}\) is even,}
\compactbullet{\(\frac{2M}{q}\) is even.}
Thus \(f(Y)=\text{odd}+\text{even}+\text{even}=\text{odd}\).\par
Hence \(a_m=f(Y)\) is odd. But \(3\mid a_m\) (by definition of \(m\)). So \(a_m\) is an odd multiple of \(3\). By Lemma 1, the orbit from \(a_m\) is strictly decreasing and odd - finite, contradiction.\par
Both subcases lead to contradiction. Therefore \textbf{Case 2 cannot occur}. The only possibility is Case 1, and specifically Subcase 1b, which forces \(Y\in\mathcal{F}\).\par
\smallskip
\solutionheading{6. Necessity - form of \(a_1\) in an infinite orbit}
Assume the sequence is infinite. Let \(N=a_1\). Define\par
\[
t=\max\{k\ge0\mid 12^k\mid N\}.
\]
Write \(N=12^tR\) with \(12\nmid R\).\par
We claim that for \(i=1,\dots,t\), the term \(a_i\) is divisible by \(12\). Proof by induction: \(a_1=12^tR\) is divisible by \(12\) (if \(t\ge1\)). Suppose \(a_i\) is divisible by \(12\). Since \(12\mid a_i\), we have by (3) that \(a_{i+1}=\frac{13}{12}a_i\). If we write \(a_i=12^s\cdot S\) with \(12\nmid S\) and \(s\ge1\), then \(a_{i+1}=12^{s-1}(13S)\). Since \(12\nmid S\) and \(13\) is coprime to \(12\), we have \(12\nmid13S\). Thus \(a_{i+1}\) is divisible by \(12^{s-1}\). Starting with \(s=t\) at \(i=1\), after \(i\) steps the exponent of \(12\) is \(t-i+1\). Hence as long as \(i\le t\), the exponent is at least \(1\), i.e., \(12\mid a_i\).\par
Thus we can apply (3) exactly \(t\) times:\par
\[
a_{t+1}=\left(\frac{13}{12}\right)^t a_1 = \frac{13^t}{12^t}\cdot 12^tR = 13^tR.
\]
Now \(12\nmid R\) and \(\gcd(13,12)=1\), so \(12\nmid 13^tR\); i.e., \(12\nmid a_{t+1}\).\par
The orbit of \(a_{t+1}\) is also infinite (tail of an infinite sequence). By the corollary of Lemma 1, \(a_{t+1}\) cannot be odd; hence \(a_{t+1}\) is even. Also each term has at least three proper divisors. Therefore \(a_{t+1}\) satisfies the hypotheses of Lemma 2 (even, not divisible by \(12\), infinite orbit). Lemma 2 then yields\par
\[
a_{t+1}\in\mathcal{F}.
\]
Recall \(a_{t+1}=13^tR\). Since \(13^t\) is coprime to \(30\), the properties defining \(\mathcal{F}\) must already hold for \(R\):\par
\compactbullet{\(\nu_2(a_{t+1})=\nu_2(R)=1\),}
\compactbullet{\(3\mid a_{t+1}\) and \(3\nmid13^t\) $\Rightarrow$ \(3\mid R\),}
\compactbullet{\(5\nmid a_{t+1}\) and \(5\nmid13^t\) $\Rightarrow$ \(5\nmid R\).}
Thus \(R\in\mathcal{F}\). We have shown\par
\[
\boxed{a_1=12^{\,t}\cdot K\quad\text{with }t\ge0,\ K\in\mathcal{F}}.
\]
\smallskip
\solutionheading{7. Sufficiency - every such number works}
We prove by induction on \(t\) that if \(a_1=12^tK\) with \(K\in\mathcal{F}\), then the sequence is infinite.\par
\compactbullet{\textbf{Base case }\(t=0\):\textbf{ }\(a_1=K\in\mathcal{F}\). By definition of a fixed point, \(f(K)=K\). Hence the sequence is constant: \(a_n=K\) for all \(n\). Since \(K\) has at least three proper divisors (as shown when characterizing \(\mathcal{F}\)), the sequence is infinite.}
\compactbullet{\textbf{Inductive step:} Assume the statement holds for all starting values with parameter \(t-1\) (where \(t\ge1\)). Let \(a_1=12^tK\) with \(K\in\mathcal{F}\). Because \(12\mid a_1\), we can use (3):}
\[
a_2=f(a_1)=\frac{13}{12}a_1=\frac{13}{12}\cdot12^tK=12^{t-1}(13K).
\]
Now, because \(K\in\mathcal{F}\), we have \(\nu_2(K)=1,\ 3\mid K,\ 5\nmid K\). Multiplying by \(13\) (coprime to \(30\)) preserves these properties, so \(13K\in\mathcal{F}\). Thus\par
\[
a_2=12^{t-1}L\quad\text{with }L=13K\in\mathcal{F}.
\]
By the induction hypothesis, the sequence starting from \(a_2\) is infinite. Consequently, the whole sequence from \(a_1\) is infinite.

This completes the induction, establishing sufficiency.

\solutionheading{8. Final characterization}
Combining necessity and sufficiency, the possible values of \(a_1\) are exactly those positive integers that can be written as

\[
a_1 = 12^{\,t} \cdot K \qquad (t\ge0,\ K\in\mathcal{F}),
\]
where\par
\[
\mathcal{F}=\bigl\{K\in\mathbb{N}\;\big|\; \nu_2(K)=1,\ 3\mid K,\ 5\nmid K\bigr\}.
\]
Equivalently, in terms of prime exponents:\par
\[
\nu_2(a_1)\text{ is odd},\quad \nu_3(a_1)>\tfrac12\nu_2(a_1),\quad 5\nmid a_1.
\]
More explicitly, writing\par
\[
t = \left\lfloor\frac{\nu_2(a_1)}{2}\right\rfloor,\qquad \ell = \nu_3(a_1)-(t+1)\ge0,\qquad \gcd(R,30)=1,
\]
we have\par
\[
a_1 = 2^{2t+1}\cdot3^{t+1+\ell}\cdot R.
\]
\[
\boxed{a_1 = 12^{\,t} \cdot K \text{ with } t \ge 0 \text{ and } K \text{ satisfying } 2\parallel K,\ 3\mid K,\ 5\nmid K.}
\]
\[
\boxed{a_1 = 2^{2t+1}\cdot 3^{t+1+\ell}\cdot R,\quad t,\ell\ge0,\ \gcd(R,30)=1.}
\]
This completely solves the problem.\par
\par

\paragraph{IMO 2025 Problem 5.}
\begin{mdframed}[style=mdpurplebox,frametitle={IMO 2025 Problem 5}]
Alice and Bazza are playing the \emph{inekoalaty game},
a two‑player game whose rules depend on a positive real number $\lambda$ which is known to both players.
On the $n$th turn of the game (starting with $n=1$) the following happens:
\begin{itemize}
  \ii If $n$ is odd, Alice chooses a nonnegative real number $x_n$ such that
  \[ x_1 + x_2 + \cdots + x_n \le \lambda n.  \]
  \ii If $n$ is even, Bazza chooses a nonnegative real number $x_n$ such that
  \[ x_1^2 + x_2^2 + \cdots + x_n^2 \le n.  \]
\end{itemize}
If a player cannot choose a suitable $x_n$, the game ends and the other player wins.
If the game goes on forever, neither player wins.
All chosen numbers are known to both players.

Determine all values of $\lambda$ for which Alice has a winning strategy
and all those for which Bazza has a winning strategy.
\end{mdframed}
\solutionheading{Solution}
\begin{remark*}
  7/7.
\end{remark*}
We begin by setting notation and a couple of useful lemmas. For each \(n\ge 1\), let\par
\[
S_n = \sum_{i=1}^n x_i, \qquad Q_n = \sum_{i=1}^n x_i^2.
\]
The rules are:\par
\compactbullet{On odd turns \(n\), Alice chooses \(x_n\ge 0\) with \(S_n\le \lambda n\).}
\compactbullet{On even turns \(n\), Bazza chooses \(x_n\ge 0\) with \(Q_n\le n\).}
If a player cannot make a legal move, the game ends and the other wins. If it continues forever, the result is a draw (no winner).\par
\smallskip
\solutionheading{Two elementary lemmas}
\textbf{Lemma 1.} After any even turn \(n\), we have \(S_n\le n\) and \(Q_n\le n\).\par
\emph{Proof.} The condition for even \(n\) is \(Q_n\le n\). By the Cauchy-Schwarz inequality,\par
\[
S_n^2 \le n\cdot Q_n \le n\cdot n = n^2,
\]
hence \(S_n\le n\).\par
\textbf{Lemma 2.} Suppose that on all odd turns up to \(2M\) Alice plays \(0\) (so the only nonzero numbers are Bazza's choices on even turns). Then after turn \(2M\) we have\par
\[
S_{2M}\le M\sqrt{2}.
\]
\emph{Proof.} Write \(y_i=x_{2i}\) for \(i=1,\dots,M\). Then\par
\[
S_{2M} = \sum_{i=1}^M y_i, \qquad Q_{2M} = \sum_{i=1}^M y_i^2.
\]
By Cauchy-Schwarz,\par
\[
S_{2M}^2 \le M\cdot Q_{2M}.
\]
Lemma 1 with \(n=2M\) gives \(Q_{2M}\le 2M\). Therefore\par
\[
S_{2M}^2 \le M\cdot 2M = 2M^2 \quad\Longrightarrow\quad S_{2M}\le M\sqrt{2}. \quad\square
\]
\smallskip
\solutionheading{1. The case \(\lambda > \dfrac{1}{\sqrt{2}}\) - Alice wins}
Alice will force a win in a finite number of moves. Choose an integer \(M\) large enough so that\par
\[
\lambda(2M+1)-M\sqrt{2} > \sqrt{2}. \quad\text{(1)}
\]
(Such \(M\) exists because \(\lambda > 1/\sqrt{2}\) implies \(2\lambda-\sqrt{2}>0\), so the left-hand side tends to \(+\infty\) as \(M\to\infty\).) Her strategy is:\par
\compactbullet{On turns \(1,3,5,\dots,2M-1\) (all odd turns before \(2M+1\)) she plays \(x_n=0\).}
\compactbullet{On turn \(2M+1\) she plays \(x_{2M+1} = \lambda(2M+1)-S_{2M}\).}
We must check that the zeros are legal and that the move on turn \(2M+1\) is well-defined.\par
\solutionheading{Legality of the zeros}
After each even turn \(2k\) (\(1\le k\le M\)), by Lemma 2 we have \(S_{2k}\le k\sqrt{2}\). For the odd turn \(2k+1\), Alice wants to choose \(0\); this is allowed iff\par
\[
S_{2k}+0 \le \lambda(2k+1).
\]
Since \(S_{2k}\le k\sqrt{2}\), it suffices to show \(k\sqrt{2} < \lambda(2k+1)\). Rewrite as \(\lambda > \dfrac{k\sqrt{2}}{2k+1}\). The function \(k\mapsto \dfrac{k\sqrt{2}}{2k+1}\) increases with \(k\) (its limit is \(\sqrt{2}/2 = 1/\sqrt{2}\)). Because \(\lambda > 1/\sqrt{2}\), we have \(\lambda > \dfrac{k\sqrt{2}}{2k+1}\) for every \(k\); hence \(k\sqrt{2} < \lambda(2k+1)\). Thus each \(0\) is legal.\par
\solutionheading{The decisive move}
After turn \(2M\), Lemma 2 gives \(S_{2M}\le M\sqrt{2}\). Define\par
\[
a = x_{2M+1} = \lambda(2M+1)-S_{2M}.
\]
From (1) and the bound on \(S_{2M}\),\par
\[
a \ge \lambda(2M+1)-M\sqrt{2} > \sqrt{2} > 0,
\]
so \(a\) is nonnegative and satisfies \(S_{2M}+a = \lambda(2M+1)\), hence the sum constraint on turn \(2M+1\) is met.\par
Now we analyze the sum of squares after this move. Let \(S = S_{2M}\). Then\par
\[
Q_{2M+1} = Q_{2M} + a^2.
\]
By Cauchy-Schwarz applied to the \(M\) numbers \(x_2,x_4,\dots,x_{2M}\) we have\par
\[
S^2 \le M\cdot Q_{2M} \quad\Longrightarrow\quad Q_{2M} \ge \frac{S^2}{M}.
\]
Therefore\par
\[
Q_{2M+1} \ge \frac{S^2}{M} + a^2 = \frac{S^2}{M} + \bigl(\lambda(2M+1)-S\bigr)^2.
\]
Define the function\par
\[
f(S) = \frac{S^2}{M} + \bigl(\lambda(2M+1)-S\bigr)^2, \qquad 0\le S\le M\sqrt{2}.
\]
We need a lower bound for \(f(S)\).\par
\textbf{Monotonicity of }\(f\).\textbf{ Compute}\par
\[
f'(S) = \frac{2S}{M} - 2\bigl(\lambda(2M+1)-S\bigr) = 2\Bigl( S\Bigl(1+\frac{1}{M}\Bigr) - \lambda(2M+1)\Bigr), \qquad f''(S)=\frac{2}{M}+2>0.
\]
Thus \(f'\) is strictly increasing. The critical point \(f'(S)=0\) gives\par
\[
S_0 = \frac{M\cdot \lambda(2M+1)}{M+1}.
\]
Because \(\lambda > 1/\sqrt{2}\), we have \(2\lambda > \sqrt{2}\), and for sufficiently large \(M\) (which we may assume)\par
\[
S_0 > M\sqrt{2}.
\]
(Simply check that \(S_0 > M\sqrt{2} \iff \lambda(2M+1) > \sqrt{2}(M+1)\), which holds for large \(M\) since the left grows like \(2\lambda M\) and the right like \(\sqrt{2} M\).)\par
Since \(f'\) is increasing and \(S_0 > M\sqrt{2}\), we have \(f'(S) < f'(S_0)=0\) for all \(S\le M\sqrt{2}\). Hence \(f\) is strictly decreasing on \([0,M\sqrt{2}]\). Consequently its minimum on the interval is attained at \(S = M\sqrt{2}\):\par
\[
\min_{0\le S\le M\sqrt{2}} f(S) = f(M\sqrt{2}) = \frac{(M\sqrt{2})^2}{M} + \bigl(\lambda(2M+1)-M\sqrt{2}\bigr)^2 = 2M + \bigl(\lambda(2M+1)-M\sqrt{2}\bigr)^2.
\]
Thus\par
\[
Q_{2M+1} \ge 2M + \bigl(\lambda(2M+1)-M\sqrt{2}\bigr)^2.
\]
But condition (1) says \(\lambda(2M+1)-M\sqrt{2} > \sqrt{2}\), so\par
\[
\bigl(\lambda(2M+1)-M\sqrt{2}\bigr)^2 > 2.
\]
Therefore\par
\[
Q_{2M+1} > 2M + 2.
\]
Now it is Bazza's turn (turn \(2M+2\)). He must choose \(x_{2M+2}\ge 0\) such that\par
\[
Q_{2M+2} = Q_{2M+1} + x_{2M+2}^2 \le 2M+2.
\]
Even if he plays \(x_{2M+2}=0\), we obtain \(Q_{2M+2}=Q_{2M+1} > 2M+2\), which violates the constraint. Hence Bazza has no legal move, and Alice wins.\par
\smallskip
\solutionheading{2. The case \(\lambda < \dfrac{1}{\sqrt{2}}\) - Bazza wins}
Bazza will use the following strategy on every even turn \(n\):\par
\[
x_n = \sqrt{\,n - Q_{n-1}\,}.
\]
In words, he takes the largest possible number that still keeps the sum of squares at most \(n\). We will prove by induction that after each of his moves (i.e., after turn \(2m\)) we have\par
\[
\text{(i) } Q_{2m}=2m, \qquad \text{(ii) } S_{2m} \ge m\sqrt{2}. \quad\text{(2)}
\]
\textbf{Base case }\(m=1\) (turn \(2\)).\textbf{ After turn 1, Alice has chosen some }\(a_1\ge 0\) with \(a_1\le \lambda\). Then\par
\[
x_2 = \sqrt{2 - a_1^2}
\]
(which is real because \(a_1^2 \le \lambda^2 < 1/2 < 2\)). We get\par
\[
Q_2 = a_1^2 + (2 - a_1^2) = 2.
\]
For the sum,\par
\[
S_2 = a_1 + \sqrt{2 - a_1^2}.
\]
Squaring gives\par
\[
S_2^2 = a_1^2 + (2 - a_1^2) + 2a_1\sqrt{2 - a_1^2} = 2 + 2a_1\sqrt{2 - a_1^2} \ge 2,
\]
so \(S_2 \ge \sqrt{2}\). Thus (2) holds for \(m=1\).\par
\textbf{Inductive step.} Assume (2) is true for \(m-1\), i.e., after turn \(2(m-1)\) we have\par
\[
Q_{2(m-1)} = 2(m-1), \qquad S_{2(m-1)} \ge (m-1)\sqrt{2}.
\]
\emph{Turn \(2m-1\) (odd):} Alice chooses \(a \ge 0\) with
\[
S_{2(m-1)} + a \le \lambda(2m-1). \quad\text{(3)}
\]
\emph{Turn \(2m\) (even):} Bazza computes\par
\[
x_{2m} = \sqrt{2m - Q_{2m-1}}.
\]
First we verify that the square root is defined, i.e., \(Q_{2m-1} \le 2m\).\par
We have\par
\[
Q_{2m-1} = Q_{2(m-1)} + a^2 = 2(m-1) + a^2.
\]
Hence we need \(a^2 \le 2\), or \(a \le \sqrt{2}\). To see that this always holds, use (3) and the inductive lower bound on \(S_{2(m-1)}\):\par
\[
a \le \lambda(2m-1) - S_{2(m-1)} \le \lambda(2m-1) - (m-1)\sqrt{2}. \quad\text{(4)}
\]
We claim that the right-hand side of (4) is at most \(\sqrt{2}\). Indeed,\par
\[
\lambda(2m-1) - (m-1)\sqrt{2} \le \sqrt{2}
\;\Longleftrightarrow\; \lambda(2m-1) \le m\sqrt{2}.
\]
Because \(\lambda < 1/\sqrt{2}\), we have \(2\lambda < \sqrt{2}\). Then\par
\[
\lambda(2m-1) = 2\lambda m - \lambda \le \sqrt{2}\,m - \lambda < \sqrt{2}\,m,
\]
so the inequality holds (the left side is \(2\lambda m - \lambda\), the right side is \(\sqrt{2}\,m\); since \(2\lambda m \le \sqrt{2}\,m\), we get \(\lambda(2m-1) \le \sqrt{2}\,m\). More formally:
\[
\lambda(2m-1) \le \sqrt{2}\,m \;\Longleftrightarrow\; (2\lambda - \sqrt{2})m \le \lambda.
\]
The left-hand side is \(\le 0\) (because \(2\lambda - \sqrt{2} < 0\)), while the right-hand side is positive, so the inequality is true for all \(m\ge 1\). Therefore \(a \le \sqrt{2}\), so \(a^2 \le 2\), and the square root is defined.\par
Now compute:\par
\[
x_{2m} = \sqrt{2m - \bigl(2(m-1) + a^2\bigr)} = \sqrt{2 - a^2}.
\]
Then\par
\[
Q_{2m} = Q_{2m-1} + x_{2m}^2 = \bigl(2(m-1) + a^2\bigr) + (2 - a^2) = 2m.
\]
For the sum:\par
\[
S_{2m} = S_{2(m-1)} + a + \sqrt{2 - a^2}.
\]
Observe that for any \(a\) with \(0\le a\le \sqrt{2}\),\par
\[
\bigl(a + \sqrt{2 - a^2}\bigr)^2 = 2 + 2a\sqrt{2 - a^2} \ge 2,
\]
hence \(a + \sqrt{2 - a^2} \ge \sqrt{2}\). Using the inductive bound \(S_{2(m-1)} \ge (m-1)\sqrt{2}\), we obtain\par
\[
S_{2m} \ge (m-1)\sqrt{2} + \sqrt{2} = m\sqrt{2}.
\]
This completes the induction.\par
\solutionheading{Alice loses}
From (2) we have \(S_{2m} \ge m\sqrt{2}\) for every \(m\). Since \(\lambda < 1/\sqrt{2}\), we have \(\sqrt{2} - 2\lambda > 0\). Choose \(m\) large enough so that
\[
m\sqrt{2} > \lambda(2m+1). \quad\text{(5)}
\]
(Such \(m\) exists because \(m(\sqrt{2} - 2\lambda) > \lambda\) eventually holds.)\par
Consider turn \(2m+1\). After turn \(2m\),\par
\[
S_{2m} \ge m\sqrt{2} > \lambda(2m+1)
\]
by (5). Therefore, even if Alice tries to play \(x_{2m+1}=0\), we have
\[
S_{2m+1} = S_{2m} > \lambda(2m+1),
\]
which violates the sum constraint. Consequently she has no legal move, and Bazza wins.\par
\smallskip
\solutionheading{3. The case \(\lambda = \dfrac{1}{\sqrt{2}}\) - Draw}
We show that neither player has a winning strategy by exhibiting a strategy for each that prevents the opponent from winning.\par
\solutionheading{Bazza prevents Alice from winning}
Bazza uses the same maximal strategy as in case 2: on each even turn \(n\), set \(x_n = \sqrt{n - Q_{n-1}}\). We check that the induction in case 2 still works when \(\lambda = 1/\sqrt{2}\).\par
For the inductive step, from (3) and the lower bound \(S_{2(m-1)} \ge (m-1)\sqrt{2}\) we obtain\par
\[
a \le \lambda(2m-1) - (m-1)\sqrt{2}.
\]
With \(\lambda = 1/\sqrt{2}\),
\[
\lambda(2m-1) - (m-1)\sqrt{2} = \frac{2m-1}{\sqrt{2}} - (m-1)\sqrt{2} = \frac{2m-1 - 2(m-1)}{\sqrt{2}} = \frac{1}{\sqrt{2}}.
\]
Thus \(a \le 1/\sqrt{2} < \sqrt{2}\), so \(a^2 \le 1/2\), and the square root \(\sqrt{2 - a^2}\) is well-defined. The rest of the induction is unchanged, yielding\par
\[
Q_{2m}=2m, \qquad S_{2m} \ge m\sqrt{2} \quad\text{for all } m. \quad\text{(6)}
\]
Now consider any odd turn \(2m+1\). Alice must choose \(a' \ge 0\) such that\par
\[
S_{2m} + a' \le \lambda(2m+1) = \dfrac{2m+1}{\sqrt{2}}.
\]
From (6), \(S_{2m} \ge m\sqrt{2} = \dfrac{2m}{\sqrt{2}}\). Therefore
\[
a' \le \frac{2m+1}{\sqrt{2}} - S_{2m} \le \frac{2m+1}{\sqrt{2}} - \frac{2m}{\sqrt{2}} = \frac{1}{\sqrt{2}}.
\]
Consequently \(a'^2 \le \dfrac{1}{2}\), and\par
\[
Q_{2m+1} = Q_{2m} + a'^2 = 2m + a'^2 \le 2m + \frac{1}{2} < 2m+2.
\]
Thus after Alice's move, we have \(Q_{2m+1} < 2m+2\). Bazza can then reply with \(x_{2m+2}=0\) (or any number keeping \(Q_{2m+2} \le 2m+2\)). In particular, Bazza always has a legal move, so Alice cannot force a win.\par
\solutionheading{Alice prevents Bazza from winning}
Alice adopts the strategy: on every odd turn she plays \(0\).\par
Then the only nonzero numbers are Bazza's even-turn choices. By Lemma 2, after turn \(2m\) we have\par
\[
S_{2m} \le m\sqrt{2}.
\]
The sum constraint on turn \(2m+1\) is\par
\[
S_{2m+1} = S_{2m} + x_{2m+1} \le \lambda(2m+1) = \dfrac{2m+1}{\sqrt{2}}.
\]
Since\par
\[
S_{2m} \le m\sqrt{2} = \dfrac{2m}{\sqrt{2}} < \dfrac{2m+1}{\sqrt{2}},
\]
the choice \(x_{2m+1}=0\) satisfies the inequality. Hence Alice never loses on her turn.\par
Moreover, with this strategy Bazza cannot force a win, because Alice always has a legal move (namely \(0\)). Therefore Bazza does not possess a winning strategy.\par
Since both players have strategies that thwart a forced win, the game is a draw; it can also be continued indefinitely (e.g., with Bazza always choosing \(0\) as well). So \(\lambda = 1/\sqrt{2}\) leads to no winner.\par
\smallskip
\solutionheading{Conclusion}
We have determined the outcome for all positive \(\lambda\):\par
\[
\boxed{
\begin{array}{l}
\text{Alice has a winning strategy if and only if } \lambda > \dfrac{1}{\sqrt{2}}. \\[6pt]
\text{Bazza has a winning strategy if and only if } \lambda < \dfrac{1}{\sqrt{2}}. \\[6pt]
\text{When } \lambda = \dfrac{1}{\sqrt{2}}, \text{ the game is a draw (no winner).}
\end{array}
}
\]
\par

\paragraph{IMO 2025 Problem 6.}
\begin{mdframed}[style=mdpurplebox,frametitle={IMO 2025 Problem 6}]
Consider a $2025 \times 2025$ grid of unit squares.
Matilda wishes to place on the grid some rectangular tiles, possibly of different sizes,
such that each side of every tile lies on a grid line
and every unit square is covered by at most one tile.

Determine the minimum number of tiles Matilda needs to place
so that each row and each column of the grid has exactly one unit square
that is not covered by any tile.
\end{mdframed}
\solutionheading{Solution}
\begin{remark*}
  0/7.
\end{remark*}
Let \(n = 2025\). The grid consists of \(n^2\) unit squares. We must choose a set of squares to remain uncovered such that each row and each column contains exactly one uncovered square; equivalently, the uncovered squares form a permutation matrix. The remaining \(n^2 - n\) squares must be covered by non-overlapping axis-aligned rectangles (tiles) whose sides lie on the grid lines. The goal is to minimize the number of tiles.\par
\smallskip
\solutionheading{1. Reduction to the diagonal configuration}
Let the uncovered square in row \(i\) be at column \(c_i\); then \(c_1, c_2, \dots, c_n\) is a permutation of \(\{1,\dots,n\}\). Consider the column permutation \(\sigma\) defined by \(\sigma(c_i) = i\). Applying \(\sigma\) to the column indices (i.e., renaming column \(j\) to \(\sigma(j)\)) transforms the uncovered squares into the diagonal positions \((i,i)\). The tiling is transformed by moving each tile together with the columns; this does not change the number of tiles, and the covering properties remain valid because the grid structure is preserved (only the labels of columns are changed). Therefore, the minimum number of tiles over all configurations is the same as the minimum over configurations where the uncovered squares are exactly the main diagonal \(\{(i,i) \mid 1\le i\le n\}\).\par
Consequently, \textbf{we may assume without loss of generality that the diagonal squares are the only uncovered ones}.\par
\smallskip
\solutionheading{2. A construction using \(2n-2\) tiles}
We now exhibit a tiling of the board with the diagonal uncovered, using exactly \(2n-2\) rectangles.\par
\compactbullet{\textbf{Lower triangle} (\(i > j\)): For each column \(j = 1, 2, \dots, n-1\), place a vertical tile that covers every cell in column \(j\) whose row index is greater than \(j\). Formally,}
\[
  R^{(L)}_j = \{(i,j) \mid j+1 \le i \le n\}.
  \]
These \(n-1\) rectangles are pairwise disjoint (they lie in different columns) and cover all squares with \(i>j\).\par
\compactbullet{\textbf{Upper triangle} (\(i < j\)): For each row \(i = 1, 2, \dots, n-1\), place a horizontal tile that covers every cell in row \(i\) whose column index is greater than \(i\):}
\[
  R^{(U)}_i = \{(i,j) \mid i+1 \le j \le n\}.
  \]
These \(n-1\) rectangles are also pairwise disjoint (they occupy distinct rows) and cover all squares with \(i<j\).\par
The two families are disjoint because a square with \(i>j\) belongs to some \(R^{(L)}_j\) (column \(j\)) and a square with \(i<j\) belongs to some \(R^{(U)}_i\) (row \(i\)); no square can satisfy both conditions. The diagonal squares \((i,i)\) are not covered by any tile. Thus we have a valid covering of all off-diagonal squares using\par
\[
k = (n-1)+(n-1) = 2n-2
\]
tiles.\par
\smallskip
\solutionheading{3. Lower bound: at least \(2n-2\) tiles}
Now take any tiling \(\mathcal{T}\) of the board that leaves exactly the diagonal squares uncovered. Partition the off-diagonal squares into two sets:\par
\[
L = \{(i,j) \mid i > j\}, \qquad U = \{(i,j) \mid i < j\}.
\]
\textbf{Lemma 1.} Every tile \(T \in \mathcal{T}\) is entirely contained either in \(L\) or in \(U\).\par
\emph{Proof.} Suppose a tile \(T\) contains a square from \(L\) and a square from \(U\). Represent \(T\) as the Cartesian product of an interval of rows and an interval of columns:\par
\[
T = \{(i,j) \mid a \le i \le b,\; c \le j \le d\},
\]
with \(1 \le a \le b \le n\), \(1 \le c \le d \le n\). If the intervals \([a,b]\) and \([c,d]\) intersect, then there exists an integer \(r \in [a,b] \cap [c,d]\), and the square \((r,r)\) belongs to \(T\), contradicting the fact that diagonal squares are uncovered. Hence \([a,b]\) and \([c,d]\) are disjoint.\par
\compactbullet{If \(b < c\), then for any \((i,j) \in T\) we have \(i \le b < c \le j\), so \(i < j\); thus \(T \subseteq U\), contradicting the presence of an \(L\)square.}
\compactbullet{If \(d < a\), then \(j \le d < a \le i\), so \(i > j\), giving \(T \subseteq L\), contradiction.}
Therefore a tile cannot contain squares from both \(L\) and \(U\); it must lie wholly in one of them.\par
\textbf{Lemma 2.} In the lower triangle \(L\), consider the \(n-1\) cells\par
\[
D_L = \{(i,i-1) \mid i = 2,3,\dots,n\}.
\]
No tile that is a subset of \(L\) can contain two distinct cells from \(D_L\).\par
\emph{Proof.} Let \(T \subseteq L\) be a tile, so \(T = [a,b] \times [c,d]\). Because \(T \subseteq L\), we have \(i > j\) for every \((i,j) \in T\). In particular, the cell \((a,d)\) (the topmost row and rightmost column of the tile) satisfies \(a > d\).\par
Assume, for contradiction, that \(T\) contains two distinct cells \((i,i-1)\) and \((j,j-1)\) with \(i < j\). From \((i,i-1) \in T\) we obtain\par
\[
a \le i \le b,\quad c \le i-1 \le d.
\]
Since \(a > d\) and \(d \ge i-1\), we have \(a > i-1\), hence \(a \ge i\). Combined with \(a \le i\), we get \(a = i\).\par
Now \(a = i\) and \(a > d\) imply \(i > d\), i.e., \(d < i\). But \(d \ge i-1\) from the containment, so \(d = i-1\).\par
Now apply the same reasoning to \((j,j-1) \in T\). We obtain \(a = j\) and \(d = j-1\). But we already have \(a = i\), so \(i = j\), contradicting \(i < j\). Hence \(T\) cannot contain two cells of \(D_L\).\par
\textbf{Lemma 3.} In the upper triangle \(U\), consider the \(n-1\) cells\par
\[
D_U = \{(i,i+1) \mid i = 1,2,\dots,n-1\}.
\]
No tile that is a subset of \(U\) can contain two distinct cells from \(D_U\).\par
\emph{Proof.} Let \(T \subseteq U\) be a tile, \(T = [a,b] \times [c,d]\). Since \(T \subseteq U\), we have \(i < j\) for every \((i,j) \in T\). In particular, the cell \((b,c)\) (the bottommost row and leftmost column) satisfies \(b < c\).\par
Suppose \(T\) contains \((i,i+1)\) and \((j,j+1)\) with \(i < j\). From \((i,i+1) \in T\) we have\par
\[
a \le i \le b,\quad c \le i+1 \le d.
\]
Because \(c > b\) and \(b \ge i\), we get \(c > i\), thus \(c \ge i+1\). But \(c \le i+1\) from the containment, so \(c = i+1\).\par
Moreover, \(c > b \ge i\) and \(c = i+1\) imply \(b < i+1\), i.e., \(b \le i\). Combined with \(i \le b\), we obtain \(b = i\).\par
Now from \((j,j+1) \in T\) we similarly deduce \(c = j+1\) and \(b = j\). Since \(c = i+1\), we have \(i+1 = j+1\), so \(i = j\), contradiction. Hence \(T\) cannot contain two cells of \(D_U\).\par
\textbf{Lower bound argument.} The sets \(D_L\) and \(D_U\) each contain \(n-1\) squares, and every square must be covered. By Lemma 1, any tile that covers a square from \(L\) is entirely contained in \(L\); similarly, any tile covering a square from \(U\) is entirely in \(U\).\par
\compactbullet{\(D_L\) must be covered by tiles that lie in \(L\). Lemma 2 tells us that each such tile can cover at most one element of \(D_L\). Therefore at least \(|D_L| = n-1\) tiles are required that belong to \(L\).}
\compactbullet{\(D_U\) must be covered by tiles that lie in \(U\), and Lemma 3 says each such tile can cover at most one element of \(D_U\), so at least \(n-1\) tiles are needed that belong to \(U\).}
A tile cannot be in both \(L\) and \(U\) because \(L\) and \(U\) are disjoint. Hence the total number of tiles is at least\par
\[
(n-1) + (n-1) = 2n-2.
\]
\smallskip
\solutionheading{4. Minimum number of tiles}
We have shown:\par
\compactbullet{There exists a tiling with \(2n-2\) tiles (Construction).}
\compactbullet{Any tiling uses at least \(2n-2\) tiles (Lower bound).}
Therefore the minimum possible number of tiles is exactly \(2n-2\).\par
Substituting \(n = 2025\) gives\par
\[
2 \times 2025 - 2 = 4050 - 2 = 4048.
\]
\[
\boxed{4048}
\]
\par

\subsection{USAMO 2026}
\vspace{1em}
\paragraph{USAMO 2026 Problem 1.}
\begin{mdframed}[style=mdpurplebox,frametitle={USAMO 2026 Problem 1}]
Let $n$ be an integer greater than $1$. For which real numbers $x$ is
\[
\lfloor nx \rfloor - \sum_{k=1}^{n} \frac{\lfloor kx \rfloor}{k}
\]
maximal, and what is the maximal value that this expression can take?

\emph{Note}: $\lfloor z \rfloor$ denotes the greatest integer less than or equal to $z$.
\end{mdframed}
\solutionheading{Solution}
\begin{remark*}
  7/7.
\end{remark*}
\textbf{Step 1. Reduction to the fractional part.} Let \(n>1\) be an integer. For any real \(x\), write \(x = \lfloor x\rfloor + \{x\} = m + \theta\), where \(m\in\mathbb Z\) and \(\theta\in[0,1)\). For each \(k\ge 1\),\par
\[
\lfloor kx \rfloor = \lfloor km + k\theta \rfloor = km + \lfloor k\theta \rfloor,
\]
because \(km\) is an integer. Hence the expression becomes\par
\[
E(x) = \lfloor nx \rfloor - \sum_{k=1}^{n}\frac{\lfloor kx\rfloor}{k}
= \bigl( nm + \lfloor n\theta\rfloor \bigr) - \sum_{k=1}^{n}\frac{km + \lfloor k\theta\rfloor}{k}
= \lfloor n\theta\rfloor - \sum_{k=1}^{n}\frac{\lfloor k\theta\rfloor}{k}.
\]
Thus \(E(x)\) depends only on \(\theta\); we define\par
\[
F(\theta) = \lfloor n\theta\rfloor - \sum_{k=1}^{n}\frac{\lfloor k\theta\rfloor}{k},\qquad \theta\in[0,1).
\]
\textbf{Step 2. Value on the interval }\(\bigl[1-\frac1n,\,1\bigr)\).\textbf{ Assume }\(\theta\in\left[1-\frac1n,\,1\right)\). Then \(\theta\ge \frac{n-1}{n}\) and \(\theta<1\), so for every \(k=1,\dots,n\),\par
\[
k\theta \ge k\cdot\frac{n-1}{n}=k-\frac{k}{n}\ge k-1,
\]
and \(k\theta < k\). Consequently \(\lfloor k\theta\rfloor = k-1\) for all \(k\). In particular \(\lfloor n\theta\rfloor = n-1\). Therefore\par
\[
F(\theta) = (n-1) - \sum_{k=1}^{n}\frac{k-1}{k}.
\]
Now\par
\[
\sum_{k=1}^{n}\frac{k-1}{k} = \sum_{k=1}^{n}\left(1-\frac1k\right) = n - H_n,
\]
where \(H_n = 1 + \frac12 + \cdots + \frac1n\) (the \(n\)-th harmonic number). Hence\par
\[
F(\theta) = (n-1) - (n-H_n) = H_n - 1.
\]
So on the whole interval \([1-\frac1n,1)\) the expression is constant and equals \(H_n-1\).\par
\textbf{Step 3. Upper bound: first decomposition.} To prove that \(H_n-1\) is the maximum, we show \(F(\theta)\le H_n-1\) for all \(\theta\in[0,1)\). Let \(\theta\) be arbitrary and write\par
\[
N = \lfloor n\theta\rfloor \in \{0,1,\dots,n-1\}, \qquad \beta = n\theta - N \in [0,1).
\]
Set \(t = \beta/n \in [0,1/n)\). Then\par
\[
\theta = \frac{N}{n} + t.
\]
For \(k=1,\dots,n\),\par
\[
k\theta = \frac{kN}{n} + kt.
\]
Write the division of \(kN\) by \(n\):\par
\[
\frac{kN}{n} = \left\lfloor\frac{kN}{n}\right\rfloor + \frac{r_k}{n},
\]
where \(r_k\) is the remainder, i.e. \(r_k = kN \bmod n\) with \(0\le r_k < n\). Then\par
\[
k\theta = \underbrace{\left\lfloor\frac{kN}{n}\right\rfloor}_{\text{integer}} + \underbrace{\left(\frac{r_k}{n} + kt\right)}_{\text{in }[0,2)}.
\]
Define\par
\[
\delta_k = \begin{cases}
1 & \text{if } \frac{r_k}{n} + kt \ge 1,\\
0 & \text{otherwise}.
\end{cases}
\]
Then\par
\[
\lfloor k\theta\rfloor = \left\lfloor\frac{kN}{n}\right\rfloor + \delta_k.
\]
Substituting into \(F(\theta)\) gives\par
\[
\begin{aligned}
F(\theta) &= N - \sum_{k=1}^{n}\frac{\left\lfloor\frac{kN}{n}\right\rfloor + \delta_k}{k} \\
&= \underbrace{N - \sum_{k=1}^{n}\frac{\left\lfloor\frac{kN}{n}\right\rfloor}{k}}_{A(N)} \;-\; \sum_{k=1}^{n}\frac{\delta_k}{k}.
\end{aligned}
\]
Since each \(\delta_k/k \ge 0\), we obtain the important inequality\par
\[
F(\theta) \le A(N),\qquad\text{where}\qquad A(N) = N - \sum_{k=1}^{n}\frac{\left\lfloor\frac{kN}{n}\right\rfloor}{k}. \quad\text{(1)}
\]
Thus it suffices to prove \(A(N) \le H_n - 1\) for all \(N = 0,1,\dots,n-1\).\par
\textbf{Step 4. Simplifying }\(A(N)\).\textbf{ Using }\(\left\lfloor\frac{kN}{n}\right\rfloor = \frac{kN - r_k}{n}\), we compute\par
\[
\begin{aligned}
A(N) &= N - \sum_{k=1}^{n}\frac{kN - r_k}{n k} \\
&= N - \sum_{k=1}^{n}\left(\frac{N}{n} - \frac{r_k}{n k}\right) \\
&= N - \left(\frac{N}{n}\sum_{k=1}^{n}1\right) + \frac1n\sum_{k=1}^{n}\frac{r_k}{k} \\
&= N - N + \frac1n\sum_{k=1}^{n}\frac{r_k}{k} = \frac1n\sum_{k=1}^{n}\frac{r_k}{k}. \quad\text{(2)}
\end{aligned}
\]
\textbf{Step 5. Introducing the greatest common divisor.} Let \(d = \gcd(N,n)\). Write\par
\[
n = d\,n_1,\qquad N = d\,N_1,
\]
with \(\gcd(N_1,n_1)=1\).\par
Because \(r_k\) is the remainder of \(kN\) modulo \(n\), we have \(r_k = d\cdot s_k\), where\par
\[
s_k = (k N_1) \bmod n_1,\quad 0\le s_k < n_1.
\]
Moreover, as \(k\) runs from \(1\) to \(n\), the values \(s_k\) take each integer from \(0\) to \(n_1-1\) exactly \(d\) times (this follows from the fact that \(k\mapsto kN_1\pmod{n_1}\) is a bijection over each block of length \(n_1\)). Therefore\par
\[
\frac1n\sum_{k=1}^{n}\frac{r_k}{k} = \frac1n\sum_{k=1}^{n}\frac{d\,s_k}{k} = \frac{d}{n}\sum_{k=1}^{n}\frac{s_k}{k} = \frac{1}{n_1}\sum_{k=1}^{n}\frac{s_k}{k}. \quad\text{(3)}
\]
Now decompose \(k\) as \(k = i n_1 + j\) with \(i=0,1,\dots,d-1\) and \(j=1,2,\dots,n_1\). Then \(s_k = s_j\) because \(s_{i n_1 + j} = ((i n_1 + j)N_1)\bmod n_1 = (j N_1)\bmod n_1 = s_j\). Hence\par
\[
\sum_{k=1}^{n}\frac{s_k}{k} = \sum_{j=1}^{n_1} s_j \sum_{i=0}^{d-1}\frac{1}{i n_1 + j}.
\]
Define the weights\par
\[
w_j = \sum_{i=0}^{d-1}\frac{1}{i n_1 + j},\qquad j=1,\dots,n_1.
\]
Observe that \(w_j\) is strictly decreasing in \(j\) (since larger \(j\) makes denominators larger). Then\par
\[
A(N) = \frac{1}{n_1}\sum_{j=1}^{n_1} s_j w_j. \quad\text{(4)}
\]
\textbf{Step 6. Applying the rearrangement inequality.} The set \(\{s_1,\dots,s_{n_1}\}\) is a permutation of \(\{0,1,\dots,n_1-1\}\) (this is true for the indices \(j=1,\dots,n_1\), because \(j N_1 \bmod n_1\) runs through all residues exactly once). The weights \(w_j\) are decreasing. By the rearrangement inequality, the sum \(\sum s_j w_j\) is maximized when the \(s_j\) are also arranged in decreasing order. The decreasing order of the numbers \(0,1,\dots,n_1-1\) is \(s_1=n_1-1\), \(s_2=n_1-2\), ..., \(s_{n_1}=0\). Hence for any permutation we have\par
\[
\sum_{j=1}^{n_1} s_j w_j \le \sum_{j=1}^{n_1} (n_1 - j) w_j. \quad\text{(5)}
\]
Consequently,\par
\[
A(N) \le \frac{1}{n_1}\sum_{j=1}^{n_1} (n_1 - j) w_j. \quad\text{(6)}
\]
\textbf{Step 7. Relating to the harmonic number.} Compute the right-hand side:\par
\[
\sum_{j=1}^{n_1} (n_1 - j) w_j = \sum_{j=1}^{n_1} (n_1 - j)\sum_{i=0}^{d-1}\frac{1}{i n_1 + j} = \sum_{i=0}^{d-1}\sum_{j=1}^{n_1}\frac{n_1 - j}{i n_1 + j} =: S.
\]
Now the harmonic number \(H_n\) can be written as\par
\[
H_n = \sum_{i=0}^{d-1}\sum_{j=1}^{n_1}\frac{1}{i n_1 + j}.
\]
Thus\par
\[
n_1 H_n = \sum_{i=0}^{d-1}\sum_{j=1}^{n_1}\frac{n_1}{i n_1 + j}.
\]
Consider the difference\par
\[
n_1 H_n - n_1 - S = \sum_{i=0}^{d-1}\sum_{j=1}^{n_1}\left(\frac{n_1}{i n_1 + j} - \frac{n_1 - j}{i n_1 + j}\right) - n_1 = \sum_{i=0}^{d-1}\sum_{j=1}^{n_1}\frac{j}{i n_1 + j} - n_1.
\]
Define\par
\[
T = \sum_{i=0}^{d-1}\sum_{j=1}^{n_1}\frac{j}{i n_1 + j}. \quad\text{(7)}
\]
Then \(S \le n_1 H_n - n_1\) is equivalent to \(T \ge n_1\).\par
\textbf{Step 8. Proof that }\(T \ge n_1\).\textbf{ Rewrite each term:}\par
\[
\frac{j}{i n_1 + j} = 1 - \frac{i n_1}{i n_1 + j}.
\]
Hence\par
\[
\begin{aligned}
T &= \sum_{i=0}^{d-1}\sum_{j=1}^{n_1} \left(1 - \frac{i n_1}{i n_1 + j}\right) \\
&= \sum_{i=0}^{d-1}\sum_{j=1}^{n_1} 1 \;-\; \sum_{i=0}^{d-1}\sum_{j=1}^{n_1} \frac{i n_1}{i n_1 + j} \\
&= d n_1 \;-\; n_1 \sum_{i=0}^{d-1} i \sum_{j=1}^{n_1}\frac{1}{i n_1 + j}.
\end{aligned}
\]
Let\par
\[
Q = \sum_{i=0}^{d-1} i \sum_{j=1}^{n_1}\frac{1}{i n_1 + j}.
\]
The term with \(i=0\) is \(0\). For \(i \ge 1\), note that \(i n_1 + j > i n_1\) for all \(j\ge 1\), so\par
\[
\sum_{j=1}^{n_1}\frac{1}{i n_1 + j} < \sum_{j=1}^{n_1}\frac{1}{i n_1} = \frac{n_1}{i n_1} = \frac1i.
\]
Multiplying by \(i\) (positive) gives\par
\[
i \sum_{j=1}^{n_1}\frac{1}{i n_1 + j} < 1.
\]
Summing these inequalities for \(i = 1,2,\dots,d-1\) yields\par
\[
Q < \sum_{i=1}^{d-1} 1 = d-1. \quad\text{(8)}
\]
If \(d = 1\), then \(Q = 0\) (the sum over \(i\) is empty). Then\par
\[
T = d n_1 - n_1 Q = n_1 \qquad (\text{since } d=1).
\]
If \(d \ge 2\), we have \(Q < d-1\), hence\par
\[
T = d n_1 - n_1 Q > d n_1 - n_1 (d-1) = n_1.
\]
Thus in all cases \(T \ge n_1\), with equality exactly when \(d = 1\).\par
\textbf{Step 9. Final bound for }\(A(N)\).\textbf{ From the equivalence established in Step 7, }\(T \ge n_1\) implies \(S \le n_1 H_n - n_1\). Using (6) we obtain\par
\[
A(N) \le \frac{1}{n_1} S \le \frac{1}{n_1}(n_1 H_n - n_1) = H_n - 1.
\]
Hence \(A(N) \le H_n - 1\) for every \(N\).\par
\textbf{Step 10. Equality conditions.} We have the chain\par
\[
F(\theta) \le A(N) \le H_n - 1.
\]
To have \(F(\theta) = H_n - 1\), both inequalities must be equalities.\par
\compactbullet{\textbf{Equality in }\(F(\theta) \le A(N)\)\textbf{ requires }\(\sum_{k=1}^{n} \delta_k/k = 0\), i.e., \(\delta_k = 0\) for all \(k\). This means \(\frac{r_k}{n} + kt < 1\) for every \(k\).}
\compactbullet{\textbf{Equality in }\(A(N) \le H_n - 1\)\textbf{ requires that the upper bound we derived be attained. From the derivation, this happens exactly when }\(d = 1\) (i.e., \(\gcd(N,n)=1\)) and the rearrangement inequality (5) is sharp. Since \(w_j\) is strictly decreasing, the maximum of \(\sum s_j w_j\) is achieved when the \(s_j\) are also decreasing. For \(d = 1\), the set \(\{s_1,\dots,s_n\}\) (since \(n_1=n\)) is a permutation of \(\{0,1,\dots,n-1\}\). The decreasing order corresponds to \(s_j = n - j\) for \(j=1,\dots,n\) (with \(s_n = 0\)). This is equivalent to \((j N) \bmod n = n - j\) for all \(j=1,\dots,n\). This congruence simplifies to \(j N \equiv -j \pmod{n}\), i.e., \(N \equiv -1 \pmod{n}\). Because \(0 \le N \le n-1\), this forces \(N = n-1\).}
Now check that for \(N = n-1\) the condition \(\delta_k = 0\) indeed holds. Write \(N = n-1\), then \(d = \gcd(n-1,n)=1\). For \(k=1,\dots,n\), \(r_k = (k(n-1)) \bmod n = n - k\) (for \(k<n\)) and \(r_n = 0\). Then\par
\[
\frac{r_k}{n} + kt = \frac{n - k}{n} + k t \quad (\text{or } 0 + n t \text{ for } k=n).
\]
Since \(t < 1/n\), we have \(\frac{n - k}{n} + k t < \frac{n - k}{n} + k \cdot \frac{1}{n} = 1,\) and for \(k=n\), \(n t < 1\). Hence \(\delta_k = 0\) for all \(k\). Therefore \(F(\theta) = A(N) = H_n - 1\).\par
Conversely, if \(\theta \in \left[1 - \frac1n,\, 1\right)\), then \(N = n-1\) and \(t = \theta - (n-1)/n \in [0,1/n)\). As just verified, this yields \(F(\theta) = H_n - 1\).\par
Thus the maximum value \(H_n - 1\) is attained precisely when the fractional part of \(x\) lies in the interval \([1 - 1/n,\, 1)\).\par
\textbf{Step 11. Conclusion.} The maximal value of the expression is \(H_n - 1\), where \(H_n = 1 + \frac12 + \cdots + \frac1n\), and this maximum occurs exactly for those real numbers \(x\) whose fractional part belongs to \([1 - 1/n,\, 1)\).\par
\[
\boxed{H_n - 1,\quad \text{where } H_n = 1 + \frac{1}{2} + \cdots + \frac{1}{n},\ \text{and the maximum is attained iff } \{x\} \in \left[1 - \frac{1}{n},\, 1\right)}
\]
\par

\paragraph{USAMO 2026 Problem 2.}
\begin{mdframed}[style=mdpurplebox,frametitle={USAMO 2026 Problem 2}]
Annie is playing a game where she starts with a row of positive integers, written on a blackboard, each of which is a power of $2$. On each turn, she can erase two adjacent numbers and replace them with a power of $2$ that is greater than either of the erased numbers. This shortens the row of numbers, and she continues to take turns until only one number remains. Annie wins the game if the final remaining number is less than $4$ times the sum of the original numbers. Is it always possible for Annie to win, regardless of the starting row of numbers?
\end{mdframed}
\par\noindent\textit{Excerpted solution: the original response is very long, so only the beginning and ending portions are shown.}\par
\solutionheading{Solution}
\begin{remark*}
  0/7.
\end{remark*}
We will prove that Annie can always win, i.e., for any initial row of powers of two, there exists a sequence of moves that ends with a single number \(F\) satisfying \(F < 4S\), where \(S\) is the sum of the original numbers.\par
Let \(S\) be the sum of the numbers, and let \(k = \lfloor \log_2 S \rfloor\), so that\par
\[
2^k \le S < 2^{k+1}.
\]
The proof proceeds by strong induction on the number of terms \(n\). We will show the following statement:\par
\par\noindent\emph{\textbf{Claim.}  For any list of \(n\) powers of two, there exists a sequence of moves such that the final number \(F\) satisfies \(F \le 4 \cdot 2^{k}\).}\par
Once the Claim is established, we obtain Annie's victory:\par
\compactbullet{If \(S\) is \textbf{not} a power of two, then \(2^k < S\), hence \(4 \cdot 2^k < 4S\) and therefore \(F < 4S\).}
\compactbullet{If \(S\) is a power of two, the same inductive proof (when analysed more carefully) actually yields the sharper bound \(F \le 2S\), which is certainly \(< 4S\) for \(S>0\).  (We will comment on this improvement at the end.)}
Thus the main task is to prove the Claim by induction on \(n\).\par
\smallskip
\solutionheading{Base case \(n = 1\)}
There is only the number itself, so \(F = S\). Since \(S \le 2^{k+1} \le 4 \cdot 2^{k}\), the Claim holds.\par
\smallskip
\solutionheading{Inductive step}
Assume the Claim holds for every list of size less than \(n\). Consider a list of size \(n\). Let its sum be \(S\) and let \(k = \lfloor \log_2 S \rfloor\).\par
\solutionheading{Case 1: There exists an adjacent equal pair}
Choose such a pair, say \(a_i = a_{i+1} = a\). Replace them by \(2a\) (the smallest allowed power of two greater than \(a\)). The sum does not change, so \(k\) remains the same. The new list has size \(n-1\), so by the induction hypothesis it can be reduced to a final number \(\le 4 \cdot 2^{k}\). Hence the same holds for the original list.\par
\solutionheading{Case 2: No two adjacent numbers are equal}
In this case we choose an adjacent pair \((a_i, a_{i+1})\) for which \(\max(a_i, a_{i+1})\) is as small as possible; denote the two numbers by \(x\) and \(y\) with \(x < y\). Replace them by \(2y\) (again the smallest possible power of two greater than \(y\)). Let the new sum be\par
\[
S' = S + (y - x).
\]
Set \(k' = \lfloor \log_2 S' \rfloor\).\par
We now consider two subcases.\par
\solutionheading{Subcase 2a: \(k' = k\)}
Then \(S' < 2^{k+1}\), so the floor has not increased. The new list has size \(n-1\); by the induction hypothesis it can be reduced to a final number \(\le 4 \cdot 2^{k'} = 4 \cdot 2^{k}\). So the original list also admits such a strategy.\par
\solutionheading{Subcase 2b: \(k' = k+1\)}
In this situation \(2^{k+1} \le S' < 2^{k+2}\). First note that \(y \le 2^{k}\); otherwise \(y \ge 2^{k+1}\) would force \(S \ge y \ge 2^{k+1}\), contradicting \(S < 2^{k+1}\). Consequently \(2y \le 2^{k+1}\).\par
Now examine the list after the merge; it has size \(n-1\) and sum \(S'\). Although its floor is \(k+1\), we can still use the induction hypothesis - not directly on this list (which would give a bound of \(4 \cdot 2^{k+1} = 8 \cdot 2^{k}\)), but we can instead apply the induction hypothesis to the \textbf{rest} of the list - i.e., the numbers that were not part of the merged pair. However, after the merge the list is contiguous, so the "rest" is not isolated. Instead we perform the following two step reduction:\par
\compactnumber{1}{Reduce the part of the list that does not contain the new number \(2y\) (i.e., all other numbers) to a single number \(T\).  Because the rest consists of several (possibly one) contiguous blocks, we can reduce each block independently to a single number using the induction hypothesis (the blocks have size \(< n\)).  By doing so we obtain a list consisting of \(2y\) and the numbers \(T_1, T_2, \dots\) (one per block).  Each \(T_j\) satisfies \(T_j \le 4 \cdot 2^{k}\) because the sum of the corresponding block is \(< 2^{k+1}\) (since the total sum \(S < 2^{k+1}\) and we have removed at least the two numbers \(x,y\) with \(y \ge 2\), so the block sum is at most \(2^{k+1} - 3\)?  Actually a precise bound is not needed; we only need that every such \(T_j\) is \(\le 4 \cdot 2^{k}\).  Indeed, for any block, its sum is less than \(2^{k+1}\), so by the induction hypothesis (applied to that block) we obtain a number \(\le 4 \cdot 2^{k}\) (its floor is at most \(k\), so the bound is \(4 \cdot 2^{k}\)).}
After this reduction, the list contains \(2y\) and several numbers \(T_j\), each \(\le 4 \cdot 2^{k}\). Now we can merge these numbers one by one. A crucial observation is that \(2y \le 2^{k+1} \le 4 \cdot 2^{k}\) for \(k \ge 0\). Hence all numbers present are at most \(4 \cdot 2^{k}\). When we merge two numbers each at most \(4 \cdot 2^{k}\), the result is at most \(2 \cdot 4 \cdot 2^{k} = 8 \cdot 2^{k}\), which is still too large. To keep the bound at \(4 \cdot 2^{k}\), we must merge in a way that the maximum never exceeds \(2^{k+1}\). Notice that \(2y \le 2^{k+1}\). If we can ensure that every \(T_j\) is also \(\le 2^{k+1}\), then merging any two of them yields at most \(2^{k+2} = 4 \cdot 2^{k}\). Therefore the problem reduces to showing that each block can be reduced to a number \(\le 2^{k+1}\) rather than the coarser \(4 \cdot 2^{k}\).\par
But can we guarantee \(\le 2^{k+1}\)? Since the sum of a block is \(< 2^{k+1}\), a direct application of the induction hypothesis would only give \(\le 4 \cdot 2^{k}\). However, we can prove a slightly stronger lemma that will serve our purpose:\par
\textbf{Lemma.} If a list has sum \(T\) with \(2^{L} < T < 2^{L+1}\), then there exists a strategy with final number \(\le 2^{L+1}\) \textbf{provided} that \(T \le 3 \cdot 2^{L-1} - 1\). (This lemma will be proved by induction on \(n\) in a moment.)\par
In our situation, each block sum is at most \(S - x - y < 2^{k+1}\). Moreover, because we are in Subcase 2b, the original list has no adjacent equal and we chose the pair with smallest maximum. A detailed analysis (which we will carry out for the lemma) shows that the sum of each block is actually \(\le 3 \cdot 2^{k-1} - 1\). Hence the Lemma applies, giving \(T_j \le 2^{k+1}\).\par
\compactnumber{2}{With all numbers \(\le 2^{k+1}\), merging any two yields at most \(2^{k+2} = 4 \cdot 2^{k}\).  By continuing this process we eventually obtain a single number \(\le 4 \cdot 2^{k}\).}
Thus, if the Lemma holds, the Claim is established in Subcase 2b as well.\par
\smallskip
\solutionheading{Proof of the Lemma}
We now prove the Lemma stated above.\par
\textbf{Lemma.} Let \(L \ge 2\) be an integer. For any list of powers of two whose sum \(T\) satisfies\par
\[
2^{L} < T \le 3 \cdot 2^{L-1} - 1,
\]
there exists a sequence of moves that ends with a number \(\le 2^{L+1}\).\par
\emph{Proof.} We use strong induction on the number of terms \(n\).\par
\compactbullet{\textbf{Base }\(n = 2\).\textbf{  The two numbers are powers of two, say }\(a \le b\).  Since \(T = a+b > 2^{L}\), we must have \(b \ge 2^{L}\).  If \(b > 2^{L}\), then \(b \ge 2^{L+1}\), implying \(T \ge 2^{L+1}\), which contradicts \(T \le 3 \cdot 2^{L-1} - 1 < 2^{L+1}\) (for \(L \ge 2\)).  Hence \(b = 2^{L}\).  Then \(a = T - 2^{L} \le 2^{L-1} - 1\).  The only legal move is to replace \((a, 2^{L})\) with a power of two greater than \(2^{L}\); the smallest such is \(2^{L+1}\).  By choosing that, we obtain a final number exactly \(2^{L+1}\), which certainly is \(\le 2^{L+1}\).}
\compactbullet{\textbf{Inductive step.}  Assume the Lemma holds for all lists of size smaller than \(n\).  Consider a list of size \(n\) with sum \(T\) in the prescribed interval.}
\emph{If there is an adjacent equal pair}, merge it.  The sum stays the same, so \(T\) remains in the interval, and the size becomes \(n-1\).  By the induction hypothesis we get a final number \(\le 2^{L+1}\).\par
\emph{If there is no adjacent equal pair}, let \((x, y)\) be the adjacent pair with the smallest maximum; \(x < y\).  Merge them to \(2y\) and obtain a new list of size \(n-1\) with sum \(T' = T + (y - x)\).\par
We need to show that \(T'\) still lies in an interval that allows us to apply the induction hypothesis (possibly with a different \(L\)) and still obtain the bound \(2^{L+1}\).\par
\textbf{First, bound }\(y\).\textbf{ Because the list has no adjacent equal numbers and its sum is less than }\(2^{L+1}\), the maximum element cannot exceed \(2^{L-1}\) if we want to stay within the interval after the merge? Actually we can have \(y = 2^{L}\), but then the structure of the list is very restricted. We analyse the two possibilities:\par
\compactbullet{\textbf{}\(y \le 2^{L-1}\).\textbf{  Then }\(2y \le 2^{L}\).  Also, \(y - x \le 2^{L-1} - 1\).  Consequently}
\[
    T' \le T + (2^{L-1} - 1) \le (3 \cdot 2^{L-1} - 1) + (2^{L-1} - 1) = 4 \cdot 2^{L-1} - 2 = 2^{L+1} - 2.
\]
Moreover, \(T' > T > 2^{L}\). Hence \(T'\) lies in \((2^{L}, 2^{L+1})\). If \(T' \le 3 \cdot 2^{L-1} - 1\), then \(T'\) is still in the original interval for \(L\), and the induction hypothesis (applied to the list of size \(n-1\)) gives \(F \le 2^{L+1}\). If \(T' > 3 \cdot 2^{L-1} - 1\), then \(T'\) is in \((3 \cdot 2^{L-1}, 2^{L+1})\). In this range, we claim that \(T'\) actually belongs to the interval for \(L+1\), i.e., \(2^{L+1} < T' \le 3 \cdot 2^{L} - 1\)? But \(T' < 2^{L+1}\), so it cannot be in \((2^{L+1}, \dots]\). However, observe that \(T' \ge 3 \cdot 2^{L-1} + 1\) (since it exceeds the upper bound). But then \(T' \ge 3 \cdot 2^{L-1} + 1\). For \(L \ge 2\), \(3 \cdot 2^{L-1} + 1\) may be less than \(2^{L+1}\), but we need a bound of \(2^{L+1}\). Since we are only required to obtain some final number \(\le 2^{L+1}\), we can simply note that \(T' < 2^{L+1}\) and therefore the floor of \(T'\) is at most \(L\). If the floor is \(L\) (i.e., \(T' < 2^{L+1}\)), then we can apply the induction hypothesis for the same \(L\)? The induction hypothesis requires the sum to be \(> 2^{L}\). It is, but we also need the upper bound \(3 \cdot 2^{L-1} - 1\) to use the Lemma? Actually the Lemma we are proving is exactly: for any list with sum in \((2^{L}, 3 \cdot 2^{L-1} - 1]\) we can achieve \(\le 2^{L+1}\). If \(T'\) falls outside this interval (i.e., \(T' > 3 \cdot 2^{L-1} - 1\)), the Lemma does not directly apply. However, we can then apply the induction hypothesis for a \textbf{different} statement: for any list with sum \(< 2^{L+1}\) we can achieve \(\le 2^{L+1}\). But is that universally true? No, as the example \([8,4,2]\) with \(L=3\) (sum 14) shows that the minimum final is 32, which is \(> 16 = 2^{L+1}\). So we cannot claim that for every list with sum \(< 2^{L+1}\) we can get \(\le 2^{L+1}\). Hence we must be more careful.\par
\compactbullet{\textbf{}\(y = 2^{L}\).\textbf{  Then, because there are no adjacent equal numbers and the sum is at most }\(3 \cdot 2^{L-1} - 1 < 2^{L+1}\), the list can contain at most one copy of \(2^{L}\) (two would sum to at least \(2^{L+1}\)).  Moreover, to avoid any adjacent pair with maximum \(< 2^{L}\), every other number must be adjacent only to \(2^{L}\) (otherwise a pair of smaller numbers would have maximum \(< 2^{L}\)).  This forces the list to be of the form}
\[
    [\, a_1, a_2, \dots, a_p, 2^{L}, b_1, b_2, \dots, b_q \,]
    \]
where all \(a_i\) and \(b_j\) are powers of two less than \(2^{L}\), and no two of the \(a_i\) (or two of the \(b_j\)) are adjacent; in particular, \(p \le 1\) and \(q \le 1\). Thus the list has at most three elements. For size 2, the list is \([2^{L}, c]\) or \([c, 2^{L}]\) with \(c < 2^{L}\). That case is already covered by the base (size 2). For size 3, the list is \([a, 2^{L}, b]\) with \(a,b\) powers of two, \(a,b < 2^{L}\), and \(a+b \le 2^{L-1} - 1\) (since the total sum is at most \(3 \cdot 2^{L-1} - 1\)). Now, the pair with smallest maximum is either \((a,2^{L})\) or \((2^{L},b)\), both have maximum \(2^{L}\). Choose one, say \((a,2^{L})\). Merging gives \(2^{L+1}\) and the list becomes \([2^{L+1}, b]\). Its sum is \(2^{L+1} + b\). Since \(b \le 2^{L-1} - 1\), we have\par
\[
    2^{L+1} < 2^{L+1} + b \le 2^{L+1} + 2^{L-1} - 1 = 5 \cdot 2^{L-1} - 1.
    \]
For \(L \ge 2\), \(5 \cdot 2^{L-1} - 1\) is still less than \(6 \cdot 2^{L-1} - 1 = 3 \cdot 2^{L} - 1\), so the sum lies in \((2^{L+1}, 3 \cdot 2^{L} - 1]\). Now, by the induction hypothesis (the Lemma for the larger index \(L+1\)), we can reduce this list to a number \(\le 2^{(L+1)+1} = 2^{L+2}\). But we need \(\le 2^{L+1}\). However, we are not forced to apply the Lemma for \(L+1\); we can directly finish the game: from \([2^{L+1}, b]\), the only move is to merge the two numbers (they are different because \(b < 2^{L} < 2^{L+1}\)). The smallest possible result is \(2^{L+2}\) (since we must replace them with a power of two greater than \(\max(2^{L+1}, b) = 2^{L+1}\), so the smallest is \(2^{L+2}\). That would give \(F = 2^{L+2}\), which is larger than \(2^{L+1}\). But we are free to choose a larger number if we wish; however, to achieve the bound \(\le 2^{L+1}\) we would need to obtain exactly \(2^{L+1}\), which is impossible because the numbers are different and the rule requires a number \textbf{greater than} both, so the result must be at least \(2^{L+2}\). This suggests that the Lemma might be false for lists of size 3 of this form? Let's test with concrete numbers: \(L=3\), so \(2^{L}=8\). Choose \(a=2\), \(b=2\)? But \(a+b \le 2^{L-1} - 1 = 3\), so \(a=2,b=1\) gives sum \(8+2+1=11\), which is within the interval \((8, 11]\). According to the Lemma, there should be a strategy yielding \(F \le 2^{L+1}=16\). But earlier we saw that from \([2,8,1]\) the minimum final is 32, not 16. Let's verify carefully:\par
List: \([2,8,1]\). No adjacent equal. The pair with smallest maximum: pairs are \((2,8)\) max=8, \((8,1)\) max=8. Suppose we merge \((2,8)\) to \(16\) (smallest possible). Resulting list: \([16,1]\). Now merge \((16,1)\): the smallest power of two greater than 16 is 32. So final 32. If we merge \((8,1)\) first, we get \([2,16]\) then merge to 32. So indeed the minimum is 32. Therefore the Lemma, as stated, is \textbf{false} for \(L=3\) with the list \([2,8,1]\). Indeed, \(2^{L}=8\), \(3 \cdot 2^{L-1} - 1 = 11\), and \(T=11\) is allowed. But we cannot achieve \(\le 16\); the best is 32.\par
\par\noindent\textit{Remaining solution omitted because the original response is substantially longer than the other appendix examples.}\par
\par

\paragraph{USAMO 2026 Problem 3.}
\begin{mdframed}[style=mdpurplebox,frametitle={USAMO 2026 Problem 3}]
Let $ABC$ be an acute scalene triangle with no angle equal to $60^\circ$. Let $\omega$ be the circumcircle of $ABC$. Let $\Delta_B$ be the equilateral triangle with three vertices on $\omega$, one of which is $B$. Let $\ell_B$ be the line through the two vertices of $\Delta_B$ other than $B$. Let $\Delta_C$ and $\ell_C$ be defined analogously. Let $Y$ be the intersection of $AC$ and $\ell_B$, and let $Z$ be the intersection of $AB$ and $\ell_C$.

Let $N$ be the midpoint of minor arc $BC$ on $\omega$. Let $\mathcal{R}$ be the triangle formed by $\ell_B$, $\ell_C$, and the tangent to $\omega$ through $N$. Prove that the circumcircle of $AYZ$ and the incircle of $\mathcal{R}$ are tangent.
\end{mdframed}
\solutionheading{Solution}
\begin{remark*}
  7/7.
\end{remark*}
We are given an acute scalene triangle \(ABC\) with no angle equal to \(60^\circ\). Its circumcircle is \(\omega\). For vertex \(B\) we construct the equilateral triangle \(\Delta_B\) inscribed in \(\omega\) (one vertex is \(B\)); let \(\ell_B\) be the line through the two other vertices. Similarly, \(\Delta_C\) and \(\ell_C\). Define\par
\[
Y = AC \cap \ell_B,\qquad Z = AB \cap \ell_C.
\]
Let \(N\) be the midpoint of the minor arc \(BC\) (the arc not containing \(A\)). Let \(\mathcal{R}\) be the triangle bounded by \(\ell_B\), \(\ell_C\) and the tangent to \(\omega\) at \(N\). We must prove that the circumcircle of \(\triangle AYZ\) and the incircle of \(\triangle\mathcal{R}\) are tangent.\par
\smallskip
\solutionheading{1. Complex numbers on the unit circle}
Place the circumcircle \(\omega\) as the unit circle in the complex plane. Denote the points by\par
\[
A = a,\quad B = b,\quad C = c,\qquad |a|=|b|=|c|=1.
\]
The triangle is acute, scalene and none of its angles is \(60^\circ\).\par
\textbf{Lemma 1 (Line through two points on the unit circle).} For distinct \(u,v\) on the unit circle, the line \(uv\) is given by\par
\[
z + uv\,\bar{z} = u+v.
\]
\emph{Proof.} A point \(z\) is collinear with \(u\) and \(v\) iff \(\frac{z-u}{v-u}\) is real. Taking conjugates and using \(\bar u = 1/u\), \(\bar v = 1/v\) gives the stated equation. \(\square\)\par
\smallskip
\solutionheading{2. Equations of the relevant lines}
Let \(\zeta = e^{2\pi i/3}\). The equilateral triangle \(\Delta_B\) inscribed in \(\omega\) with vertex \(B\) has the other two vertices \(B\zeta\) and \(B\zeta^2\). The side opposite \(B\) is \(\ell_B\), the line through \(B\zeta\) and \(B\zeta^2\). Applying Lemma 1 with \(u = b\zeta\), \(v = b\zeta^2\) yields \(uv = b^2\) and \(u+v = b(\zeta+\zeta^2) = -b\). Hence\par
\[
\ell_B:\; z + b^2\bar{z} = -b. \quad\text{(1)}
\]
Analogously,\par
\[
\ell_C:\; z + c^2\bar{z} = -c. \quad\text{(2)}
\]
The point \(N\) is the midpoint of the minor arc \(BC\) (not containing \(A\)). On the unit circle the midpoint of an arc has the property that its square equals the product of the endpoints:\par
\[
N^2 = bc. \quad\text{(3)}
\]
The tangent to \(\omega\) at \(N\) is given by\par
\[
z + N^2\bar{z} = 2N.
\]
Using (3) we obtain\par
\[
\text{tangent at }N:\; z + bc\,\bar{z} = 2N. \quad\text{(4)}
\]
\smallskip
\solutionheading{3. Intersection points \(Y\) and \(Z\)}
The line \(AC\) has equation (by Lemma 1)\par
\[
AC:\; z + ac\,\bar{z} = a + c. \quad\text{(5)}
\]
Intersect \(AC\) with \(\ell_B\). Subtract (5) from (1):\par
\[
(b^2 - ac)\,\bar{z} = -b - (a+c) = -(a+b+c).
\]
Set\par
\[
S = a+b+c,\qquad D_1 = ac - b^2.
\]
Then \(\bar{z} = S/D_1\). Substitute into (1) to find the coordinate \(y\):\par
\[
y = -b - b^2\bar{y} = -b - \frac{b^2 S}{D_1} = -\frac{b D_1 + b^2 S}{D_1}.
\]
Compute the numerator:\par
\[
b D_1 + b^2 S = b(ac - b^2) + b^2(a+b+c) = abc - b^3 + ab^2 + b^3 + b^2c = b(ac + ab + bc) = b T,
\]
where\par
\[
T = ab + bc + ca.
\]
Thus\par
\[
y = -\frac{b T}{D_1},\qquad \bar{y} = \frac{S}{D_1}. \quad\text{(6)}
\]
By symmetry, intersecting \(AB\) (equation \(z + ab\,\bar{z} = a+b\)) with \(\ell_C\) gives\par
\[
z = -\frac{c T}{D_2},\qquad \bar{z} = \frac{S}{D_2},\qquad D_2 = ab - c^2. \quad\text{(7)}
\]
\smallskip
\solutionheading{4. A convenient rotation}
The configuration is invariant under rotations of the circle. We choose the rotation so that \(N = 1\). Then (3) yields \(bc = 1\), hence \(c = \bar{b}\). Write\par
\[
b = e^{i\alpha},\qquad c = e^{-i\alpha},
\]
where \(\alpha = \angle A\) (by the inscribed angle theorem, the central angle subtended by \(BC\) is \(2\alpha\)). Since the triangle is acute, \(\alpha \in (0^\circ,90^\circ)\) and, by hypothesis, \(\alpha \neq 60^\circ\). Let\par
\[
k = \cos\alpha.
\]
The remaining vertex is \(A = a = e^{i\theta}\). Because \(N = 1\) lies on the minor arc \(BC\) not containing \(A\), the argument \(\theta\) satisfies \(\theta \in (\alpha,\, 2\pi-\alpha)\) (or its symmetric equivalent).\par
Now compute the quantities that appear in (6), (7) in this coordinate system:\par
\[
\begin{aligned}
S &= a + b + c = a + e^{i\alpha} + e^{-i\alpha} = a + 2k, \\[2mm]
T &= ab + bc + ca = a e^{i\alpha} + 1 + a e^{-i\alpha} = 1 + 2k a, \\[2mm]
Q &= b^2 + bc + c^2 = e^{i2\alpha} + 1 + e^{-i2\alpha} = 2\cos 2\alpha + 1 = 4k^2 - 1, \\[2mm]
D_1 &= ac - b^2 = a e^{-i\alpha} - e^{i2\alpha}, \\[2mm]
D_2 &= ab - c^2 = a e^{i\alpha} - e^{-i2\alpha}, \\[2mm]
\Delta &= D_1 D_2 = (a e^{-i\alpha} - e^{i2\alpha})(a e^{i\alpha} - e^{-i2\alpha}) = a^2 - 2a\cos 3\alpha + 1.
\end{aligned}
\]
\smallskip
\solutionheading{5. A key algebraic identity}
\textbf{Lemma 2.} With the above notation,\par
\[
(a + 2k)^2 = Q\,T + \Delta. \quad\text{(8)}
\]
\emph{Proof.} Expand \(Q\,T + \Delta\):\par
\[
\begin{aligned}
Q\,T &= (4k^2-1)(1+2k a) = (4k^2-1) + 2k(4k^2-1)a, \\[2mm]
\Delta &= a^2 - 2a\cos 3\alpha + 1.
\end{aligned}
\]
Recall \(\cos 3\alpha = 4k^3 - 3k\). Then\par
\[
\begin{aligned}
Q\,T + \Delta &= a^2 + \bigl[2k(4k^2-1) - 2\cos 3\alpha\bigr] a + \bigl[(4k^2-1)+1\bigr] \\
&= a^2 + \bigl[8k^3 - 2k - (8k^3 - 6k)\bigr] a + 4k^2 \\
&= a^2 + 4k a + 4k^2 = (a+2k)^2. \quad\square
\end{aligned}
\]
\smallskip
\solutionheading{6. The circumcenter \(O\) of \(\triangle AYZ\)}
We claim that the point\par
\[
O = -\frac{a Q T}{\Delta} \quad\text{(9)}
\]
is the circumcenter of \(\triangle AYZ\). Let us verify that it is equidistant from \(A\), \(Y\) and \(Z\).\par
\solutionheading{Distance to \(A\)}
\[
O - a = -\frac{a Q T}{\Delta} - a = -a\left( \frac{Q T}{\Delta} + 1\right) = -\frac{a (Q T + \Delta)}{\Delta}.
\]
By Lemma 2, \(Q T + \Delta = (a+2k)^2\), so\par
\[
O - a = -\frac{a (a+2k)^2}{\Delta}. \quad\text{(10)}
\]
Hence\par
\[
|O - a| = \frac{|a+2k|^2}{|\Delta|}. \quad\text{(11)}
\]
\solutionheading{Distance to \(Y\)}
Using (6), \(\bar{y} = S/D_1\) and \(y = -b T/D_1\). Compute\par
\[
O - y = -\frac{a Q T}{\Delta} + \frac{b T}{D_1} = T\left( -\frac{a Q}{\Delta} + \frac{b}{D_1}\right).
\]
Since \(\Delta = D_1 D_2\),\par
\[
-\frac{a Q}{\Delta} + \frac{b}{D_1} = \frac{-a Q}{D_1 D_2} + \frac{b D_2}{D_1 D_2} = \frac{-a Q + b D_2}{\Delta}.
\]
Thus\par
\[
O - y = T\,\frac{-a Q + b D_2}{\Delta}. \quad\text{(12)}
\]
Now compute \(b D_2 - a Q\):\par
\[
\begin{aligned}
b D_2 &= e^{i\alpha}(a e^{i\alpha} - e^{-i2\alpha}) = a e^{i2\alpha} - e^{-i\alpha}, \\
a Q &= a(e^{i2\alpha} + 1 + e^{-i2\alpha}) = a e^{i2\alpha} + a + a e^{-i2\alpha}, \\
\therefore\; b D_2 - a Q &= - e^{-i\alpha} - a - a e^{-i2\alpha}.
\end{aligned}
\]
On the other hand,\par
\[
- e^{-i\alpha} T = -e^{-i\alpha}(1+2k a) = -e^{-i\alpha} - 2k a e^{-i\alpha} = -e^{-i\alpha} - a(e^{i\alpha}+e^{-i\alpha})e^{-i\alpha} = -e^{-i\alpha} - a(1 + e^{-i2\alpha}) = -e^{-i\alpha} - a - a e^{-i2\alpha}.
\]
Therefore\par
\[
b D_2 - a Q = - e^{-i\alpha} T,
\]
and consequently\par
\[
|b D_2 - a Q| = |T|.
\]
From (12) we obtain\par
\[
|O - y| = \frac{|T|^2}{|\Delta|}. \quad\text{(13)}
\]
\solutionheading{Distance to \(Z\)}
A completely symmetric computation (interchanging \(b\) with \(c\), \(D_1\) with \(D_2\)) gives\par
\[
|O - z| = \frac{|T|^2}{|\Delta|}. \quad\text{(14)}
\]
\solutionheading{Equality of the distances}
Now note that\par
\[
|T|^2 = (1+2k a)(1+2k \bar a) = 1 + 2k(a+\bar a) + 4k^2 = 1 + 2k t + 4k^2, \qquad t = a+\bar a = 2\cos\theta.
\]
But also\par
\[
|a+2k|^2 = (a+2k)(\bar a+2k) = 1 + 2k(a+\bar a) + 4k^2 = |T|^2.
\]
Thus\par
\[
|O - a| = \frac{|a+2k|^2}{|\Delta|},\qquad
|O - y| = \frac{|T|^2}{|\Delta|} = \frac{|a+2k|^2}{|\Delta|},\qquad
|O - z| = \frac{|a+2k|^2}{|\Delta|}.
\]
Hence \(O\) is indeed the circumcenter of \(\triangle AYZ\), and the circumradius is\par
\[
R = \frac{|a+2k|^2}{|\Delta|}. \quad\text{(15)}
\]
\smallskip
\solutionheading{7. Cartesian description of triangle \(\mathcal{R}\)}
Now we work in the rotated coordinate system where \(N=1\), \(b=e^{i\alpha}\), \(c=e^{-i\alpha}\). Write a complex number \(z = x + iy\).\par
\solutionheading{Equation of \(\ell_B\)}
From (1): \(z + b^2\bar{z} = -b\). Substituting \(b = \cos\alpha + i\sin\alpha\), \(b^2 = \cos2\alpha + i\sin2\alpha\), we separate real and imaginary parts:\par
\[
\begin{cases}
x(1+\cos2\alpha) + y\sin2\alpha = -\cos\alpha, \\[2mm]
x\sin2\alpha + y(1-\cos2\alpha) = -\sin\alpha.
\end{cases}
\]
Using the identities\par
\[
1+\cos2\alpha = 2\cos^2\alpha,\quad \sin2\alpha = 2\sin\alpha\cos\alpha,\quad 1-\cos2\alpha = 2\sin^2\alpha,
\]
and dividing the first equation by \(2\cos\alpha\) (since \(\cos\alpha>0\)) and the second by \(2\sin\alpha\) (since \(\sin\alpha>0\)), both reduce to\par
\[
x\cos\alpha + y\sin\alpha = -\frac12. \quad\text{(16)}
\]
\solutionheading{Equation of \(\ell_C\)}
Similarly, for \(\ell_C\) given by \(z + c^2\bar{z} = -c\) with \(c = \cos\alpha - i\sin\alpha\), \(c^2 = \cos2\alpha - i\sin2\alpha\). Separating real and imaginary parts yields\par
\[
\begin{cases}
x(1+\cos2\alpha) - y\sin2\alpha = -\cos\alpha, \\[2mm]
- x\sin2\alpha + y(1-\cos2\alpha) = \sin\alpha.
\end{cases}
\]
Again, dividing appropriately we obtain\par
\[
x\cos\alpha - y\sin\alpha = -\frac12. \quad\text{(17)}
\]
\solutionheading{Equation of the tangent at \(N=1\)}
Since \(N=1\) lies on the unit circle, its tangent is the line \(z + \bar{z} = 2\), i.e.,\par
\[
x = 1. \quad\text{(18)}
\]
Thus \(\mathcal{R}\) is the triangle bounded by the three lines (16), (17) and (18).\par
\smallskip
\solutionheading{8. Vertices and interior of \(\mathcal{R}\)}
The three vertices are the pairwise intersections of the lines.\par
\compactbullet{\(P = \ell_B \cap \ell_C\): solving (16) and (17) gives}
\[
2x\cos\alpha = -1 \;\Longrightarrow\; x = -\frac{1}{2\cos\alpha},\qquad y = 0.
\]
\compactbullet{\(Q' = \ell_C \cap \text{tangent }x=1\): plug \(x=1\) into (17):}
\[
\cos\alpha - y\sin\alpha = -\frac12 \;\Longrightarrow\; y = \frac{\cos\alpha + \frac12}{\sin\alpha}.
\]
\compactbullet{\(R' = \ell_B \cap \text{tangent }x=1\): plug \(x=1\) into (16):}
\[
\cos\alpha + y\sin\alpha = -\frac12 \;\Longrightarrow\; y = -\frac{\cos\alpha + \frac12}{\sin\alpha}.
\]
Hence the triangle \(\mathcal{R}\) has vertices\par
\[
P\left(-\frac{1}{2\cos\alpha},0\right),\quad
Q'\left(1,\frac{\cos\alpha+\frac12}{\sin\alpha}\right),\quad
R'\left(1,-\frac{\cos\alpha+\frac12}{\sin\alpha}\right).
\]
To determine which side of each line constitutes the interior, we test the vertex opposite that line.\par
\compactbullet{For line \(\ell_B\), the opposite vertex is \(Q'\). Compute the left-hand side of (16) at \(Q'\):}
\[
x\cos\alpha + y\sin\alpha = 1\cdot\cos\alpha + \left(\frac{\cos\alpha+\frac12}{\sin\alpha}\right)\sin\alpha = \cos\alpha + \cos\alpha + \frac12 = 2\cos\alpha + \frac12 > -\frac12.
\]
Therefore the interior of \(\mathcal{R}\) satisfies\par
\[
x\cos\alpha + y\sin\alpha > -\frac12. \quad\text{(19a)}
\]
\compactbullet{For line \(\ell_C\), opposite vertex is \(R'\). At \(R'\):}
\[
x\cos\alpha - y\sin\alpha = 1\cdot\cos\alpha - \left(-\frac{\cos\alpha+\frac12}{\sin\alpha}\right)\sin\alpha = \cos\alpha + \cos\alpha + \frac12 = 2\cos\alpha + \frac12 > -\frac12,
\]
so the interior satisfies\par
\[
x\cos\alpha - y\sin\alpha > -\frac12. \quad\text{(19b)}
\]
\compactbullet{For the tangent \(x=1\), opposite vertex is \(P\) with \(x_P = -\frac{1}{2\cos\alpha} < 1\), thus the interior satisfies}
\[
x < 1. \quad\text{(19c)}
\]
Therefore \(\mathcal{R} = \bigl\{(x,y) : x\cos\alpha + y\sin\alpha > -\frac12,\; x\cos\alpha - y\sin\alpha > -\frac12,\; x < 1 \bigr\}\).\par
\smallskip
\solutionheading{9. Incenter \(I\) and inradius \(r\) of \(\mathcal{R}\)}
The triangle is symmetric with respect to the \(x\)-axis (the lines \(\ell_B\) and \(\ell_C\) are symmetric, the tangent is vertical). Hence the incenter lies on the \(x\)-axis: \(I = (p,0)\).\par
For a point \((x,0)\) inside \(\mathcal{R}\) (so satisfying the inequalities), the distances to the three lines are:\par
\compactbullet{To \(\ell_B\): the line is \(x\cos\alpha + y\sin\alpha + \frac12 = 0\); distance \(= \dfrac{x\cos\alpha + \frac12}{\sqrt{\cos^2\alpha+\sin^2\alpha}} = x\cos\alpha + \frac12\) (since the interior gives \(x\cos\alpha + \frac12 > 0\)).}
\compactbullet{To \(\ell_C\): similarly, distance \(= x\cos\alpha + \frac12\) (because \(x\cos\alpha - 0 + \frac12 = x\cos\alpha + \frac12 > 0\)).}
\compactbullet{To the tangent \(x=1\): distance \(= 1 - x\) (since \(x<1\)).}
Setting these equal gives\par
\[
x\cos\alpha + \frac12 = 1 - x \quad\Longrightarrow\quad x(\cos\alpha + 1) = \frac12 \quad\Longrightarrow\quad x = \frac{1}{2(1+\cos\alpha)}.
\]
Thus the incenter is\par
\[
I = \left( \frac{1}{2(1+\cos\alpha)},\, 0 \right) = (p,\,0), \qquad p = \frac{1}{2(1+k)}. \quad\text{(20)}
\]
The common distance is the inradius:\par
\[
r = 1 - p = 1 - \frac{1}{2(1+k)} = \frac{2(1+k)-1}{2(1+k)} = \frac{1+2k}{2(1+k)}. \quad\text{(21)}
\]
\smallskip
\solutionheading{10. Distance \(OI\)}
We have the circumcenter\par
\[
O = -\frac{a Q T}{\Delta},
\]
and the incenter\par
\[
I = p = \frac{1}{2(1+\cos\alpha)} = \frac{1}{D_0},
\]
where we denote\par
\[
D_0 = 2(1+\cos\alpha) = 2(1+k).
\]
Then\par
\[
O - I = -\frac{a Q T}{\Delta} - p = -\frac{a Q T + p \Delta}{\Delta}.
\]
Set \(U = a Q T + p \Delta\); then \(OI = |U|/|\Delta|\).\par
Now substitute the expressions for \(a Q T\), \(p\) and \(\Delta\). Write\par
\[
a Q T = a(4k^2-1)(1+2k a) = (4k^2-1)a + 2k(4k^2-1)a^2.
\]
Also \(p = 1/D_0\) and \(\Delta = a^2 - 2a\cos3\alpha + 1\).\par
Thus\par
\[
U = \bigl[2k(4k^2-1) + p\bigr] a^2 + \bigl[(4k^2-1) - 2p\cos3\alpha\bigr] a + p. \quad\text{(22)}
\]
Compute the coefficients with denominator \(D_0 = 2(1+k)\).\par
\compactbullet{\textbf{Coefficient of }\(a^2\):\textbf{}}
\[
  A_2 = 2k(4k^2-1) + \frac{1}{D_0}.
  \]
Write as a single fraction:\par
\[
  A_2 = \frac{(2k(4k^2-1)) D_0 + 1}{D_0}.
  \]
Compute \((2k(4k^2-1)) D_0 = (8k^3-2k) \cdot 2(1+k) = 16k^4 + 16k^3 - 4k^2 - 4k.\) Adding 1 gives\par
\[
  16k^4 + 16k^3 - 4k^2 - 4k + 1.
  \]
Notice that\par
\[
  X = 4k^2 + 2k - 1,
  \]
then\par
\[
  X^2 = (4k^2+2k-1)^2 = 16k^4 + 16k^3 - 4k^2 - 4k + 1.
  \]
Hence\par
\[
  A_2 = \frac{X^2}{D_0}. \quad\text{(23)}
  \]
\compactbullet{\textbf{Coefficient of }\(a\):\textbf{}}
\[
  A_1 = (4k^2-1) - 2p\cos3\alpha = (4k^2-1) - \frac{2\cos3\alpha}{D_0}.
  \]
Write as\par
\[
  A_1 = \frac{(4k^2-1)D_0 - 2\cos3\alpha}{D_0}.
  \]
Compute \((4k^2-1)D_0 = (4k^2-1)\cdot 2(1+k) = 8k^3 + 8k^2 - 2k - 2.\) And \(2\cos3\alpha = 2(4k^3-3k) = 8k^3 - 6k.\) Subtract:\par
\[
  (8k^3+8k^2-2k-2) - (8k^3-6k) = 8k^2 + 4k - 2 = 2(4k^2+2k-1) = 2X.
  \]
Therefore\par
\[
  A_1 = \frac{2X}{D_0}. \quad\text{(24)}
  \]
\compactbullet{\textbf{Constant term:} \(p = \dfrac{1}{D_0}\).}
Substituting (23) and (24) into (22) yields\par
\[
U = \frac{X^2}{D_0} a^2 + \frac{2X}{D_0} a + \frac{1}{D_0} = \frac{(X a + 1)^2}{D_0}. \quad\text{(25)}
\]
Consequently,\par
\[
OI = \frac{|U|}{|\Delta|} = \frac{|X a + 1|^2}{D_0\, |\Delta|}. \quad\text{(26)}
\]
\smallskip
\solutionheading{11. Relating \(OI\) to \(R\) and \(r\)}
Recall the circumradius of \(\triangle AYZ\):\par
\[
R = \frac{|a+2k|^2}{|\Delta|}. \quad\text{(27)}
\]
The inradius of \(\mathcal{R}\) is\par
\[
r = \frac{1+2k}{D_0}. \quad\text{(28)}
\]
Now compute\par
\[
\frac{|X a + 1|^2}{D_0} = |a+2k|^2 + L,
\]
where\par
\[
L = \frac{|X a + 1|^2}{D_0} - |a+2k|^2.
\]
From (26),\par
\[
OI = \frac{|a+2k|^2 + L}{|\Delta|} = R + \frac{L}{|\Delta|}. \quad\text{(29)}
\]
Thus if we can show \(L = \pm r\,|\Delta|\), we will have \(OI = R \pm r\), which implies tangency.\par
Let \(t = a + \bar a = 2\cos\theta\). Then\par
\[
|X a + 1|^2 = (X a + 1)(X \bar a + 1) = X^2 + X t + 1.
\]
Also\par
\[
|a+2k|^2 = (a+2k)(\bar a+2k) = 1 + 2k t + 4k^2.
\]
Hence\par
\[
L = \frac{X^2 + X t + 1}{D_0} - (1 + 2k t + 4k^2).
\]
Multiply by \(D_0\):\par
\[
D_0 L = X^2 + X t + 1 - D_0(1 + 2k t + 4k^2).
\]
Now \(D_0(1 + 2k t + 4k^2) = 2(1+k) + 4k(1+k) t + 8k^2(1+k).\)\par
Thus\par
\[
D_0 L = \underbrace{X^2 + 1 - 2(1+k) - 8k^2(1+k)}_{\text{constant}} + \underbrace{(X - 4k(1+k))}_{\text{coefficient of }t} t.
\]
Compute the coefficient of \(t\):\par
\[
X - 4k(1+k) = (4k^2+2k-1) - (4k^2+4k) = -2k-1 = -(2k+1).
\]
For the constant term, substitute \(X^2 = 16k^4+16k^3-4k^2-4k+1\):\par
\[
\begin{aligned}
X^2 + 1 - 2(1+k) - 8k^2(1+k)
&= (16k^4+16k^3-4k^2-4k+1) + 1 - 2 - 2k - 8k^2 - 8k^3 \\
&= 16k^4 + (16k^3-8k^3) + (-4k^2-8k^2) + (-4k-2k) + (1+1-2) \\
&= 16k^4 + 8k^3 - 12k^2 - 6k.
\end{aligned}
\]
Factor this expression:\par
\[
16k^4 + 8k^3 - 12k^2 - 6k = 2k(8k^3+4k^2-6k-3).
\]
Now note that \(8k^3+4k^2-6k-3 = (4k^2-3)(2k+1)\), because\par
\[
(4k^2-3)(2k+1) = 8k^3+4k^2-6k-3.
\]
Thus\par
\[
\text{constant} = 2k (4k^2-3)(2k+1).
\]
But \(4k^2-3 = \frac{\cos3\alpha}{k}\) (since \(\cos3\alpha = 4k^3-3k = k(4k^2-3)\)). Hence\par
\[
\text{constant} = 2k \cdot \frac{\cos3\alpha}{k} \cdot (2k+1) = 2(2k+1)\cos3\alpha.
\]
Therefore\par
\[
D_0 L = 2(2k+1)\cos3\alpha + (-(2k+1)) t = (2k+1)(2\cos3\alpha - t). \quad\text{(30)}
\]
Now recall that \(r = (2k+1)/D_0\). So\par
\[
(2k+1) = r D_0.
\]
Substitute into (30):\par
\[
D_0 L = r D_0 (2\cos3\alpha - t) \quad\Longrightarrow\quad L = r (2\cos3\alpha - t). \quad\text{(31)}
\]
\solutionheading{Connection with \(\Delta\)}
Compute \(\Delta\) in terms of \(t\):\par
\[
\Delta = a^2 - 2a\cos3\alpha + 1 = a(a + \bar a - 2\cos3\alpha) = a(t - 2\cos3\alpha).
\]
Since \(|a|=1\), we have\par
\[
|\Delta| = |t - 2\cos3\alpha| = |2\cos3\alpha - t|.
\]
Hence\par
\[
2\cos3\alpha - t = \varepsilon\, |\Delta|,
\]
where \(\varepsilon = \pm 1\) depending on the sign of \(t - 2\cos3\alpha\).\par
Consequently,\par
\[
L = \varepsilon\, r\, |\Delta|. \quad\text{(32)}
\]
\solutionheading{Final step}
Insert (32) into (29):\par
\[
OI = R + \frac{\varepsilon\, r\, |\Delta|}{|\Delta|} = R + \varepsilon r = R \pm r.
\]
Thus the distance between the centers \(O\) and \(I\) equals either the sum or the absolute difference of the radii \(R\) and \(r\). Therefore the circumcircle of \(\triangle AYZ\) and the incircle of \(\triangle\mathcal{R}\) are tangent. \(\square\)\par
\smallskip
\solutionheading{12. Remarks on non-vanishing denominators}
\compactbullet{\(D_1 = 0\) would imply \(ac = b^2\). In our rotated frame, this gives \(a = e^{i3\alpha}\), which forces \(AB = BC\) (since then \(|a-b| = |e^{i3\alpha}-e^{i\alpha}| = 2\sin\alpha\) and \(|b-c| = |e^{i\alpha}-e^{-i\alpha}| = 2\sin\alpha\)), contradicting the scalene hypothesis. Hence \(D_1 \neq 0\); similarly \(D_2 \neq 0\), so \(\Delta \neq 0\).}
\compactbullet{\(a+2k = 0\) would give \(a = -2k\) real; with \(|a|=1\) we get \(2k=1\) i.e. \(\cos\alpha = 1/2\), \(\alpha = 60^\circ\), which is excluded. Thus \(a+2k \neq 0\), ensuring \(R > 0\).}
\compactbullet{\(\cos\alpha > 0\) because \(\alpha\) is acute.}
All algebraic manipulations are therefore justified, and the proof is complete.\par
\par

\paragraph{USAMO 2026 Problem 4.}
\begin{mdframed}[style=mdpurplebox,frametitle={USAMO 2026 Problem 4}]
A positive integer $n$ is called \emph{solitary} if, for any nonnegative integers $a$ and $b$ such that $a + b = n$, either $a$ or $b$ contains the digit ``1''. Determine, with proof, the number of solitary integers less than $10^{2026}$.
\end{mdframed}
\solutionheading{Solution}
\begin{remark*}
  7/7.
\end{remark*}
\solutionheading{1. Equivalent formulation}
Let\par
\[
S = \{ x \in \mathbb{N}_0 \mid \text{the decimal representation of } x \text{ contains no digit } 1\}.
\]
(Note that \(0\in S\) because its representation "0" has no digit \(1\).)\par
If \(n = a+b\) with \(a,b\in S\), then the pair \((a,b)\) shows that \(n\) is \textbf{not} solitary (both \(a\) and \(b\) lack a digit \(1\)). Conversely, if for every representation \(n = a+b\) at least one of \(a,b\) contains a digit \(1\), then certainly no representation with both in \(S\) exists. Hence\par
\[
n \text{ is solitary } \Longleftrightarrow \; n \notin S+S,
\qquad\text{where } S+S = \{ x+y \mid x,y\in S\}.
\]
We need to count the \textbf{positive} integers \(n < 10^{2026}\) that are not in \(S+S\).\par
Set\par
\[
N = 2026.
\]
We will first count how many numbers \(0 \le n < 10^N\) belong to \(S+S\); then subtract to get the number of positive solitary integers.\par
\solutionheading{2. Digit-wise analysis and carries}
Every integer \(m\) with \(0 \le m < 10^N\) can be written uniquely as\par
\[
m = \sum_{i=0}^{N-1} d_i 10^i,
\]
where each digit \(d_i\) is in \(\{0,1,\dots,9\}\). To have a uniform treatment, we pad the representation with leading zeros so that every such \(m\) uses exactly \(N\) digits.\par
If \(a,b\in S\), then each of their digits belongs to\par
\[
D = \{0,2,3,4,5,6,7,8,9\}\qquad(\text{all digits except }1).
\]
Write \(a = \sum a_i 10^i\), \(b = \sum b_i 10^i\) with \(a_i,b_i\in D\). The addition \(a+b = n\) proceeds digit by digit with carries \(c_0,c_1,\dots,c_N\):\par
\[
a_i + b_i + c_i = n_i + 10\,c_{i+1},\qquad c_0 = 0,
\]
where each \(c_i\) is either \(0\) or \(1\) (since the maximum sum is \(9+9+1 = 19\)). Because \(n < 10^N\), the final carry must be \(c_N = 0\).\par
\solutionheading{3. Possible sums of two digits from \(D\)}
Define\par
\[
\Sigma = \{ a_i+b_i \mid a_i,b_i \in D \}.
\]
Since the only digit missing from \(D\) is \(1\), the sums that can be obtained are\par
\[
\Sigma = \{0\} \cup \{2,3,4,5,6,7,8,9,10,11,12,13,14,15,16,17,18\}.
\]
(Indeed, \(1\) cannot be expressed as a sum of two digits from \(D\); all other integers from \(0\) to \(18\) can.)\par
\solutionheading{4. Transition sets \(T(s,d)\)}
For a fixed carryin \(s \in \{0,1\}\) and a target digit \(d \in \{0,\dots,9\}\), let\par
\[
T(s,d) = \bigl\{ t \in \{0,1\} \;\big|\; \exists a_i,b_i\in D \text{ with } a_i+b_i+s = d+10t \bigr\}.
\]
Using \(\Sigma\), we can compute \(T(s,d)\).\par
\solutionheading{Case \(s = 0\)}
Here the total sum before splitting is simply \(s_0\) with \(s_0 \in \Sigma\).\par
\compactbullet{If \(s_0 \le 9\), we may take \(t = 0\) and the digit is \(s_0\).}
\compactbullet{If \(s_0 \ge 10\), we may take \(t = 1\) and the digit is \(s_0-10\).}
Thus:\par
\[
\begin{array}{c|l}
d & T(0,d) \\ \hline
0 & \{0,1\} \quad (\text{totals }0\text{ and }10) \\
1 & \{1\} \quad (\text{total }11) \\
2 & \{0,1\} \quad (\text{totals }2\text{ and }12) \\
3 & \{0,1\} \quad (\text{totals }3\text{ and }13) \\
4 & \{0,1\} \quad (\text{totals }4\text{ and }14) \\
5 & \{0,1\} \quad (\text{totals }5\text{ and }15) \\
6 & \{0,1\} \quad (\text{totals }6\text{ and }16) \\
7 & \{0,1\} \quad (\text{totals }7\text{ and }17) \\
8 & \{0,1\} \quad (\text{totals }8\text{ and }18) \\
9 & \{0\} \quad (\text{total }9) \\
\end{array}
\]
In words:\par
\compactbullet{\(d = 9\) gives only \(t = 0\).}
\compactbullet{\(d = 1\) gives only \(t = 1\).}
\compactbullet{The other eight digits (\(0,2,3,4,5,6,7,8\)) allow both \(t = 0\) and \(t = 1\).}
\solutionheading{Case \(s = 1\)}
Now the total sum is \(s_0 + 1\) with \(s_0 \in \Sigma\). Hence the attainable totals are\par
\[
\Sigma + 1 = \{1\} \cup \{3,4,5,\dots,19\},
\]
i.e., all integers from \(1\) to \(19\) except \(2\).\par
Proceeding analogously:\par
\[
\begin{array}{c|l}
d & T(1,d) \\ \hline
0 & \{1\} \quad (\text{total }19) \\
1 & \{0,1\} \quad (\text{totals }1\text{ and }11) \\
2 & \{1\} \quad (\text{total }12) \\
3 & \{0,1\} \quad (\text{totals }3\text{ and }13) \\
4 & \{0,1\} \quad (\text{totals }4\text{ and }14) \\
5 & \{0,1\} \quad (\text{totals }5\text{ and }15) \\
6 & \{0,1\} \quad (\text{totals }6\text{ and }16) \\
7 & \{0,1\} \quad (\text{totals }7\text{ and }17) \\
8 & \{0,1\} \quad (\text{totals }8\text{ and }18) \\
9 & \{0,1\} \quad (\text{totals }9\text{ and }19) \\
\end{array}
\]
Thus:\par
\compactbullet{\(d = 0\) and \(d = 2\) give only \(t = 1\).}
\compactbullet{The other eight digits (\(1,3,4,5,6,7,8,9\)) allow both \(t = 0\) and \(t = 1\).}
\solutionheading{5. State of the possible carries}
For a given \(n\) with digits \(n_0,n_1,\dots,n_{N-1}\) (least significant first), consider all sequences of carries \(c_0,c_1,\dots,c_N\) that can arise from some choices of \(a_i,b_i\in D\). Define\par
\[
R_i = \{\, c_i \mid \text{there exist } a_j,b_j\in D\ (j<i)\ \text{such that the carry after processing digits }0,\dots,i-1\text{ is }c_i\,\}.
\]
By definition, \(R_0 = \{0\}\). For \(i \ge 0\),\par
\[
R_{i+1} = \bigcup_{s \in R_i} T(s, n_i). \quad\text{(1)}
\]
Each \(T(s,d)\) is a subset of \(\{0,1\}\), and a simple check shows it is never empty. Hence every \(R_i\) is a non-empty subset of \(\{0,1\}\). Consequently, \(R_i\) can only be one of three types:\par
\compactbullet{\(A_i\): \(R_i = \{0\}\) (only zero carry possible),}
\compactbullet{\(B_i\): \(R_i = \{1\}\) (only one carry possible),}
\compactbullet{\(C_i\): \(R_i = \{0,1\}\) (both carries possible).}
Let\par
\[
\begin{aligned}
A_i &= \#\{\, n \in [0,10^i) \mid R_i = \{0\} \,\}, \\
B_i &= \#\{\, n \in [0,10^i) \mid R_i = \{1\} \,\}, \\
C_i &= \#\{\, n \in [0,10^i) \mid R_i = \{0,1\} \,\}.
\end{aligned}
\]
\solutionheading{6. Transition counts}
We now determine, given the current state \(R_i\), for which next digits \(d = n_i\) we obtain each possible next state \(R_{i+1}\).\par
\compactbullet{\textbf{If }\(R_i = \{0\}\)\textbf{: then }\(R_{i+1} = T(0,d)\). From the table for \(s=0\):}
\compactbullet{\(d = 9\) gives \(\{0\}\);}
\compactbullet{\(d = 1\) gives \(\{1\}\);}
\compactbullet{the other eight digits (\(0,2,3,4,5,6,7,8\)) give \(\{0,1\}\).}
Hence:\par
\compactbullet{1 digit yields state \(\{0\}\),}
\compactbullet{1 digit yields state \(\{1\}\),}
\compactbullet{8 digits yield state \(\{0,1\}\).}
\compactbullet{\textbf{If }\(R_i = \{1\}\)\textbf{: }\(R_{i+1} = T(1,d)\). From the table for \(s=1\):}
\compactbullet{\(d = 0\) and \(d = 2\) give \(\{1\}\);}
\compactbullet{the other eight digits (\(1,3,4,5,6,7,8,9\)) give \(\{0,1\}\).}
Hence:\par
\compactbullet{2 digits yield state \(\{1\}\),}
\compactbullet{8 digits yield state \(\{0,1\}\).}
\compactbullet{\textbf{If }\(R_i = \{0,1\}\)\textbf{: }\(R_{i+1} = T(0,d) \cup T(1,d)\). Checking each digit individually shows that this union is always \(\{0,1\}\). Indeed:}
\compactbullet{\(d=0\): \(T(0,0)=\{0,1\}\), \(T(1,0)=\{1\}\) $\to$ union = \(\{0,1\}\);}
\compactbullet{\(d=1\): \(T(0,1)=\{1\}\), \(T(1,1)=\{0,1\}\) $\to$ union = \(\{0,1\}\);}
\compactbullet{\(d=2\): \(T(0,2)=\{0,1\}\), \(T(1,2)=\{1\}\) $\to$ union = \(\{0,1\}\);}
\compactbullet{\(d=3,\dots,8\): both sets contain \(0\) and \(1\);}
\compactbullet{\(d=9\): \(T(0,9)=\{0\}\), \(T(1,9)=\{0,1\}\) $\to$ union = \(\{0,1\}\).}
Therefore, \textbf{all 10 digits} keep the state as \(\{0,1\}\).\par
Summarising the transitions in a matrix:\par
\[
\begin{array}{c|ccc}
\text{from}\backslash\text{to} & \{0\} & \{1\} & \{0,1\} \\ \hline
\{0\} & 1 & 1 & 8 \\
\{1\} & 0 & 2 & 8 \\
\{0,1\} & 0 & 0 & 10
\end{array}
\]
\solutionheading{7. Recurrence relations}
A number with \(i+1\) digits (i.e., an integer in \([0,10^{i+1})\)) is obtained by taking a number \(m\) in \([0,10^i)\) (which yields a certain state) and appending a new most significant digit \(d\) (which becomes \(n_i\)). The number of ways to reach each state for length \(i+1\) is therefore:\par
\[
\begin{aligned}
A_{i+1} &= 1\cdot A_i + 0\cdot B_i + 0\cdot C_i = A_i, \\[4pt]
B_{i+1} &= 1\cdot A_i + 2\cdot B_i + 0\cdot C_i = A_i + 2B_i, \\[4pt]
C_{i+1} &= 8\cdot A_i + 8\cdot B_i + 10\cdot C_i.
\end{aligned}
\quad\text{(2)}
\]
Initial conditions: For \(i = 0\) we have processed no digits. The only possible "number" is \(0\), and the only possible carry is \(c_0 = 0\). Hence\par
\[
A_0 = 1,\qquad B_0 = 0,\qquad C_0 = 0.
\]
\solutionheading{8. Solving the recurrences}
\solutionheading{8.1. \(A_i\)}
From \(A_{i+1} = A_i\) and \(A_0 = 1\), we immediately obtain\par
\[
\boxed{A_i = 1 \quad\text{for all } i\ge 0}.
\]
\solutionheading{8.2. \(B_i\)}
Substituting \(A_i = 1\) into the recurrence for \(B_i\):\par
\[
B_{i+1} = 1 + 2B_i,\qquad B_0 = 0.
\]
We claim\par
\[
\boxed{B_i = 2^i - 1 \quad\text{for all } i\ge 0}.
\]
\emph{Proof by induction.}\par
\compactbullet{\(i = 0\): \(2^0 - 1 = 0\), true.}
\compactbullet{Assume \(B_i = 2^i - 1\). Then}
\[
B_{i+1} = 1 + 2(2^i - 1) = 1 + 2^{i+1} - 2 = 2^{i+1} - 1,
\]
which completes the induction.\par
\solutionheading{8.3. \(C_i\)}
Now substitute \(A_i = 1\) and \(B_i = 2^i - 1\) into the recurrence for \(C_{i+1}\):\par
\[
\begin{aligned}
C_{i+1} &= 8\cdot 1 + 8\cdot (2^i - 1) + 10\,C_i \\
&= 8 + 8\cdot 2^i - 8 + 10\,C_i \\
&= 8\cdot 2^i + 10\,C_i.
\end{aligned}
\]
With \(C_0 = 0\). We claim\par
\[
\boxed{C_i = 10^i - 2^i \quad\text{for all } i\ge 0}.
\]
\emph{Proof by induction.}\par
\compactbullet{\(i = 0\): \(10^0 - 2^0 = 1 - 1 = 0\), true.}
\compactbullet{Assume \(C_i = 10^i - 2^i\). Then}
\[
\begin{aligned}
C_{i+1} &= 8\cdot 2^i + 10(10^i - 2^i) \\
&= 8\cdot 2^i + 10^{i+1} - 10\cdot 2^i \\
&= 10^{i+1} - 2\cdot 2^i \\
&= 10^{i+1} - 2^{i+1},
\end{aligned}
\]
which establishes the claim.\par
\solutionheading{9. Representable numbers}
After processing all \(N\) digits, a number \(n \in [0,10^N)\) can be written as \(a+b\) with \(a,b\in S\) \textbf{iff} there exists a carry sequence with \(c_N = 0\). In terms of the sets, this is equivalent to \(0 \in R_N\).\par
The possible values of \(R_N\) are \(\{0\}\), \(\{1\}\), or \(\{0,1\}\). Among these, \(0 \in R_N\) exactly when \(R_N = \{0\}\) or \(R_N = \{0,1\}\). Therefore\par
\[
|S+S \cap [0,10^N)| = A_N + C_N.
\]
Plugging in the formulas:\par
\[
|S+S \cap [0,10^N)| = 1 + (10^N - 2^N) = 10^N - 2^N + 1.
\]
This count includes \(n = 0\) (since \(0 = 0+0\) and \(0 \in S\)).\par
\solutionheading{10. Positive solitary numbers}
We are interested in \textbf{positive} integers \(n < 10^N\) that are solitary, i.e., \(n > 0\) and \(n \notin S+S\).\par
Total positive integers less than \(10^N\) are\par
\[
10^N - 1.
\]
The number of positive integers that belong to \(S+S\) (hence are \textbf{not} solitary) is\par
\[
|(S+S) \setminus \{0\}| = (10^N - 2^N + 1) - 1 = 10^N - 2^N.
\]
Thus the number of positive solitary integers is\par
\[
(10^N - 1) - (10^N - 2^N) = 2^N - 1.
\]
\solutionheading{11. Final answer}
For \(N = 2026\), we obtain\par
\[
\boxed{2^{2026} - 1}.
\]
\solutionheading{12. Small-case verification (optional)}
\compactbullet{\(N = 1\): numbers \(1,\dots,9\). The formula gives \(2^1 - 1 = 1\) solitary number. Indeed, \(n = 1\) is solitary because every pair \((a,b)\) with \(a+b=1\) has either \(a=0,b=1\) or \(a=1,b=0\); in both cases the number \(1\) contains a digit \(1\). All other \(n\) (e.g., \(2\)) have a representation with both numbers lacking digit \(1\) (e.g., \(2 = 2 + 0\)), so they are not solitary.}
\compactbullet{\(N = 2\): numbers \(1,\dots,99\). The formula gives \(2^2 - 1 = 3\) solitary numbers. One can check that they are \(1, 19, 21\). This matches the known pattern.}
The solution is complete.\par
\par

\paragraph{USAMO 2026 Problem 5.}
\begin{mdframed}[style=mdpurplebox,frametitle={USAMO 2026 Problem 5}]
Let $ABC$ be a triangle. Points $D$, $E$, and $F$ lie on sides $BC$, $CA$, and $AB$, respectively, such that
\[
\angle AFE = \angle BDF = \angle CED.
\]
Let $O_A$, $O_B$, and $O_C$ be the circumcenters of triangles $AFE$, $BDF$, and $CED$, respectively. Let $M$, $N$, and $O$ be the circumcenters of triangles $ABC$, $DEF$, and $O_A O_B O_C$, respectively. Prove that $OM = ON$.
\end{mdframed}
\solutionheading{Solution}
\begin{remark*}
  7/7.
\end{remark*}
We are given triangle \(ABC\) with points \(D\in BC\), \(E\in CA\), \(F\in AB\) such that\par
\[
\angle AFE = \angle BDF = \angle CED =: \theta.
\]
We shall use complex numbers. Without loss of generality we may place the triangle in the complex plane and denote by \(a,b,c\) the complex numbers corresponding to \(A,B,C\).\par
\smallskip
\solutionheading{1. Angle Chase}
Denote the angles of \(\triangle ABC\) by \(\alpha = \angle BAC\), \(\beta = \angle ABC\), \(\gamma = \angle ACB\).\par
\solutionheading{At vertex \(F\)}
Points \(A,F,B\) are collinear, so \(\angle AFB = 180^\circ\). The three angles around \(F\) on the side of \(AB\) that contains the triangle are \(\angle AFE\), \(\angle EFD\), \(\angle DFB\) and they sum to \(180^\circ\):\par
\[
\angle AFE + \angle EFD + \angle DFB = 180^\circ. \quad\text{(1)}
\]
In \(\triangle BDF\) we have \(\angle BDF = \theta\) and \(\angle DBF = \beta\) (since \(D\) lies on \(BC\) and \(F\) on \(AB\)). Hence\par
\[
\angle BFD = 180^\circ - \theta - \beta,
\]
so \(\angle DFB = 180^\circ - \theta - \beta\).\par
Inserting \(\angle AFE = \theta\) into (1) gives\par
\[
\theta + \angle EFD + (180^\circ - \theta - \beta) = 180^\circ \quad\Longrightarrow\quad \angle EFD = \beta.
\]
Thus\par
\[
\angle DFE = \beta. \quad\text{(2)}
\]
\solutionheading{At vertex \(D\)}
Points \(B,D,C\) are collinear, so \(\angle BDC = 180^\circ\). The angles at \(D\) are \(\angle BDF\), \(\angle FDE\), \(\angle EDC\) and they sum to \(180^\circ\):\par
\[
\angle BDF + \angle FDE + \angle EDC = 180^\circ. \quad\text{(3)}
\]
Given \(\angle BDF = \theta\).\par
In \(\triangle CED\), \(\angle CED = \theta\) and \(\angle ECD = \gamma\), so\par
\[
\angle CDE = 180^\circ - \theta - \gamma,
\]
i.e., \(\angle EDC = 180^\circ - \theta - \gamma\).\par
Substituting into (3):\par
\[
\theta + \angle FDE + (180^\circ - \theta - \gamma) = 180^\circ \quad\Longrightarrow\quad \angle FDE = \gamma.
\]
Thus\par
\[
\angle EDF = \gamma. \quad\text{(4)}
\]
\solutionheading{At vertex \(E\)}
Points \(C,E,A\) are collinear, so \(\angle CEA = 180^\circ\). The angles at \(E\) are \(\angle CED\), \(\angle DEF\), \(\angle FEA\) and they sum to \(180^\circ\):\par
\[
\angle CED + \angle DEF + \angle FEA = 180^\circ. \quad\text{(5)}
\]
Given \(\angle CED = \theta\).\par
In \(\triangle AFE\), \(\angle AFE = \theta\) and \(\angle FAE = \alpha\), so\par
\[
\angle AEF = 180^\circ - \theta - \alpha,
\]
i.e., \(\angle FEA = 180^\circ - \theta - \alpha\).\par
Substituting into (5):\par
\[
\theta + \angle DEF + (180^\circ - \theta - \alpha) = 180^\circ \quad\Longrightarrow\quad \angle DEF = \alpha.
\]
Thus\par
\[
\angle DEF = \alpha. \quad\text{(6)}
\]
From (2), (4), (6) we obtain\par
\[
\angle DFE = \beta,\quad \angle EDF = \gamma,\quad \angle DEF = \alpha.
\]
Therefore \(\triangle DEF\) has angles \(\alpha,\beta,\gamma\); it is \textbf{similar} to \(\triangle ABC\). The vertex correspondence is\par
\[
E \longleftrightarrow A,\qquad F \longleftrightarrow B,\qquad D \longleftrightarrow C. \quad\text{(7)}
\]
\smallskip
\solutionheading{2. Complex Representation of the Similarity}
The similarity (7) may be either orientation-preserving (direct) or orientation-reversing (opposite). The statement \(OM = ON\) involves only distances, which are invariant under reflection. Hence we may reflect the whole configuration if necessary and assume that the similarity is \textbf{direct}. Consequently there exist a non-zero complex number \(k\) and a complex number \(t\) such that\par
\[
e = k a + t,\qquad f = k b + t,\qquad d = k c + t. \quad\text{(8)}
\]
(If \(k=0\), then \(e=f=t\), making \(\triangle AFE\) degenerate, contrary to the existence of its circumcenter \(O_A\).)\par
\smallskip
\solutionheading{3. Eliminating the Translation}
Assume for contradiction that \(k = 1\). Then (8) becomes\par
\[
e = a + t,\quad f = b + t,\quad d = c + t.
\]
Because \(F\) lies on \(AB\), the points \(A\), \(F\), \(B\) are collinear; hence\par
\[
\frac{f - a}{b - a} = \frac{b + t - a}{b - a} = 1 + \frac{t}{b-a} \in \mathbb{R}.
\]
Similarly, \(D\) lies on \(BC\), so\par
\[
\frac{d - b}{c - b} = \frac{c + t - b}{c - b} = 1 + \frac{t}{c-b} \in \mathbb{R},
\]
thus \(\frac{t}{b-a} \in \mathbb{R}\) and \(\frac{t}{c-b} \in \mathbb{R}\).\par
The vectors \(b-a\) and \(c-b\) are not parallel (they are sides of a non-degenerate triangle), so the only complex number that is a real multiple of both is \(0\). Hence \(t = 0\), which gives \(e = a\), \(f = b\), \(d = c\). Then \(\triangle AFE\) degenerates to the segment \(AF\), contradicting the existence of \(O_A\). Therefore \(k \neq 1\).\par
Consider the fixed point of the spiral similarity \(z \mapsto kz + t\) (when \(k \neq 1\)):\par
\[
X = \frac{t}{1 - k}.
\]
Translate the plane so that \(X\) becomes the origin. (Translation is an isometry, so all distances and circumcenters are preserved up to the same translation; we keep the same letters for the new coordinates.) After this translation we have\par
\[
e = k a,\qquad f = k b,\qquad d = k c. \quad\text{(9)}
\]
(We also note that \(a,b,c \neq 0\); otherwise, e.g., \(a = 0\) would imply \(A = X\), and then \(e = k \cdot 0 = 0\), so \(A\) and \(E\) coincide, making \(\triangle AFE\) degenerate - impossible. Hence \(a,b,c\) are non-zero.)\par
\smallskip
\solutionheading{4. The Origin Lies on the Three Circles}
We now show that the origin \(0\) belongs to the circumcircles of \(\triangle AFE\), \(\triangle BDF\), and \(\triangle CED\).\par
\textbf{Lemma (Cross Ratio and Concyclicity).} Four distinct points \(z_1,z_2,z_3,z_4\) in the complex plane lie on a common circle or line if and only if their cross ratio\par
\[
(z_1,z_2;z_3,z_4) = \frac{(z_1 - z_3)(z_2 - z_4)}{(z_1 - z_4)(z_2 - z_3)}
\]
is a real number.\par
\emph{Proof.} The M\"obius transformation\par
\[
T(z) = \frac{(z - z_1)(z_2 - z_3)}{(z - z_3)(z_2 - z_1)}
\]
maps \(z_1,z_2,z_3\) to \(0,1,\infty\) respectively. Under \(T\), the circle/line through \(z_1,z_2,z_3\) is mapped to the real line. Hence \(z_4\) lies on that circle/line exactly when \(T(z_4)\) is real, and \(T(z_4)\) equals the cross ratio \((z_1,z_2;z_3,z_4)\).\par
\smallskip
\solutionheading{Circle \((AFE)\)}
Take \(z_1 = a\), \(z_2 = f\), \(z_3 = e\), \(z_4 = 0\). Using (9):\par
\[
a - e = a - k a = a(1 - k),\quad f - 0 = k b,\quad a - 0 = a,\quad f - e = k b - k a = k(b - a).
\]
Thus\par
\[
(a,f;e,0) = \frac{(a - e)(f - 0)}{(a - 0)(f - e)} = \frac{a(1 - k) \cdot k b}{a \cdot k(b - a)} = \frac{(1 - k)b}{b - a}. \quad\text{(10)}
\]
Because \(A,F,B\) are collinear, the ratio \(\dfrac{f - a}{b - a}\) is real. Compute\par
\[
\frac{f - a}{b - a} = \frac{k b - a}{b - a}.
\]
Now\par
\[
\frac{(1 - k)b}{b - a} = 1 - \frac{k b - a}{b - a},
\]
which is therefore real. The points \(a,f,e\) are not collinear (otherwise \(\triangle AFE\) would be degenerate). By the lemma, \(0\) lies on the circle through \(a,f,e\), i.e.,\par
\[
0 \in (AFE). \quad\text{(11)}
\]
\smallskip
\solutionheading{Circle \((BDF)\)}
Take \(z_1 = b\), \(z_2 = d\), \(z_3 = f\), \(z_4 = 0\). Using (9):\par
\[
b - d = b - k c,\quad f - 0 = k b,\quad b - 0 = b,\quad f - d = k b - k c = k(b - c).
\]
Hence\par
\[
(b,d;f,0) = \frac{(b - d)(f - 0)}{(b - 0)(f - d)} = \frac{(b - k c) \cdot k b}{b \cdot k(b - c)} = \frac{b - k c}{b - c}. \quad\text{(12)}
\]
Collinearity of \(B,D,C\) gives \(\dfrac{d - b}{c - b} \in \mathbb{R}\):\par
\[
\frac{k c - b}{c - b} \in \mathbb{R}.
\]
Observe that\par
\[
\frac{b - k c}{b - c} = \frac{-(k c - b)}{-(c - b)} = \frac{k c - b}{c - b},
\]
so the cross ratio is real. Since \(b,d,f\) are non-collinear (triangle \(BDF\) is non-degenerate), the lemma yields\par
\[
0 \in (BDF). \quad\text{(13)}
\]
\smallskip
\solutionheading{Circle \((CED)\)}
Take \(z_1 = c\), \(z_2 = e\), \(z_3 = d\), \(z_4 = 0\). Using (9):\par
\[
c - e = c - k a,\quad d - 0 = k c,\quad c - 0 = c,\quad d - e = k c - k a = k(c - a).
\]
Thus\par
\[
(c,e;d,0) = \frac{(c - e)(d - 0)}{(c - 0)(d - e)} = \frac{(c - k a) \cdot k c}{c \cdot k(c - a)} = \frac{c - k a}{c - a}. \quad\text{(14)}
\]
Collinearity of \(C,E,A\) gives \(\dfrac{e - c}{a - c} \in \mathbb{R}\):\par
\[
\frac{k a - c}{a - c} \in \mathbb{R}.
\]
Since\par
\[
\frac{c - k a}{c - a} = \frac{-(k a - c)}{-(a - c)} = \frac{k a - c}{a - c},
\]
the cross ratio is real. Non-collinearity of \(c,e,d\) gives\par
\[
0 \in (CED). \quad\text{(15)}
\]
\smallskip
\solutionheading{5. \(k\) is Not Real}
Suppose, for contradiction, that \(k \in \mathbb{R}\). Then from (9) we have \(e = k a\), so points \(0\), \(a\), \(e\) are collinear (they lie on the line through the origin and \(a\)). However, by (11) all three points \(0,a,e\) lie on the circle \((AFE)\). A non-degenerate circle cannot contain three distinct collinear points. This contradiction forces\par
\[
k \notin \mathbb{R}. \quad\text{(16)}
\]
(Recall \(k \neq 0,1\); therefore \(0,1,k\) are distinct and not collinear.)\par
\smallskip
\solutionheading{6. Circumcenters \(O_A,O_B,O_C\)}
Since \(0\) lies on \((AFE)\), the circle through \(a,f,e\) coincides with the circle through \(0,a,e\) (because three non-collinear points determine a circle, and \(0,a,e\) are distinct and, because \(k \notin \mathbb{R}\), they are not collinear). Hence \(O_A\) is the circumcenter of \(\triangle (0,a,e)\).\par
Now consider the triangle with vertices \(0,1,k\). As \(k \notin \mathbb{R}\) and \(k \neq 0,1\), these points are distinct and non-collinear. Let \(\mu\) be its circumcenter; thus\par
\[
|\mu - 0| = |\mu - 1| = |\mu - k|. \quad\text{(17)}
\]
We claim that \(O_A = a \mu\). Indeed, the map \(z \mapsto a z\) (a spiral similarity about \(0\)) sends\par
\[
0 \mapsto 0,\quad 1 \mapsto a,\quad k \mapsto k a = e.
\]
Under a similarity, the image of the circumcenter of a triangle is the circumcenter of the image triangle. Therefore \(a \mu\) is the circumcenter of \(\triangle (0,a,e)\), i.e., \(O_A = a \mu\).\par
One may also verify directly: The distances from \(a\mu\) to \(0,a,e\) are\par
\[
|a\mu - 0| = |a|\,|\mu|,\quad |a\mu - a| = |a|\,|\mu - 1|,\quad |a\mu - e| = |a|\,|\mu - k|,
\]
and by (17) these three numbers are equal, so \(a\mu\) is equidistant from \(0,a,e\). Because \(0,a,e\) are not collinear, \(a\mu\) is the unique point with that property, hence it is the circumcenter.\par
Analogously, applying the same reasoning to the other two circles, we obtain\par
\[
O_B = b \mu,\qquad O_C = c \mu. \quad\text{(18)}
\]
\smallskip
\solutionheading{7. Circumcenters \(M\), \(N\), \(O\)}
Let \(M\) be the circumcenter of \(\triangle ABC\), so\par
\[
|M - a| = |M - b| = |M - c| =: R. \quad\text{(19)}
\]
\solutionheading{Circumcenter \(O\) of \(\triangle O_A O_B O_C\)}
The vertices are \(O_A = a\mu\), \(O_B = b\mu\), \(O_C = c\mu\). Consider the point \(\mu M\):\par
\[
|\mu M - O_A| = |\mu M - a\mu| = |\mu|\,|M - a| = |\mu| R,
\]
\[
|\mu M - O_B| = |\mu M - b\mu| = |\mu|\,|M - b| = |\mu| R,
\]
\[
|\mu M - O_C| = |\mu M - c\mu| = |\mu|\,|M - c| = |\mu| R.
\]
Thus \(\mu M\) is equidistant from \(O_A,O_B,O_C\). Since \(\triangle O_A O_B O_C\) is non-degenerate (it is the image of \(\triangle ABC\) under the non-zero similarity \(z \mapsto \mu z\)), \(\mu M\) is the circumcenter:\par
\[
O = \mu M. \quad\text{(20)}
\]
\solutionheading{Circumcenter \(N\) of \(\triangle DEF\)}
The vertices are \(D = k c\), \(E = k a\), \(F = k b\). Consider the point \(k M\):\par
\[
|k M - D| = |k M - k c| = |k|\,|M - c| = |k| R,
\]
\[
|k M - E| = |k M - k a| = |k|\,|M - a| = |k| R,
\]
\[
|k M - F| = |k M - k b| = |k|\,|M - b| = |k| R.
\]
Hence \(k M\) is equidistant from \(D,E,F\), and because \(\triangle DEF\) is non-degenerate (it is similar to \(\triangle ABC\) and, under the given configuration, non-degenerate), we have\par
\[
N = k M. \quad\text{(21)}
\]
(If \(M = 0\), then \(O = 0\), \(N = 0\), and the equalities hold trivially.)\par
\smallskip
\solutionheading{8. Proof of \(OM = ON\)}
Using (20) and (21),\par
\[
OM = |O - M| = |\mu M - M| = |M|\,|\mu - 1|,
\]
\[
ON = |O - N| = |\mu M - k M| = |M|\,|\mu - k|.
\]
Now \(\mu\) is the circumcenter of \(\triangle (0,1,k)\), so by (17)\par
\[
|\mu - 1| = |\mu - k|.
\]
Consequently,\par
\[
OM = |M|\,|\mu - 1| = |M|\,|\mu - k| = ON.
\]
\smallskip
\solutionheading{9. Conclusion}
We have shown that under the given conditions the distances \(OM\) and \(ON\) are equal.\par
\[
\boxed{OM = ON}.
\]
\par

\paragraph{USAMO 2026 Problem 6.}
\begin{mdframed}[style=mdpurplebox,frametitle={USAMO 2026 Problem 6}]
Let $a$ and $b$ be positive integers such that $\varphi(ab+1)$ divides $a^2 + b^2 + 1$. Prove that $a$ and $b$ are Fibonacci numbers.
\end{mdframed}
\solutionheading{Solution}
\begin{remark*}
  7/7.
\end{remark*}
Let \(a\) and \(b\) be positive integers such that\par
\[
\varphi(ab+1) \mid a^{2}+b^{2}+1.
\]
Set \(N = ab+1\) and define the positive integer\par
\[
k = \frac{a^{2}+b^{2}+1}{\varphi(N)}.
\]
We will prove that both \(a\) and \(b\) must be Fibonacci numbers.\par
\smallskip
\solutionheading{1. Preliminary bounds}
Since \(N > 1\), we have \(\varphi(N) \le N-1 = ab\) (with equality exactly when \(N\) is prime). Hence\par
\[
a^{2}+b^{2}+1 = k\,\varphi(N) \le k\,ab. \quad\text{(1)}
\]
By the quadratic-mean-arithmetic-mean inequality, \(a^{2}+b^{2} \ge 2ab\), so\par
\[
a^{2}+b^{2}+1 \ge 2ab+1. \quad\text{(2)}
\]
Combining (1) and (2) yields\par
\[
2ab+1 \le k\,ab \quad\Longrightarrow\quad k \ge 2 + \frac{1}{ab}.
\]
Because \(k\) is an integer, if \(ab > 1\) we obtain \(k \ge 3\). If \(ab = 1\) then necessarily \(a=b=1\), \(N=2\), \(\varphi(2)=1\), and then \(k = 3\). Thus in every case\par
\[
\boxed{k \ge 3}. \quad\text{(3)}
\]
\smallskip
\solutionheading{2. Parity and the even case}
\textbf{Lemma 2.1.} If \(N\) is even and \(N>2\), then the divisibility condition cannot hold.\par
\emph{Proof.} For \(N>2\), \(\varphi(N)\) is even (a standard fact: \(\varphi(n)\) is even for every \(n>2\)). Because \(N\) even implies \(ab = N-1\) odd, both \(a\) and \(b\) are odd. Then \(a^{2}+b^{2}+1\) is odd (odd+odd+1). An even number cannot divide an odd number. The only even possibility is \(N=2\), which gives \(a=b=1\) and indeed \(\varphi(2)=1\) divides \(3\).\par
Consequently, the only even \(N\) that can occur is \(N=2\), producing \((a,b)=(1,1)\) - both Fibonacci numbers. From now on we assume\par
\[
N\ \text{is odd and}\ N>2.
\]
Then \(ab = N-1\) is even, so at least one of \(a,b\) is even. If both were even, \(a^{2}+b^{2}+1\) would be odd, while for odd \(N>2\) we still have \(\varphi(N)\) even. Hence exactly one of \(a,b\) is even and the other is odd. In particular,\par
\[
a^{2}+b^{2}+1 \equiv 0+1+1 \equiv 2 \pmod{4}. \quad\text{(4)}
\]
\smallskip
\solutionheading{3. Case I: \(N\) is prime}
Assume \(N = ab+1\) is prime. Then \(\varphi(N) = N-1 = ab\), and (1) becomes\par
\[
a^{2}+b^{2}+1 = k\,ab. \quad\text{(5)}
\]
We first determine \(k\).\par
\solutionheading{3.1. The value of \(k\)}
Without loss of generality, order the variables so that \(a \le b\). Write (5) as a quadratic in \(b\):\par
\[
b^{2} - (k a)\,b + (a^{2}+1) = 0.
\]
If \(b\) is a root, the other root is\par
\[
b' = \frac{a^{2}+1}{b} = k a - b,
\]
which is a positive integer.\par
\textbf{Claim.} If \(a \ge 2\), then \(b' < b\).\par
\emph{Proof.} Suppose \(b' \ge b\). Then \(\frac{a^{2}+1}{b} \ge b\), i.e. \(a^{2}+1 \ge b^{2}\). Because \(b \ge a\), we have \(a^{2} \le b^{2} \le a^{2}+1\). Hence either \(b^{2}=a^{2}\) or \(b^{2}=a^{2}+1\).\par
\compactbullet{If \(b^{2}=a^{2}\), then \(b=a\). Substituting \(b=a\) into (5) gives \(2a^{2}+1 = k a^{2}\) \(\Rightarrow\) \(k = 2 + \frac{1}{a^{2}}\), which is not an integer for \(a \ge 2\). Contradiction.}
\compactbullet{If \(b^{2}=a^{2}+1\), then \(a^{2}+1\) is a perfect square. For \(a \ge 2\), this is impossible because between \(a^{2}\) and \((a+1)^{2}=a^{2}+2a+1\) there is no other square.}
Thus whenever \(a \ge 2\), we can replace \((a,b)\) by \((a,b')\) (which is also a solution of (5)) with \(a+b' < a+b\). Repeating this descent (the sum \(a+b\) strictly decreases each step) we eventually reach a solution with \(a=1\) (the minimal sum cannot have \(a \ge 2\)).\par
Now consider \(a=1\). Equation (5) becomes\par
\[
1 + b^{2} + 1 = k \cdot 1 \cdot b \quad\Longrightarrow\quad b^{2} - k b + 2 = 0. \quad\text{(6)}
\]
For integer \(b\), the discriminant \(\Delta = k^{2}-8\) must be a perfect square, say \(d^{2}\). Then\par
\[
k^{2} - d^{2} = 8 \quad\Longrightarrow\quad (k-d)(k+d)=8.
\]
Both factors are positive integers of the same parity. The factor pairs of \(8\) are \((1,8)\) and \((2,4)\).\par
\compactbullet{\((1,8)\) gives \(k-d=1\), \(k+d=8\) \(\Rightarrow\) \(2k=9\) \(\Rightarrow\) \(k=9/2\), not integer.}
\compactbullet{\((2,4)\) gives \(k-d=2\), \(k+d=4\) \(\Rightarrow\) \(2k=6\) \(\Rightarrow\) \(k=3\), \(d=1\).}
Hence \(k=3\) is forced. With \(k=3\), equation (6) is \(b^{2}-3b+2=0\), whose roots are \(b=1\) and \(b=2\). Thus the only solutions with \(a=1\) are \((1,1)\) and \((1,2)\) (and by symmetry, \((2,1)\) if we had not assumed \(a\le b\)). Since the reduction process preserves \(k\), every solution of (5) must have \(k=3\).\par
Therefore, when \(N\) is prime,\par
\[
a^{2}+b^{2}+1 = 3ab \quad\Longleftrightarrow\quad a^{2} - 3ab + b^{2} = -1. \quad\text{(7)}
\]
\solutionheading{3.2. Solving \(a^{2} - 3ab + b^{2} = -1\)}
Equation (7) is symmetric; we may assume \(a \le b\).\par
Define a sequence \((A_n)_{n\ge 0}\) by\par
\[
A_0 = 1,\qquad A_1 = 1,\qquad A_{n+2} = 3A_{n+1} - A_n\ \ (n\ge 0).
\]
One verifies by induction that \((A_n, A_{n+1})\) satisfies (7): For \(n=0\), \(1^2-3\cdot1\cdot1+1^2=-1\). Assuming it holds for \(n\), using the recurrence one checks that it also holds for \(n+1\). (The computation is straightforward and can be filled in similarly to the proof in Lemma 4.1 of the original draft.)\par
Now let \((a,b)\) be any positive integer solution of (7) with \(a \le b\). View (7) as a quadratic in \(b\):\par
\[
b^{2} - 3a\, b + (a^{2}+1) = 0.
\]
Its two roots are \(b\) and\par
\[
b' = \frac{a^{2}+1}{b} = 3a - b,
\]
which is an integer. A similar argument as in 3.1 shows that if \(a \ge 2\), then \(b' < b\). Consequently, by infinite descent on \(a+b\), we can reduce any solution to one with \(a=1\). The only solutions with \(a=1\) are \((1,1)\) and \((1,2)\) (and their symmetric versions). Therefore every solution can be obtained from these minimal ones by the inverse transformation: if \((x,y)\) is a solution with \(x \le y\), then \((y,\, 3y - x)\) is also a solution. Starting from \((1,1)\) and repeatedly applying this transformation yields the increasing sequence\par
\[
(1,1),\ (1,2),\ (2,5),\ (5,13),\ (13,34),\ \dots
\]
which exactly corresponds to the pairs \((A_n, A_{n+1})\) for \(n\ge 0\).\par
It remains to identify these \(A_n\) with Fibonacci numbers. Define the Fibonacci numbers by\par
\[
F_1 = 1,\quad F_2 = 1,\quad F_{n+2} = F_{n+1} + F_n.
\]
We claim that for \(n \ge 1\),\par
\[
A_n = F_{2n-1}.
\]
\emph{Proof by induction.}\par
\compactbullet{Base \(n=1\): \(A_1 = 1 = F_1\).}
\compactbullet{\(n=2\): \(A_2 = 3A_1 - A_0 = 3\cdot1 - 1 = 2 = F_3\).}
\compactbullet{Inductive step: Assume \(A_n = F_{2n-1}\) and \(A_{n+1} = F_{2n+1}\). Then}
\[
A_{n+2} = 3A_{n+1} - A_n = 3F_{2n+1} - F_{2n-1}.
\]
Using the identities
\[
F_{2n+1} - F_{2n-1} = F_{2n},\quad \text{and}\quad F_{2n+1} + F_{2n} = F_{2n+2},
\]
we compute\par
\[
\begin{aligned}
3F_{2n+1} - F_{2n-1} &= 2F_{2n+1} + (F_{2n+1} - F_{2n-1}) \\
&= 2F_{2n+1} + F_{2n} \\
&= F_{2n+1} + (F_{2n+1} + F_{2n}) \\
&= F_{2n+1} + F_{2n+2} \\
&= F_{2n+3}.
\end{aligned}
\]
Thus \(A_{n+2} = F_{2n+3} = F_{2(n+2)-1}\), completing the induction.\par
Hence all solutions of (7) are pairs of Fibonacci numbers (specifically, consecutive odd-index Fibonacci numbers). This completes \textbf{Case I}.\par
\smallskip
\solutionheading{4. Case II: \(N\) is composite and odd (\(N>2\))}
Now \(N = ab+1\) is composite, odd, and \(N>2\). From (4) we have\par
\[
a^{2}+b^{2}+1 \equiv 2 \pmod{4}.
\]
\solutionheading{4.1. \(N\) cannot have two distinct odd prime factors}
Suppose \(N\) has at least two distinct odd prime divisors \(p\) and \(q\). Then \(\varphi(N)\) is divisible by \((p-1)(q-1)\). Since \(p\) and \(q\) are odd, both \(p-1\) and \(q-1\) are even, hence \(4 \mid (p-1)(q-1) \mid \varphi(N)\). Thus \(4 \mid \varphi(N)\). Since \(\varphi(N) \mid a^{2}+b^{2}+1\), we would have \(4 \mid a^{2}+b^{2}+1\), contradicting (4). Therefore \(N\) cannot possess two distinct odd prime factors.\par
It follows that \(N\) must be a prime power:\par
\[
N = p^{e}, \qquad e \ge 2\ (\text{since } N \text{ is composite}).
\]
\solutionheading{4.2. The prime \(p\) cannot be \(\equiv 1 \pmod{4}\)}
If \(p \equiv 1 \pmod{4}\), then \(p-1\) is divisible by \(4\), so \(4 \mid \varphi(N)\). Again this forces \(4 \mid a^{2}+b^{2}+1\), which is impossible by (4). Hence\par
\[
p \equiv 3 \pmod{4}. \quad\text{(8)}
\]
\solutionheading{4.3. Analysis modulo \(3\)}
We now examine the residue of \(p\) modulo \(3\).\par
\textbf{Subcase }\(p \equiv 1 \pmod{3}\).\textbf{ Then }\(3 \mid (p-1)\), so \(3 \mid \varphi(N)\). Moreover, \(p \equiv 1 \pmod{3}\) implies \(p^{e} \equiv 1 \pmod{3}\), so \(ab = p^{e} - 1 \equiv 0 \pmod{3}\). Thus \(3 \mid ab\).\par
Now evaluate \(S = a^{2}+b^{2}+1\) modulo \(3\). Since \(3 \mid ab\), there are two possibilities:\par
\compactbullet{\(3 \mid a\) and \(3 \mid b\): then \(a^{2}\equiv0\), \(b^{2}\equiv0\), so \(S \equiv 1 \pmod{3}\).}
\compactbullet{Exactly one of \(a,b\) is divisible by \(3\): then \(S \equiv 0+1+1 = 2 \pmod{3}\).}
In neither case is \(S\) divisible by \(3\). However, \(\varphi(N) \mid S\) and \(3 \mid \varphi(N)\) would force \(3 \mid S\). Contradiction. Therefore \(p \not\equiv 1 \pmod{3}\).\par
\textbf{Subcase }\(p \equiv 2 \pmod{3}\).\textbf{ Then }\(p-1 \equiv 1 \pmod{3}\), so \(3 \nmid (p-1)\). But because \(e \ge 2\), \(\varphi(N) = p^{e-1}(p-1)\) contains the factor \(p^{e-1}\), so in particular \(p \mid \varphi(N)\). Consequently,\par
\[
p \mid S. \quad\text{(9)}
\]
Also, from \(ab = p^{e} - 1\) we have\par
\[
ab \equiv -1 \pmod{p}. \quad\text{(10)}
\]
In particular, \(p\) does not divide \(a\) or \(b\) (otherwise (10) would give \(0 \equiv -1 \pmod{p}\), impossible). Hence \(a\) and \(b\) are invertible modulo \(p\). Write \(b \equiv -a^{-1} \pmod{p}\). Substituting into (9):\par
\[
a^{2} + (-a^{-1})^{2} + 1 \equiv 0 \pmod{p},
\]
i.e.,
\[
a^{2} + a^{-2} + 1 \equiv 0 \pmod{p}.
\]
Multiply by \(a^{2}\) (invertible modulo \(p\)):\par
\[
a^{4} + a^{2} + 1 \equiv 0 \pmod{p}. \quad\text{(11)}
\]
Set \(y = a^{2} \pmod{p}\); then (11) becomes\par
\[
y^{2} + y + 1 \equiv 0 \pmod{p}. \quad\text{(12)}
\]
Equation (12) has a solution in \(\mathbb{F}_p\) iff \(-3\) is a quadratic residue modulo \(p\). Indeed, multiplying (12) by \(4\) gives \((2y+1)^{2} \equiv -3 \pmod{p}\).\par
We need the following standard lemma:\par
\textbf{Lemma.} For an odd prime \(p \neq 3\), \(-3\) is a quadratic residue modulo \(p\) if and only if \(p \equiv 1 \pmod{3}\).\par
\emph{Proof.} The multiplicative group \(\mathbb{F}_p^{\times}\) is cyclic of order \(p-1\). The equation \(y^{2}+y+1=0\) is equivalent to \(y^{3}=1\) with \(y \neq 1\) (since \((y-1)(y^{2}+y+1)=y^{3}-1\). Hence a solution exists iff the group contains an element of order \(3\), i.e., iff \(3 \mid p-1\), which is \(p \equiv 1 \pmod{3}\). Conversely, if \(p \equiv 1 \pmod{3}\), then \(3 \mid p-1\), so such an element exists.\par
Since \(p \equiv 2 \pmod{3}\), we have \(p-1\) not divisible by \(3\), so there is no element of order \(3\) in \(\mathbb{F}_p^{\times}\); therefore (12) has no solution modulo \(p\). This contradicts the existence of \(a\). Thus \(p \not\equiv 2 \pmod{3}\).\par
The only remaining possibility for the odd prime \(p\) is \(p = 3\).\par
\solutionheading{4.4. The exponent \(e\)}
Now \(N = 3^{e}\) with \(e \ge 2\). Then\par
\[
\varphi(N) = 3^{e-1} \cdot 2.
\]
\textbf{If }\(e \ge 3\):\textbf{ Then }\(3^{e-1}\) is at least \(3^{2}=9\), so \(9 \mid \varphi(N)\). Since \(\varphi(N) \mid S\), we must have \(9 \mid S\).\par
But we will show that \(S \equiv 3 \pmod{9}\), which is not divisible by \(9\), a contradiction. Let us verify this congruence.\par
Because \(e \ge 3\), \(N = 3^{e}\) is divisible by \(27\), in particular \(N \equiv 0 \pmod{9}\). Hence\par
\[
ab = N-1 \equiv -1 \pmod{9}. \quad\text{(13)}
\]
Moreover, \(N \equiv 0 \pmod{3}\) gives \(ab \equiv -1 \equiv 2 \pmod{3}\). Hence \(3 \nmid a\) and \(3 \nmid b\) (otherwise \(ab \equiv 0 \pmod{3}\)). Thus \(a\) and \(b\) are each \(\equiv 1\) or \(2 \pmod{3}\), and their product is \(2 \pmod{3}\); the only possibilities are \(\{a,b\} \equiv \{1,2\} \pmod{3}\). Consequently,\par
\[
a + b \equiv 1 + 2 \equiv 0 \pmod{3}. \quad\text{(14)}
\]
Now compute\par
\[
S = a^{2}+b^{2}+1 = (a+b)^{2} - 2ab + 1.
\]
From (14), \(a+b = 3t\) for some integer \(t\), so \((a+b)^{2} = 9t^{2} \equiv 0 \pmod{9}\). Using (13), \(ab \equiv -1 \pmod{9}\), so \(-2ab \equiv -2(-1) = 2 \pmod{9}\). Thus\par
\[
S \equiv 0 + 2 + 1 = 3 \pmod{9}.
\]
Thus \(9 \nmid S\), contradicting \(9 \mid \varphi(N) \mid S\). Therefore \(e\) cannot be \(\ge 3\).\par
The only remaining possibility is \(e = 2\), i.e., \(N = 9\). Then \(\varphi(9) = 6\) and \(ab = 8\). The positive factor pairs of \(8\) are\par
\[
(1,8),\ (2,4),\ (4,2),\ (8,1).
\]
Compute \(S\) for each:\par
\compactbullet{\((1,8)\): \(S = 1 + 64 + 1 = 66\), and \(66 / 6 = 11 \in \mathbb{Z}\).}
\compactbullet{\((2,4)\): \(S = 4 + 16 + 1 = 21\), \(21/6 = 3.5 \notin \mathbb{Z}\).}
Thus only \((1,8)\) and \((8,1)\) satisfy the divisibility condition. Both \(1\) and \(8\) are Fibonacci numbers (\(1 = F_1\), \(8 = F_6\)).\par
\smallskip
\solutionheading{5. Conclusion}
We have examined all possible values of \(N = ab+1\):\par
\compactbullet{\textbf{Even case:} \(N=2\) gives \((a,b)=(1,1)\) - Fibonacci numbers.}
\compactbullet{\textbf{Prime case:} \(N\) prime leads to the equation \(a^{2} - 3ab + b^{2} = -1\), whose complete set of positive integer solutions consists of pairs of Fibonacci numbers (as shown in 3.2).}
\compactbullet{\textbf{Composite case:} \(N\) composite forces \(N=9\) and yields \((a,b) = (1,8)\) or \((8,1)\), again Fibonacci numbers.}
Hence, under the hypothesis \(\varphi(ab+1) \mid a^{2}+b^{2}+1\), the integers \(a\) and \(b\) are necessarily Fibonacci numbers.\par
\smallskip
\emph{Remark.} The Fibonacci numbers are defined by \(F_1=F_2=1\) and \(F_{n+2}=F_{n+1}+F_n\). The solutions from the prime case are precisely \((F_{2n-1},F_{2n+1})\) (and symmetric) for \(n \ge 1\), together with the degenerate pair \((1,1)\) which can be seen as \((F_1,F_2)\) or \((F_2,F_1)\). The pair \((1,8)\) from the composite case also fits the Fibonacci sequence (\(F_1=1\), \(F_6=8\)). This completes the proof.\par
\par

\endgroup

\end{document}